\documentclass{article}

\usepackage[preprint]{neurips_2026}
\usepackage[utf8]{inputenc}
\usepackage[T1]{fontenc}
\usepackage{hyperref}
\usepackage{xurl}
\usepackage{booktabs}
\usepackage{amsfonts}
\usepackage{nicefrac}
\usepackage[final,
            protrusion=true,
            expansion=true,
            factor=1100,
            stretch=10,
            shrink=10,
            step=1,
            tracking=smallcaps
           ]{microtype}
\SetTracking{encoding={*}, shape=sc}{40}
\UseMicrotypeSet[protrusion]{basicmath}

\usepackage{graphicx}
\usepackage{natbib}
\usepackage{doi}
\renewcommand*{\doi}[1]{doi:\href{https://doi.org/#1}{{\urlstyle{same}\nolinkurl{#1}}}}

\usepackage{amsmath,amssymb}
\usepackage{textcomp}
\usepackage{array}
\usepackage{calc}
\usepackage{multirow}
\usepackage{longtable}
\usepackage{etoolbox}
\usepackage{xcolor}
\usepackage{tikz}
\usepackage{adjustbox}
\usetikzlibrary{arrows.meta,positioning,decorations.pathreplacing,calc,shapes.geometric,fit}

\hypersetup{colorlinks=true,citecolor=blue,linkcolor=blue,urlcolor=blue}
\makeatletter
\g@addto@macro\normalsize{%
  \setlength\abovedisplayskip{8pt plus 2pt minus 2pt}%
  \setlength\belowdisplayskip{8pt plus 2pt minus 2pt}%
  \setlength\abovedisplayshortskip{8pt plus 2pt minus 2pt}%
  \setlength\belowdisplayshortskip{8pt plus 2pt minus 2pt}}
\patchcmd{\LT@makecaption}{\sbox\@tempboxa}{\normalsize\sbox\@tempboxa}{}{\PackageError{main}{LT caption patch failed}{}}
\makeatother
\definecolor{supplier}{HTML}{F76C6C}
\definecolor{intermediate}{HTML}{B9BEF4}
\definecolor{manuf}{HTML}{C3E59F}
\definecolor{threepl}{HTML}{F9D68D}
\definecolor{pathblue}{HTML}{4A90D9}
\definecolor{labelblue}{HTML}{1F6FC0}
\definecolor{future}{HTML}{DDE0FA}
\definecolor{bracecol}{HTML}{5B9BD5}
\definecolor{dbfill}{HTML}{DAE8FC}\definecolor{dbdraw}{HTML}{6C8EBF}
\definecolor{mlyellow}{HTML}{FFF2CC}\definecolor{mlyellowdraw}{HTML}{D6B656}
\definecolor{mlpanel}{HTML}{F4F3FD}\definecolor{mlpaneldraw}{HTML}{E3E1F7}\definecolor{mllabel}{HTML}{6666FF}

\expandafter\ifx\csname iflatexml\endcsname\relax
  \expandafter\newif\csname iflatexml\endcsname
\fi

\iflatexml
  \newcommand{\orcidicon}[1]{}%
\else
  \newcommand{\orcidicon}[1]{\href{https://orcid.org/#1}{\includegraphics[scale=0.06]{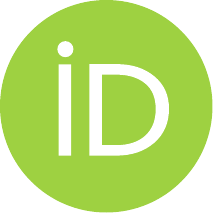}}\hspace{1mm}}
\fi

\title{A hybrid Graph-based Machine Learning Model for predicting Delivery Times in Supply Chain Logistics}

\date{November 2, 2026}

\iflatexml

\author{\orcidicon{0009-0003-3273-8931}Mattia Volpato\\ Cefriel
\and \orcidicon{0009-0008-3752-4680}Andrea Giuseppe Landella\\ Cefriel
\and \orcidicon{0009-0004-9316-2689}Bruno Puzzolante\thanks{Corresponding author. \texttt{bruno.puzzolante@cefriel.com}}\\ Cefriel}
\else

\author{%
	\orcidicon{0009-0003-3273-8931}Mattia Volpato\thanks{Work performed while at Cefriel.}%
	\hspace{1.5em}\orcidicon{0009-0008-3752-4680}Andrea Giuseppe Landella\footnotemark[1]%
	\hspace{1.5em}\orcidicon{0009-0004-9316-2689}Bruno Puzzolante\thanks{Cefriel, Viale Sarca 226, 20126 Milan, Italy}\, \thanks{Corresponding author. \texttt{bruno.puzzolante@cefriel.com}}
}
\fi

\makeatletter
\newcommand{\keywords}[1]{\par\addvspace\medskipamount{\rightskip=0pt plus1cm
  \def\and{\ifhmode\unskip\nobreak\fi\ $\cdot$ }\noindent\textbf{Keywords:}\enskip\ignorespaces#1\par}\addvspace\medskipamount}
\makeatother

\hypersetup{
pdftitle={A hybrid Graph-based Machine Learning Model for predicting Delivery Times in Supply Chain Logistics},
pdfsubject={cs.LG, math.OC},
pdfauthor={Mattia Volpato, Andrea Giuseppe Landella, Bruno Puzzolante},
pdfkeywords={Supply Chain Management, Machine Learning, Graph-Based Modelling, Supply Chain Operations, Logistics Processes, Supply Chain Resilience, Performance Indicators, Manufacturing-as-a-Service},
}

\begin{document}
\maketitle

\vskip\bigskipamount

\begin{abstract}
Global supply chains are increasingly vulnerable to disruptions arising from external events, such as natural disasters and geopolitical crises, and from internal operational inefficiencies, including supplier delays and bottlenecks in order shipping. This work, conducted within the Horizon Europe project M4ESTRO, introduces a hybrid modelling approach combining graph theory and machine learning to represent supply chain logistics operations, with a focus on predicting delays in inbound flows between industrial suppliers and manufacturing entities. The various stages of order fulfilment are modelled through a formal mathematical framework, by defining disruption indicators computed from both historical and real-time data drawn from heterogeneous sources, including traffic and weather conditions and supplier dispatch readiness. Building on a graph-based representation of the supply chain, stochastic methods based on Markov chains are applied to estimate delivery times, detect schedule deviations, quantify prediction uncertainty, and produce interpretable metrics to support operational decision-making. Within the M4ESTRO project, the proposed framework was applied to build a web application for real-time tracking of inbound logistics, and was validated in a use case with an industrial project partner. The web application successfully enabled early detection of logistics disruptions along the supply chain and provided stakeholders with the visibility needed to act before delays propagated. The underlying model proved effective as the basis of a resilient, data-driven logistics tracking system, aligned with the digital transformation objectives of industry and with Manufacturing-as-a-Service paradigms.
\end{abstract}

\vskip\bigskipamount


\clearpage
\section{Introduction}\label{introduction}

Global supply chains have entered a period in which disruption is no longer an exceptional condition but a structural feature of the operating environment. The COVID-19 pandemic produced measurable contractions in trade volumes and exposed the fragility of tightly coupled sourcing networks~\citep{ref30_Verschuur2021,ref31_Ivanov2020}; compounding geopolitical events have forced firms to redesign supplier footprints under conditions of sustained uncertainty~\citep{ref32_Roscoe2022}; and climate-related extreme events increasingly interrupt transport corridors and production sites on timescales too short for conventional planning cycles to absorb~\citep{ref33_Ghadge2020}. What distinguishes these events from ordinary variability in either demand or supply is not only their magnitude but their propagation, that is, a localized interruption cascades downstream through the network; an effect widely studied as the ripple effect in supply chain management~\citep{ref34_Ivanov2021}. Therefore, managing this environment, which is now frequently characterized as Volatile, Uncertain, Complex, and Ambiguous (VUCA)~\citep{ref35_Bennett2014}, has shifted resilience from a desirable property to an operational requirement. Supply chain resilience is commonly understood as the capacity of a system to anticipate disruptive events, absorb their impact, adapt its configuration in response, and recover its intended performance~\citep{ref36_Christopher2004,ref37_Pettit2019}. The four capacities are not equally well served by existing quantitative methods. A review of the resilience literature shows that most formal approaches address absorption and recovery, typically through network design, redundancy allocation, or post-disruption reconfiguration~\citep{ref38_Hosseini2019}. Anticipation, by contrast, requires detecting a disruption while it is unfolding and while corrective action is still possible, which imposes requirements that strategic methods are not designed to meet, such as: continuous computation over live operational data; latency compatible with the decision window; and outputs interpretable by the personnel who must act on them. 

This mismatch is visible in the dominant modelling paradigms. Simulation-based and digital-twin approaches reproduce disruption scenarios with high fidelity but are data- and computation-intensive, positioning them as instruments for strategic evaluation rather than continuous monitoring~\citep{ref31_Ivanov2020}. Predictive and machine learning methods have been applied extensively to supply chain risk, yet reviews of the field note a persistent concentration on external risk sources and on model accuracy rather than on operational integration~\citep{ref39_Baryannis2019,ref40_Brintrup2020}. Meanwhile, delay prediction has been studied productively in the transport domain, where estimated time of arrival models exploit network structure and real-time conditions~\citep{ref41_Balster2020,ref42_Servos2020}, but these models generally treat the shipment leg in isolation rather than as one stage of a procurement process that also encompasses supplier dispatch. The consequence is a gap at the operational layer: internal disruptions, such as dispatching delays and carrier underperformance~\citep{ref43_Stamelou2025}, are comparatively underserved, and the quantities produced by existing models are often not expressible as indicators that a logistics operator can monitor and act upon. The Horizon Europe project M4ESTRO (\url{https://www.m4estro-project.eu}), launched in 2023, addresses this layer directly. Building on the Manufacturing-as-a-Service paradigm, the project develops methods and tooling for resilient, transparent, and adaptable manufacturing systems, with an emphasis on timely preparedness and adaptive response rather than retrospective analysis. Within the project, the M4SCARA web application (M4ESTRO Supply Chain Resilience App) provides the operational environment in which the present work is deployed: it ingests heterogeneous data from carriers, traffic and weather services, and supplier dispatching times, and exposes continuously updated estimates of shipment progress to the manufacturer. The scope of the article is bounded to the design, formalization, and implementation of a selected subset of disruption indicators, with emphasis on the formalization of carrier routing behaviour and on the pragmatic choices imposed by limited data. This work builds upon the first author's MSc thesis~\citep{ref01_Volpato2025}, which provides the complete formal treatment, motivation, and extended experimental analysis underlying the proposed framework. The mathematical elements required to reproduce the approach are included in this manuscript, while the complete formal derivations and supplementary material are available in the original source.

\subsection{Literature Review}\label{literature-review}

Supply chain risk management has evolved significantly over the past decade, reflecting a shift from static planning approaches to dynamic, data-driven frameworks. This evolution has been propelled by advances in predictive analytics, artificial intelligence (AI), digital twins, and the integration of real-time data streams. Probabilistic and stochastic models have become central to managing uncertainty in supply chain operations, with particular emphasis on delay estimation, shipment and dispatch modelling, and the use of uncertainty intervals and Markov chains. Predictive analytics has emerged as one of the most widely adopted tools for disruption detection and delay estimation. Techniques such as time-series forecasting, regression models, and anomaly detection are routinely employed to anticipate risks based on both historical and streaming data~\citep{ref02_Carbonneau2008,ref06_Bodendorf2023}. Recent surveys highlight the application of supervised and unsupervised learning methods to real-time risk mitigation~\citep{ref03_Aljohani2023}, while deep learning architectures, including Graph Neural Networks~\citep{ref04_Wasi2024} and inductive graph transformers~\citep{ref05_Zhou2023}, exploit network structures to enhance delivery time estimation and supply chain optimization. Digital twins represent a transformative approach, enabling the creation of virtual replicas of supply chains that facilitate the simulation of disruption scenarios and the evaluation of mitigation strategies~\citep{ref07_Seco2014}. These virtual models allow decision-makers to assess alternative configurations before implementing real-world changes, supporting proactive disruption management. However, digital twin methods are data intensive and computationally demanding, which can limit their practicality for real-time logistics monitoring.

The integration of real-time data from heterogeneous sources, such as IoT devices, blockchain-enabled traceability systems, and social media, has further enhanced the transparency and agility of modern supply chains~\citep{ref08_Sallam2023}. Continuous monitoring of shipments and operations is now possible, enabling organizations to respond more effectively to emerging risks. However, uncertainty remains an inherent feature of logistics processes, particularly in dispatching and shipment steps. To address this aspect, stochastic and probabilistic models are commonly used to represent variability and support robust decision-making, for instance, scenario-based frameworks compare stochastic programming against robust optimization, using simple uncertainty sets constructed from limited or partial data~\citep{ref09_Maggioni2017}. These approaches allow decision-makers to hedge against worst-case outcomes without requiring full probabilistic knowledge, underscoring the importance of uncertainty-aware models, especially lightweight ones, in estimating delays and delivery performance when operational data are sparse. In this context, Markov chains have served as a foundational tool for modelling uncertainty in logistics and supply chain operations, as their ability to represent state transitions probabilistically makes them particularly suitable for modelling shipment progress, disruption propagation, and recovery behaviours. The typical applications include Hidden Markov Models for analysing supply sequences and forecasting delivery reliability~\citep{ref10_Borucka2023} dynamic Bayesian networks that incorporate Markovian transitions to model ripple effects of disruptions~\citep{ref11_Liu2022}, and discrete-time Markov chain models for quantifying supply chain resilience~\citep{ref13_Ayrapetov2025}.

\subsection{Research Gaps}\label{research-gaps}

Despite the breadth of modelling techniques available, several limitations persist when considering their applicability to internal disruptions such as dispatching delays or carrier underperformance. Many approaches are data-intensive, requiring extensive and high-quality datasets that may not be available in practice. In addition, there is a predominant focus on predicting or adjusting to external risks, with comparatively fewer studies addressing operational delays arising from internal supply chain processes~\citep{ref39_Baryannis2019}. Simulation-heavy methods demand substantial computational resources, which may be impractical for real-time monitoring~\citep{ref31_Ivanov2020}. Additionally, probabilistic models often remain abstract and are not easily translated into interpretable key performance indicators (KPIs), limiting their utility for operational decision-making~\citep{ref40_Brintrup2020}. Given these constraints, lightweight alternatives are gaining traction. Indicator-based approaches, which rely on a limited set of measurable variables, offer pragmatic solutions for real-time logistics monitoring~\citep{ref08_Sallam2023} KPI-driven models~\citep{ref07_Seco2014} demonstrate that even simplified frameworks can provide valuable insights when represented through performance metrics such as dispatch times, shipment durations, and their associated variability. Uncertainty intervals and bounded uncertainty sets~\citep{ref09_Maggioni2017} are employed to quantify variability in lead times and supply performance, enabling robust optimization in supply planning even when probabilistic distributions are difficult to estimate. Markov chains and Hidden Markov Models~\citep{ref10_Borucka2023,ref12_Putthakosa2023} offer lightweight approaches to modelling shipment processes, allowing calibration on limited historical records while still producing meaningful forecasts of delivery times and potential disruptions. From this review, two gaps emerge. First, internal disruptions, such as dispatching delays, shipment slowdowns, and carrier underperformance, remain comparatively underrepresented relative to the literature's focus on external risks. Second, there is a need for lightweight, interpretable indicators capable of functioning as real-time disruption detectors without relying on complex, data-hungry models. The current study addresses both gaps by proposing a framework that models dispatch and shipment processes through delay-focused indicators, leveraging uncertainty intervals and stochastic models to produce delivery time estimates even under conditions of limited data.

\subsection{Problem Statement}\label{problem-statement}

Order delivery times are subject to a wide range of external and internal disruptions. External disruptions derive from natural disasters, geopolitical events, and market fluctuations, while internal disruptions often derive from operational inefficiencies such as supplier delays, shipment bottlenecks, and carrier underperformance. These disruptions propagate along the supply chain, often leading to unpredictable deviations from nominal lead times. Traditional analytical methods struggle to capture this complexity, as they assume stable lead-time distributions, homogeneous datasets or well-structured event logs~\citep{ref30_Verschuur2021}. These conditions are rarely met in industrial environments, as well as the use case employed for this work. For a given manufacturer, the available datasets present various challenges: imbalance concentrated on a single site and carrier; and limited or heterogeneous data from shipment tracking. These issues complicate the underlying logistics process modelling, and the reliable prediction of delays. Given these constraints, this work addresses the key problem of developing a hierarchical modelling framework for a single-manufacturer, multiple-supplier supply chain. The framework is designed to represent the structural dependencies within the logistics process, extract delay-relevant indicators from limited operational data, and predict the likelihood of delivery delays despite data imbalance, sparsity, and noise. The industrial use case is based on the M4ESTRO project partner FAE Technology S.p.A. (\url{https://fae.technology/}), an Italian company specializing in embedded and custom electronic solutions, which serves as the pilot partner providing the core order and shipment dataset for this study. Its supply chain is inbound-focused, sourcing electronic components and materials from multiple distributed suppliers across various global sites, delivered via several carriers to its manufacturing facility. This use case is an ideal real-world test, since it presents with a multi-supplier, multi-carrier procurement structure that mirrors the general inbound logistics setting the model is designed to address.

\section{Modelling Approach}\label{modelling-approach}

The proposed approach is founded on a hybrid modelling strategy comprising two complementary components: the Supply Chain Graph Model (SCGM), in which carrier-specific routing choices are described by a Markov chain, and a hierarchical set of Logistics Chain Disruption Indicators (LCDIs) that quantitatively characterize the dynamics of the order fulfilment process along the logistic chain. The graph representation captures how shipments move through the network, while the LCDIs describe how long different components of the network take to operate. The remainder of this section first introduces the order fulfilment process, the SCGM structure, and then defines the LCDIs framework used to quantify dispatching, shipment, and delivery times.

\subsection{Order Fulfilment Process}\label{order-fulfilment-process}

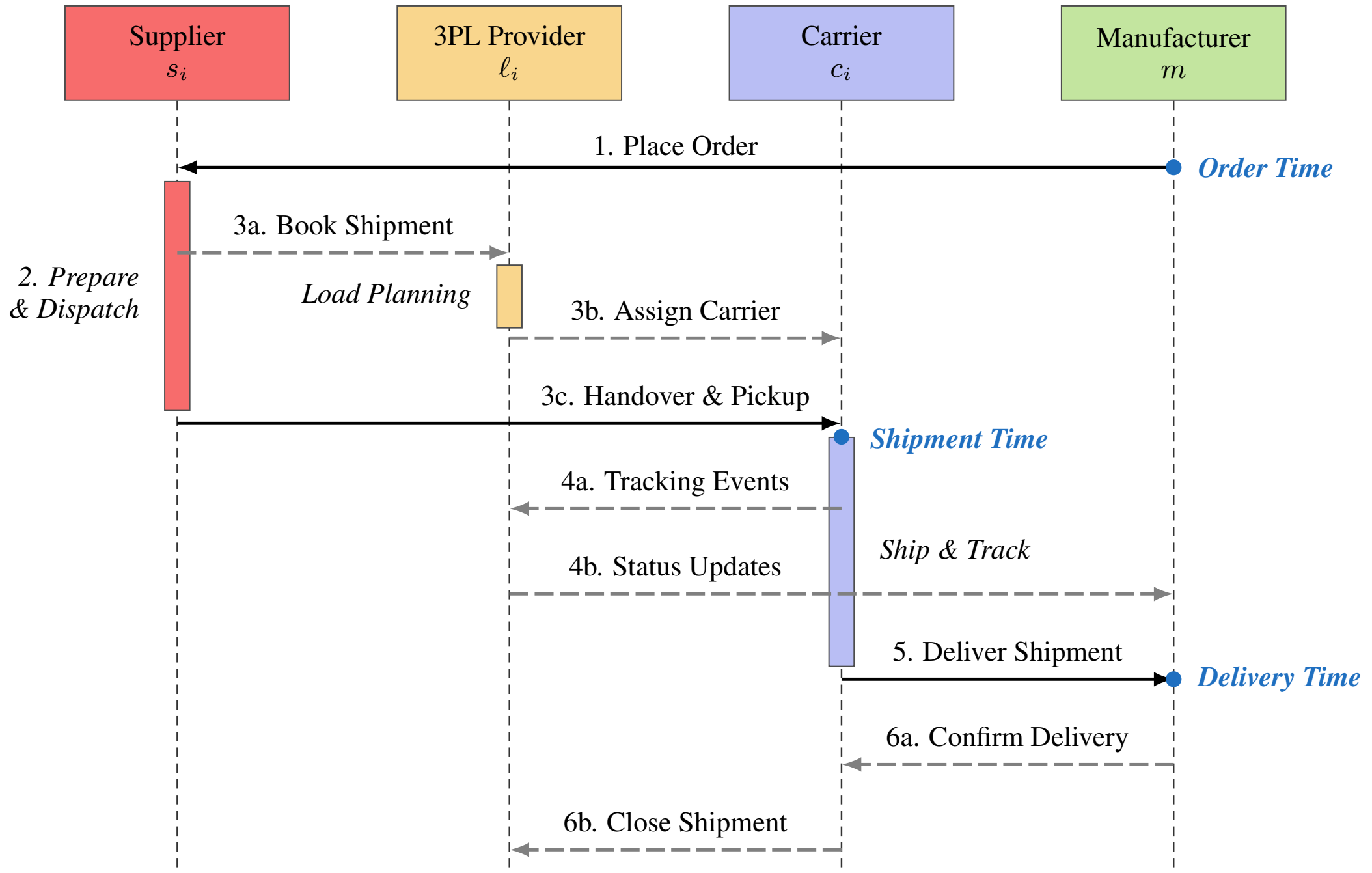
\begin{figure}[htbp]
\centering
\begin{adjustbox}{max width=\linewidth}
\begin{tikzpicture}[
  font=\small,
  actor/.style={draw=black!70, minimum width=2.5cm, minimum height=1.05cm, align=center, inner sep=2pt},
  life/.style={densely dashed, black!80, line width=0.5pt},
  msg/.style={-{Latex[length=2.2mm]}, line width=1pt},
  rmsg/.style={msg, draw=black!50, dash pattern=on 6pt off 2pt},
  act/.style={draw=black!70, line width=0.4pt, minimum width=0.28cm, inner sep=0pt},
  note/.style={font=\small\itshape},
  tstamp/.style={circle, fill=labelblue, inner sep=0pt, minimum size=5pt},
  tlabel/.style={font=\small\itshape, text=labelblue, anchor=west},
]

  \def\xS{0}\def\xL{3.7}\def\xC{7.4}\def\xM{11.1}
  \def\yEnd{-9.6}

  \node[actor, fill=supplier]     (S) at (\xS,-0.525) {Supplier\\ $s_{i}$};
  \node[actor, fill=threepl]      (L) at (\xL,-0.525) {3PL Provider\\ $\ell_{i}$};
  \node[actor, fill=intermediate] (C) at (\xC,-0.525) {Carrier\\ $c_{i}$};
  \node[actor, fill=manuf]        (M) at (\xM,-0.525) {Manufacturer\\ $m$};
  \foreach \x/\n in {\xS/S,\xL/L,\xC/C,\xM/M}
    \draw[life] (\n.south) -- (\x,\yEnd);

  \def\yA{-1.80}\def\yB{-2.75}\def\yC{-3.70}\def\yD{-4.65}\def\yE{-5.60}
  \def\yF{-6.55}\def\yG{-7.50}\def\yH{-8.45}\def\yI{-9.40}

  \node[act, fill=supplier, anchor=north, minimum height=2.55cm] at (\xS,-1.95) {};
  \node[act, fill=threepl,  anchor=north, minimum height=0.70cm] at (\xL,-2.88) {};
  \node[act, fill=intermediate, anchor=north, minimum height=2.55cm] at (\xC,-4.80) {};

  \node[note, anchor=east, align=right] at (\xS-0.3,-3.22) {2. Prepare \\\& Dispatch};
  \node[note, anchor=east] at (\xL-0.3,-3.23) {Load Planning};
  \node[note, anchor=west] at (\xC+0.3,-6.08) {Ship \& Track};

  \draw[msg]  (\xM,\yA) -- node[above] {1. Place Order}         (\xS,\yA);
  \draw[rmsg] (\xS,\yB) -- node[above] {3a. Book Shipment}      (\xL,\yB);
  \draw[rmsg] (\xL,\yC) -- node[above] {3b. Assign Carrier}     (\xC,\yC);
  \draw[msg]  (\xS,\yD) -- node[above, pos=0.75] {3c. Handover \& Pickup} (\xC,\yD);
  \draw[rmsg] (\xC,\yE) -- node[above] {4a. Tracking Events}    (\xL,\yE);
  \draw[rmsg] (\xL,\yF) -- node[above, pos=0.25] {4b. Status Updates} (\xM,\yF);
  \draw[msg]  (\xC,\yG) -- node[above] {5. Deliver Shipment}    (\xM,\yG);
  \draw[rmsg] (\xM,\yH) -- node[above] {6a. Confirm Delivery}   (\xC,\yH);
  \draw[rmsg] (\xC,\yI) -- node[above] {6b. Close Shipment}     (\xL,\yI);

  \node[tstamp] at (\xM,\yA) {};   \node[tlabel] at (\xM+0.15,\yA) {\textbf{Order Time}};
  \node[tstamp] at (\xC,-4.80) {}; \node[tlabel] at (\xC+0.2,-4.85) {\textbf{Shipment Time}};
  \node[tstamp] at (\xM,\yG) {};   \node[tlabel] at (\xM+0.15,\yG) {\textbf{Delivery Time}};
\end{tikzpicture}
\end{adjustbox}
\caption[Sequence diagram of the order fulfilment process]{Sequence diagram of the order fulfilment process. The manufacturer places an order with the supplier (1), who prepares and dispatches it (2). The supplier books the shipment with a third-party logistics (3PL) provider (3a), which plans the load and assigns a carrier (3b); the goods are then handed over to the carrier for pickup (3c). The carrier ships the goods and tracks them, reporting tracking events to the 3PL (4a), which forwards status updates to the manufacturer (4b). The carrier delivers the shipment to the manufacturer (5), who confirms delivery (6a), after which the carrier closes the shipment with the 3PL (6b). The bold blue markers indicate the order time, shipment time, and delivery time as timestamps referenced throughout the LCDI framework.}
\label{fig:1}
\end{figure}

Supply chain logistics involves the flow of materials and goods between geographically distributed facilities, coordinated through a sequence of procurement, transportation, and delivery operations. In this setting, a company hereafter referred to as the \emph{manufacturer}, has a need to acquire materials from a \emph{supplier} site. Thus, the manufacturer issues an \emph{order}, which is defined as the set of operations it instantiates and directs to a supplier. Each order comprises a \emph{shipment}, defined as the subset of operations exclusively related to the transportation of goods by a specific \emph{carrier} from the supplier site to the manufacturer. The order fulfilment process thus encompasses procurement activities, such as order placement and supplier dispatch from a Third Party Logistics (TPL) Provider, as well as transportation carried out by the selected carrier. Figure~\ref{fig:1} illustrates this process, together with the key timestamps used throughout the paper to define the LCDIs.

\subsection{Supply Chain Graph Model}\label{supply-chain-graph-model}

\begin{figure}[htbp]
\centering
\begin{tikzpicture}[x=1cm,y=-1cm,scale=0.95,
  >={Stealth[length=2.2mm]},
  sup/.style={circle,draw=black!70,fill=supplier,minimum size=7mm,inner sep=0pt},
  int/.style={circle,draw=black!70,fill=intermediate,minimum size=7mm,inner sep=0pt},
  man/.style={circle,draw=black!70,fill=manuf,minimum size=11mm,inner sep=0pt},
  e/.style={->,black,thin},
  pe/.style={->,pathblue,line width=1.2pt}]
  \node[sup] (s2) at (5.0,0.3) {$s_2$};
  \node[sup] (s1) at (1.9,1.4) {$s_1$};
  \node[sup] (s3) at (8.1,1.4) {$s_3$};
  \node[int] (i1) at (3.1,2.4) {$i_1$};
  \node[int] (i2) at (5.0,1.9) {$i_2$};
  \node[int] (i3) at (6.8,2.4) {$i_3$};
  \node[int] (i4) at (3.5,3.9) {$i_4$};
  \node[int] (i5) at (5.0,3.6) {$i_5$};
  \node[int] (i6) at (6.5,3.9) {$i_6$};
  \node[man] (m)  at (5.0,5.3) {$m$};
  \draw[e] (s1) -- (i2);  \draw[e] (s2) -- (i2);  \draw[e] (s2) -- (i3);
  \draw[e] (s3) -- (i3);  \draw[e] (i2) -- (i4);  \draw[e] (i2) -- (i5);
  \draw[e] (i2) -- (i6);  \draw[e] (i3) -- (i5);  \draw[e] (i3) -- (i6);
  \draw[e] (i5) -- (m);   \draw[e] (i6) -- (m);
  \draw[->,red,line width=1.4pt] (s1) -- (i1);
  \draw[pe] (s2) -- (i1); \draw[pe] (i1) -- (i4); \draw[pe] (i4) -- (m);
  \node[anchor=west] at (8.5,1.4) {Supplier Site};
  \node[anchor=west] at (5.6,5.3) {Manufacturer};
  \draw[decorate,decoration={brace,amplitude=5pt}] (7.6,2.2) -- (7.6,4.1)
     node[midway,right=6pt] {Intermediate Sites};
  \node[red,align=center] (rl) at (0.9,2.4) {Route\\[-2pt] \textcolor{black}{$r$}};
  \draw[red!60,thin] (1.45,2.25) -- ($(s1)!0.55!(i1)$);
  \node[cyan!70!blue,align=center] (pl) at (1.5,3.7) {Path\\[-2pt] \textcolor{black}{$\pi$}};
  \draw[cyan!60,thin] (2.0,3.55) -- ($(i1)!0.5!(i4)$);
\end{tikzpicture}
\caption[Supply Chain Graph representation]{Supply Chain Graph representation. Supplier sites constitute the source vertices and are shown in red, intermediate logistics nodes are shown in light blue, and the manufacturer acts as the sink vertex, shown in green. A route \(r\) is a single edge connecting two vertices, and a path \(\pi\) is any walk from a source to the sink vertex. The graph shown is a simplified representation with \(|S| = 3\) supplier sites, \(|I| = 6\) intermediate nodes, and a single manufacturer, for a total of \(|V| = 10\) vertices. A route is exemplified in red \(r = \left( s_{1},i_{1} \right)\), and a path is exemplified in blue \(\pi = \left( s_{2},i_{1},i_{4},m \right)\).}
\label{fig:2}
\end{figure}
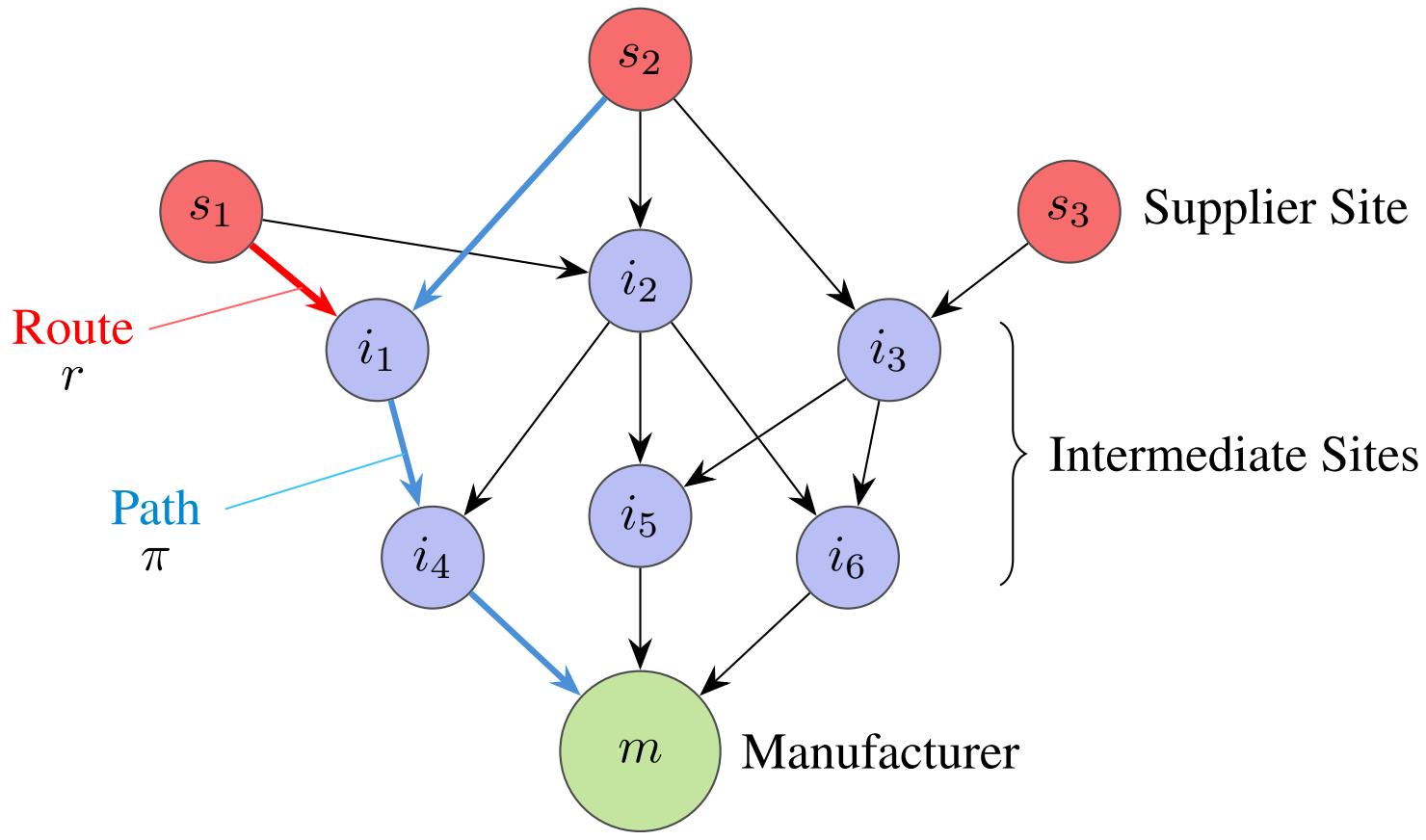

The basis of the SCGM structure is a Directed Acyclic Graph (DAG), defined as \(G = (V,A,C)\), where vertices represent entities or physical locations along the logistic chain, and directed edges represent transportation routes. The vertex set is defined as a partition \(V = S \cup I \cup \left\{ m \right\}\), where each subset identifies a specific group of entities. Suppliers (\(s_{i} \in S\)) represent the origin of shipments, each corresponding to a specific site of a supplier company. Intermediate vertices (\(i_{i} \in I\)) represent physical locations inferred from tracking events reported by logistics carriers during transportation, such as terminals, sorting centres, or processing facilities. The manufacturer vertex (\(m\)) is the sink node, the destination of all shipments related to the company issuing orders. The arcs set is a partition \(A = A_{SI} \cup A_{II} \cup A_{IM}\), where each subset identifies a specific connection between vertices. The set \(A_{SI}\) denotes arcs from supplier sites to intermediate vertices, the set \(A_{II}\) denotes arcs between intermediate vertices, while the set \(A_{IM}\) denotes arcs from intermediate vertices to the manufacturer. The graph model depends on a set of carriers \(C = \left\{ c_{1},c_{2},\ldots,c_{k} \right\}\), denoting the possible shipping companies transporting goods from a supplier to the manufacturer. The SCGM allows the definition of logistics descriptors, giving abstract graph components an operational meaning consistent with real-world shipment terminology. Specifically, a \emph{route} \(r_{i} = \left( v_{a},v_{b} \right) \in A\) is defined as a directed edge connecting two vertices, representing a single logistic step within the shipment process. A \emph{path} \(\pi_{i} = \left( v_{a},v_{b},\ \ldots,v_{k} \right) \in A\) is defined as an ordered sequence of connected vertices, representing the flux of goods between the first and last vertices, which is also equivalent to a route sequence \(\rho_{i} = \left( r_{1},r_{2},\ldots,r_{k - 1} \right)\). Acyclicity is introduced as a modelling assumption, motivated by the fact that no cyclic shipment trajectories were observed in the dataset. This simplifies path enumeration and probabilistic routing analysis, guaranteeing that every shipment corresponds to a unique maximal directed path ending at the sink vertex, where proof is provided by Theorem A.1 in Appendix~\ref{appendix-A}. Acyclicity holds whether the shipment is newly dispatched from a supplier site or it is already in transit from an intermediate vertex, allowing both cases to be handled with the same routing logic. Figure~\ref{fig:2} provides a visual summary of the SCGM representation described above.

\subsubsection{Stochastic Routing Model}\label{stochastic-routing-model}

A deterministic representation of routing is insufficient to capture how shipments traverse the network, since different carriers may follow different routes from the same origin, and any specific path taken by individual shipments cannot be known in advance. To account for this variability, carrier routing behaviour is modelled statistically, by estimating transition frequencies from historical tracking event sequences, treating each shipment's trajectory as a realization of a discrete-time stochastic process \(\left\{ X_{t} \right\}_{t\mathbb{\in N}}\) exhibiting memoryless behaviour. This is possible because routing decisions at each vertex depend only on the current location and not on the shipment's past trajectory. Defining, for each directed edge \(r_{i} = \left( v_{a},v_{b} \right)\), the number of shipments that have traversed that edge as the flow value \(f\left( v_{a},v_{b} \right)\), it is possible to prove flow conservation at every intermediate vertex~\citep{ref15_Ford1956}, as this property propagates throughout the network, implying that all shipment flows originate at source vertices and terminate at the sink (see Theorem A.2 in Appendix~\ref{appendix-A}). Building on this, routing is inferred empirically rather than assumed from a theoretical distribution: for each vertex \(v\), outgoing transition probabilities to any other vertex \(u\), namely \(P_{uv}= \mathbb{P}\left( X_{t + 1} = u\  \middle| \ X_{t} = v \right)\ \)are estimated from existing shipment data by normalizing the empirical flows \(f(v,u)\) by the total outgoing flow of the vertex \(f_{v}(v)\). These normalized flows define valid probability distributions (see Theorem A.3 in Appendix~\ref{appendix-A}) and fully characterize \(\left\{ X_{t} \right\}_{t\mathbb{\in N}}\) as a discrete-time Markov process~\citep{ref14_Norris1997}, consistent with the memoryless property whereby routing decisions depend only on the current vertex. The full derivation of flow functions, transition probabilities, and their extension to path-level probabilities and distributions is provided in the first author's MSc thesis~\citep{ref01_Volpato2025}. Given the Markovian property of the shipment process and the acyclic structure of the graph, a set of feasible paths from any vertex to the manufacturer vertex can be extracted using a modified depth-first search procedure~\citep{ref16_Tarjan1972}, with each path assigned a probability equal to the product of its transition probabilities. The authors recognize that although the number of possible paths may grow exponentially in the worst case, to the best of their knowledge, routing variability is limited in practice. Moreover, the path space can be truncated by discarding paths with a probability below a given threshold and renormalizing the remaining probabilities, ensuring computational feasibility without reducing modelling fidelity.

\subsubsection{Carrier-conditioned Routing Model}\label{carrier-conditioned-routing-model}

Transition probabilities may be conditioned on any carrier\(\ c\) to account for heterogeneous routing, since different carriers were empirically observed to follow different routes from the same vertex. A carrier-agnostic probability merges distinct routing behaviours rather than capturing each carrier's actual delivery. The procedure is analogous to the construction of the transition probabilities defined above. For each carrier \(c\), a carrier-specific flow function \(f^{c}\left( v_{a},v_{b} \right)\) is defined by restricting the count on a directed edge \(r_{i} = \left( v_{a},v_{b} \right)\), to the shipments transported by \(c\), and the corresponding outgoing transition probability \(P_{uv}^{c}= \mathbb{P}\left( X_{t + 1} = u\ |\ X_{t} = v\  \land \ C = c \right)\) from each vertex \(v\) to any other vertex \(u\) are obtained by normalizing this carrier-specific flows \(f^{c}(v,u)\) by the total outgoing carrier-specific flow of the vertex \(f_{v}^{c}(v)\). Where no shipments were observed to depart from a vertex under a given carrier, a degenerate self-loop was introduced to preserve the row-stochasticity of the transition matrix. This resulted in a family of carrier-specific Markov chains defined on the same graph structure, where differences among carriers are captured exclusively through their transition probability matrices. This formulation enables a more fine-grained analysis of shipment dynamics across carriers, accounting for heterogeneous behaviour in network traversal and routing strategy, and extends naturally to carrier-conditioned path probabilities and path distributions, obtained as products and normalized vectors of the corresponding carrier-specific transition probabilities. The full derivation of the carrier-conditioned flow functions, transition probabilities, and their extension to path-level distributions is provided in the first author's MSc thesis~\citep{ref01_Volpato2025}.

\subsection{Logistic Chain Disruption Indicators}\label{logistic-chain-disruption-indicators}

Building on the SCGM, a framework of indicators is introduced to quantify disruptions along the order fulfilment process, from an industrial supplier to the manufacturer. The proposed framework is designed to be modular, hierarchical, and independent of estimation techniques, allowing it to accommodate heterogeneous data sources and evolving modelling choices. In this context, an \emph{indicator} is defined as a measurable or estimated quantity associated with a specific component of the supply chain graph representation (such as a vertex, a route, or a path), formalized as a mapping from observed data and graph-derived structures to a quantitative variable. The value of each indicator represents either a time distribution, a pointwise estimate of a physical property of the logistics chain, or an interval-based bound, depending on the specific indicator and the level of uncertainty it is designed to capture. Within this framework, a distinction is drawn between \emph{historical} LCDIs and \emph{real-time} LCDIs, which serve complementary but separated roles; a complete tabular summary of both families, reporting the scope, type, and description of each indicator, is provided in Appendix~\ref{appendix-C}. The following sections detail this indicator framework: the first defines the historical LCDIs, while the second introduces the real-time LCDIs, presenting their formal definition and its corresponding implementation across increasing levels of granularity, that is from vertex-level, route-level, to path-level and at graph-level, reflecting the structural decomposition of the SCGM.

\subsubsection{Historical LCDIs}\label{historical-lcdis}

\begin{figure}[htbp]
\centering
\begin{tikzpicture}[x=1cm,y=1cm,scale=1]
  \draw[->,thick] (-0.4,0) -- (11.5,0) node[right] {$t$};
  \fill[supplier] (0,-0.22) rectangle (2.6,0.22); \draw (0,-0.22) rectangle (2.6,0.22);
  \fill[intermediate] (2.6,-0.22) rectangle (7.7,0.22); \draw (2.6,-0.22) rectangle (7.7,0.22);
  \node at (1.3,-0.02) {Dispatch}; \node at (5.15,-0.02) {Shipment};
  \foreach \x in {0,2.6,7.7} \draw[thick] (\x,-0.3) -- (\x,0.3);
  \draw[decorate,decoration={brace,amplitude=8pt}] (0,0.5) -- (7.7,0.5) node[midway,above=9pt] {Elapsed};
  \node[below] at (0,-0.35) {$t_0$};
  \node[below] at (2.6,-0.35) {$t_1$};
  \node[below] at (7.7,-0.35) {$t_n$};
  \node[labelblue,align=center,font=\itshape] at (0,-1.35) {Order\\Time};
  \node[labelblue,align=center,font=\itshape] at (2.6,-1.35) {Shipment\\Time};
  \node[labelblue,align=center,font=\itshape] at (7.7,-1.35) {Delivery\\Time};
\end{tikzpicture}
\caption[Time sequence of the order fulfilment process]{Time sequence of the order fulfilment process. As introduced in Figure~\ref{fig:1}, the order fulfilment process comprises the time between order placement and delivery. The total time is decomposed into two components: the \emph{dispatch interval}, from the order time \(t_{0}(s)\) to the shipment time \(t_{1}(s,c)\), and the \emph{shipment interval}, from \(t_{1}(s,c)\) to the delivery time \(t_{n}(s,c)\). Since these times are generated by individual orders, they are inherently order- and carrier-dependent, and are to be understood as such throughout, even where the dependence is omitted from the notation for brevity.}
\label{fig:3}
\end{figure}
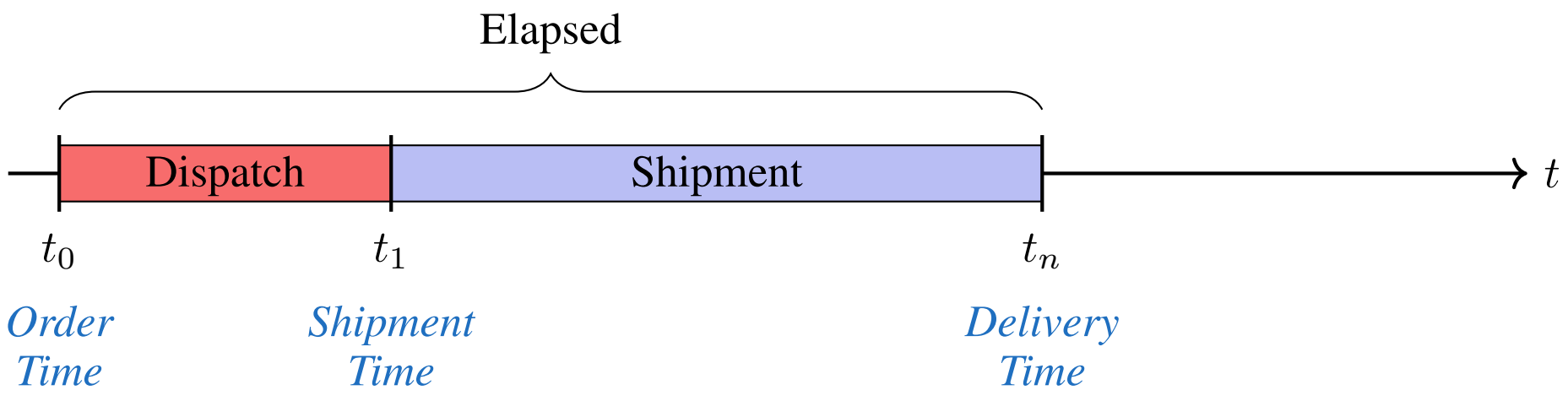

These indicators are statistical descriptors estimated from completed shipments. They characterize the operational behaviour of the logistics chain based on past orders, and provide probabilistic baselines for a given supplier-carrier pair used in the calculation of real-time LCDIs. Once estimated, they remain fixed until new data becomes available, at which point they are recomputed. They are not influenced by the state of any individual ongoing shipment, and since they are estimated from completed shipments, they implicitly capture the statistical pattern of past disruptions, such as recurring dispatch delays or transit variability, reflected in the shape of their underlying distributions. What they cannot capture is the disruption state of any individual ongoing shipment, which instead requires the real-time LCDIs introduced in the next section. Figure~\ref{fig:3} illustrates the time sequence underlying the order fulfilment process, decomposing the times between order placement and delivery, as it provides the calculation basis for the historical LCDIs. The remainder of this section defines the historical LCDIs in detail. Their summary is provided in Table~\ref{tab:C1}.

\paragraph{Dispatch Time (DT)}\label{dispatch-time-dt}

The Dispatch Time depends on the statistical distribution of dispatching times, that is the elapsed time between the shipment time \(t_{1}(s,c)\) and the order time \(t_{0}(s)\), obtained by historical data. Since a single supplier may operate multiple sites across different geographic locations, each site is assigned its own dispatch time distribution. Therefore, two forms exist for this indicator, a pointwise value (scalar), and an interval-based value (2D vector). The first one is formally the expected value of the distribution, that is
\begin{equation}
\text{DT}_{\mu}(s) = \mathbb{E}\left\lbrack \text{DT}(s) \right\rbrack
\label{eq:1}\,.
\end{equation}
While the second one is the two-sided confidence interval
\begin{equation}
\text{DT}_{\beta}(s) = \text{CI}_{\beta}\left\lbrack \text{DT}(s) \right\rbrack = \left\lbrack L_{\beta},U_{\beta} \right\rbrack
\label{eq:2}
\end{equation}
at an arbitrary confidence level \(\beta\) such that \(\mathbb{P}\left( L_{\beta} \leq X \leq U_{\beta} \right) \geq \beta\) where \(X \sim \text{DT}(s)\). This form quantifies uncertainty in an explicit way, making it more appropriate for robust planning under variable dispatch conditions.

\paragraph{Shipment Time (ST)}\label{shipment-time-st}

The Shipment Time depends on the statistical distribution of shipment times, that is the elapsed time between the delivery time \(t_{n}(s,c)\) and shipment time \(t_{1}(s,c)\), obtained by historical data. Unlike the Dispatch Time indicator, here there exists a joint dependence on the site and the carrier responsible for transportation. Since a single supplier may operate multiple sites across different geographic locations, each site is assigned its own shipment time distribution. Therefore, two forms exist for this indicator. The first one is formally the expected value of the distribution, that is
\begin{equation}
\text{ST}_{\mu}(s,c) = \mathbb{E}\left\lbrack \text{ST}(s,c) \right\rbrack
\label{eq:3}\,.
\end{equation}
While the second one is the two-sided confidence interval
\begin{equation}
\text{ST}_{\beta}(s,c) = \text{CI}_{\beta}\left\lbrack \text{ST}(s,c) \right\rbrack = \left\lbrack L_{\beta},U_{\beta} \right\rbrack
\label{eq:4}
\end{equation}
at an arbitrary confidence level \(\beta\) such that \(\mathbb{P}\left( L_{\beta} \leq X \leq U_{\beta} \right) \geq \beta\) where \(X \sim \text{ST}(s,c)\), the same method for \(\text{DT}_{\beta}(s)\).

\paragraph{Delivery Time (DLT)}\label{delivery-time-dlt}

The Delivery Time is defined as the total elapsed time between order placement and delivery, obtained as the direct sum of the Dispatch Time and Shipment Time indicators. This summation relies on the assumption that, for any given supplier site and carrier pair, the two indicators are independent, which in turn makes it possible to express
\begin{align}
\text{DLT}_{\mu}(s,c) &= \text{DT}_{\mu}(s) + \text{ST}_{\mu}(s,c)
\label{eq:5}\\[2pt]
\text{DLT}_{\beta}(s,c) &= \text{DT}_{\beta}(s) + \text{ST}_{\beta}(s,c)
\label{eq:6}\,.
\end{align}
\subsubsection{Real-time Vertex-Level LCDIs}\label{real-time-vertex-level-lcdis}

\begin{figure}[htbp]
\centering
\begin{tikzpicture}[x=1cm,y=-1cm,scale=0.95,
  >={Stealth[length=2.2mm]},
  sup/.style={circle,draw=black!70,fill=supplier,minimum size=7mm,inner sep=0pt},
  int/.style={circle,draw=black!70,fill=intermediate,minimum size=7mm,inner sep=0pt},
  man/.style={circle,draw=black!70,fill=manuf,minimum size=11mm,inner sep=0pt},
  faded/.style={draw opacity=0.2,fill opacity=0.25,text opacity=0.3},
  half/.style={draw opacity=0.6,fill opacity=0.7},
  dim/.style={->,black!22,thin},
  hot/.style={->,black,line width=1.1pt},
  ann/.style={decorate,decoration={brace,amplitude=5pt},bracecol,thin},
  lead/.style={bracecol,thin}]
  \node[sup,faded] (s2) at (5.0,0.3) {$s_2$};
  \node[sup] (s1) at (1.9,1.4) {$s_1$};
  \node[sup,faded] (s3) at (8.1,1.4) {$s_3$};
  \node[int,faded] (i1) at (3.1,2.4) {$i_1$};
  \node[int,faded] (i2) at (5.0,1.9) {$i_2$};
  \node[int,faded] (i3) at (6.8,2.4) {$i_3$};
  \node[int] (i4) at (3.5,3.9) {$i_4$};
  \node[int,faded] (i5) at (5.0,3.6) {$i_5$};
  \node[int,faded] (i6) at (6.5,3.9) {$i_6$};
  \node[man,faded] (m) at (5.0,5.3) {$m$};
  \draw[dim] (s1) -- (i1);
  \draw[dim] (s1) -- (i2);
  \draw[dim] (s2) -- (i1);
  \draw[dim] (s2) -- (i2);
  \draw[dim] (s2) -- (i3);
  \draw[dim] (s3) -- (i3);
  \draw[dim] (i1) -- (i4);
  \draw[dim] (i2) -- (i4);
  \draw[dim] (i2) -- (i5);
  \draw[dim] (i2) -- (i6);
  \draw[dim] (i3) -- (i5);
  \draw[dim] (i3) -- (i6);
  \draw[dim] (i4) -- (m);
  \draw[dim] (i5) -- (m);
  \draw[dim] (i6) -- (m);
  \node[anchor=west] at (8.5,1.4) {Supplier Site};
  \node[anchor=west] at (5.6,5.3) {Manufacturer};
  \draw[decorate,decoration={brace,amplitude=5pt}] (7.6,2.2) -- (7.6,4.1) node[midway,right=6pt] {Intermediate Sites};
  \node[anchor=east] (edt) at (0.6,1.4) {EDT};
  \draw[lead] (edt.east) -- (s1.west);
  \node[anchor=east] (ori) at (1.9,3.9) {ORI};
  \draw[lead] (ori.east) -- (i4.west);
  \node[anchor=east,align=right] at (-0.15,3.9) {$\text{VT}_{\mu} = \mathbb{E}(\text{ORI})$\\[2pt]$\text{VT}_{\beta} = \mathrm{CI}_{\beta}(\text{ORI})$};
  \draw[ann] (-0.05,3.25) -- (-0.05,4.55);
  \draw[lead] (0.12,3.9) -- (ori.west);
\end{tikzpicture}
\caption[Vertex-level indicators on the Supply Chain Graph]{Vertex-level indicators on the Supply Chain Graph. Building on the SCGM representation, the vertex-level indicators are computed at supplier sites and intermediate nodes. For a supplier site \(s = s_{1}\), the Estimated Dispatch Time (EDT) is shown, whose computation differs depending on dispatching and shipping stages. For a vertex \(v = i_{4}\), the Overall Residence Index (ORI) distribution and the statistically derived Vertex Time (VT) indicator are shown.}
\label{fig:4}
\end{figure}
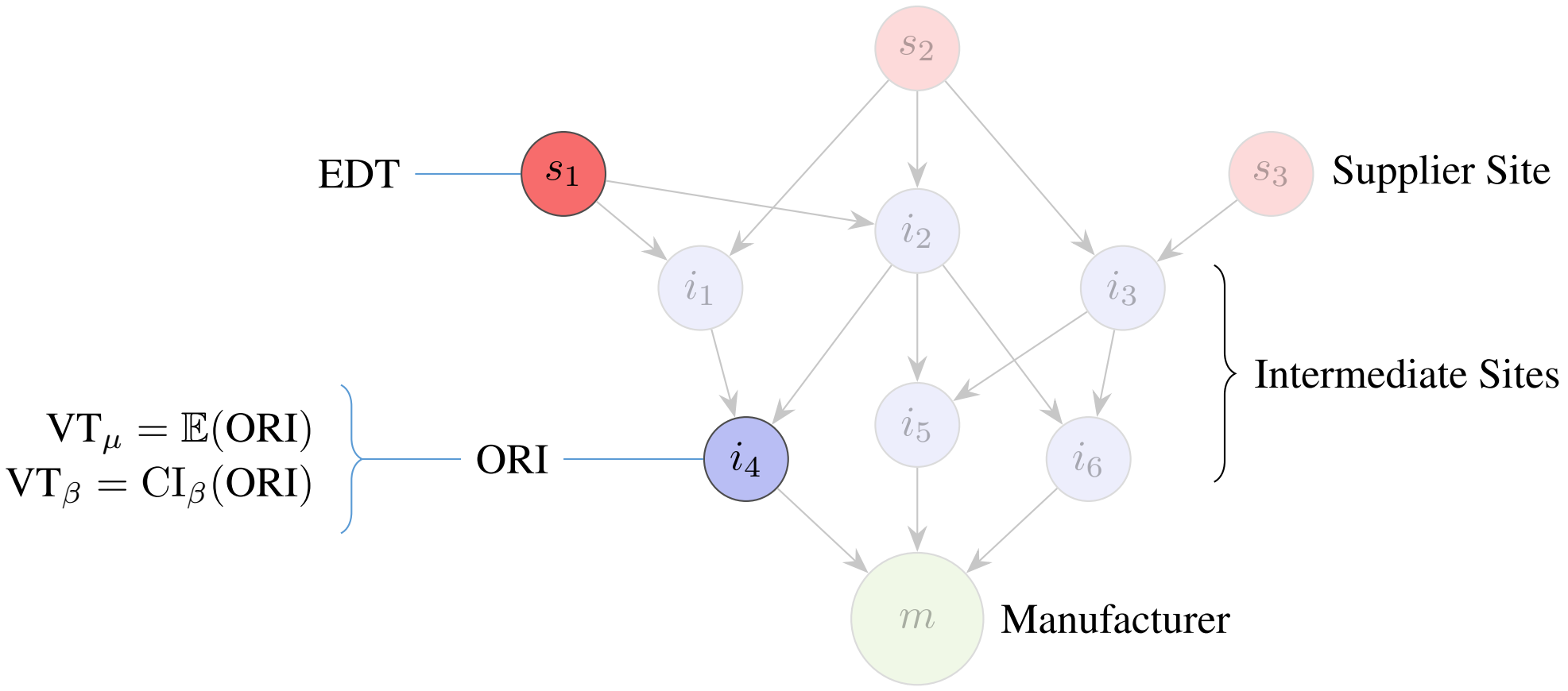

These indicators are local descriptors that characterize the time required for processing at physical facilities, such as supplier sites or intermediate vertices. In addition, they account for localized operational constraints on supplier sites, such as national and public holidays. They cannot capture the logistics performance along the possible downstream paths, which requires route-level and path-level LCDIs introduced in subsequent sections. Figure~\ref{fig:4} illustrates the vertex-level LCDIs within the SCGM representation. The remainder of this section defines each vertex-level LCDIs in detail, and their summary is provided in Table~\ref{tab:C1}.

\paragraph{Estimated Dispatch Time (EDT)}\label{estimated-dispatch-time-edt}

The Estimated Dispatch Time depends on the dispatching and shipping stages with respect to the actual order processing time \(t\) for a supplier site, that is
\begin{equation}
\text{EDT}\left( s,t_{0},t \right) = \left\{\begin{aligned}
t_{1} - t_{0}\ ,\ \ t \geq t_{1}\ ,\ \  & \text{shipping stage} \\
\max\left( t - t_{0}\ ,\ \text{DT}_{\mu}(s) \right),\ \ t < t_{1}\ ,\ \  & \text{dispatching stage}
\end{aligned}\right.
\label{eq:7}\,.
\end{equation}
At the dispatching stage, the shipment time \(t_{1}\) is unknown, thus a historical estimate is provided to account for non-operational days on the supplier's site's country. If any delay pushes the dispatching time \(\left( t - t_{0} \right)\) beyond the \(\text{DT}_{\mu}(s)\) estimate, the value is adjusted as the maximum between the two times, ensuring that the result accurately reflects both standard processing behaviour and periods of suspended operations. This indicator is also adjusted to account for supplier holidays, ensuring that facility closures are excluded from processing times. The algorithm used to determine holidays in the country of the supplier site is included in the first author's MSc thesis~\citep{ref01_Volpato2025}.

\paragraph{Overall Residence Index (ORI)}\label{overall-residence-index-ori}

The Overall Residence Index is derived from the statistical distribution of residence times at intermediate sites, each computed as the elapsed time between the first and last consecutive carrier-provided tracking events recorded at the same location, and this it is independent of any external estimation model. This indicator aggregates processing times across all carriers without distinction, since delays at a given location are generally attributable to the facility itself rather than to the carrier, and can therefore affect any carrier interacting with it. If a location appears only once in a shipment's tracking sequence, no observation is generated for that vertex.

\paragraph{Vertex Time (VT)}\label{vertex-time-vt}

The Vertex Time depends on the residence time distribution, as each intermediate vertex is assigned its ORI. Therefore, two forms exist for this indicator. The first one is formally the expected value of the distribution, that is
\begin{equation}
\text{VT}_{\mu}(v) = \mathbb{E}\left\lbrack \text{ORI}(v) \right\rbrack
\label{eq:8}\,.
\end{equation}
While the second one is the two-sided confidence interval
\begin{equation}
\text{VT}_{\beta}(v) = \text{CI}_{\beta}\left\lbrack \text{ORI}(v) \right\rbrack = \left\lbrack L_{\beta},U_{\beta} \right\rbrack
\label{eq:9}
\end{equation}
at an arbitrary confidence level \(\beta\) such that \(\mathbb{P}\left( L_{\beta} \leq X \leq U_{\beta} \right) \geq \beta\) where \(X \sim \text{ORI}(v)\).

\subsubsection{Real-time Route-Level LCDIs}\label{real-time-route-level-lcdis}

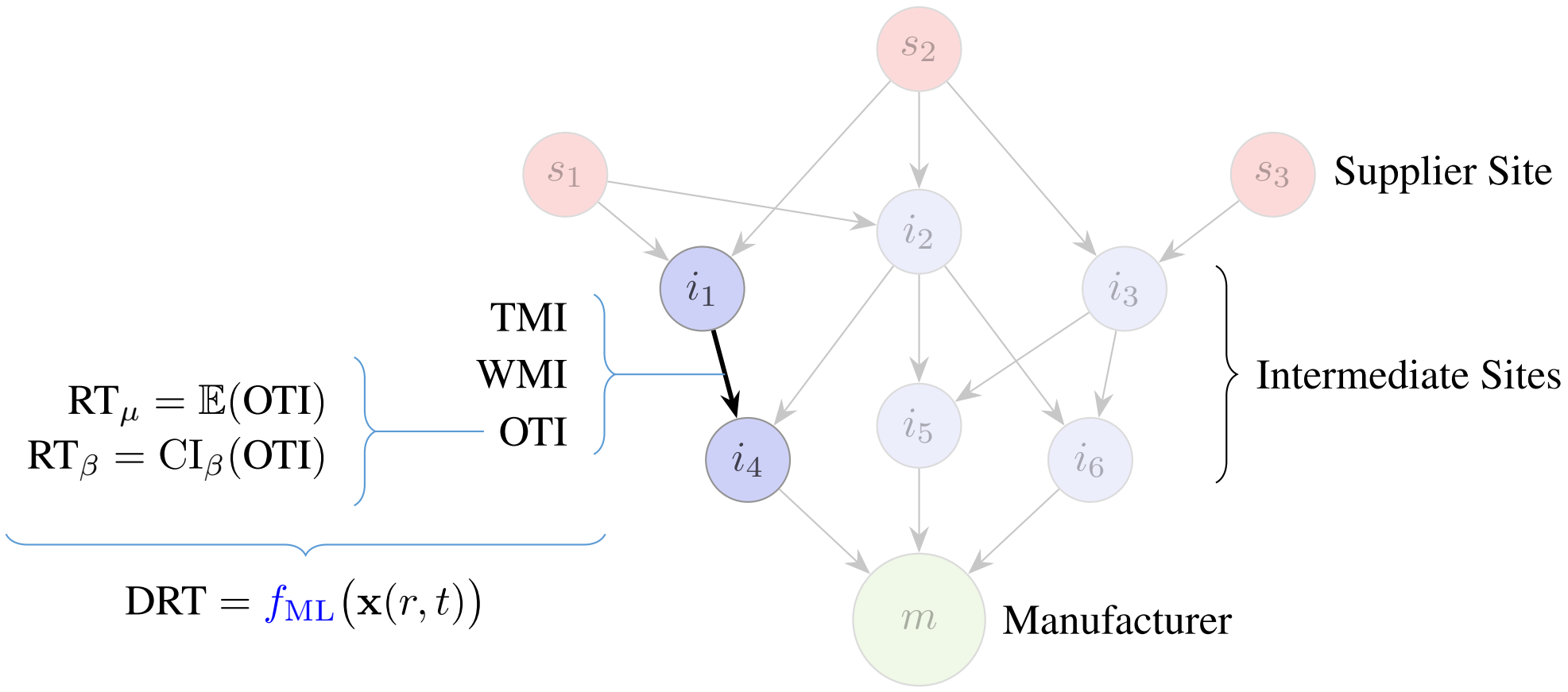
\begin{figure}[htbp]
\centering
\begin{tikzpicture}[x=1cm,y=-1cm,scale=0.95,
  >={Stealth[length=2.2mm]},
  sup/.style={circle,draw=black!70,fill=supplier,minimum size=7mm,inner sep=0pt},
  int/.style={circle,draw=black!70,fill=intermediate,minimum size=7mm,inner sep=0pt},
  man/.style={circle,draw=black!70,fill=manuf,minimum size=11mm,inner sep=0pt},
  faded/.style={draw opacity=0.2,fill opacity=0.25,text opacity=0.3},
  half/.style={draw opacity=0.6,fill opacity=0.7},
  dim/.style={->,black!22,thin},
  hot/.style={->,black,line width=1.1pt},
  ann/.style={decorate,decoration={brace,amplitude=5pt},bracecol,thin},
  lead/.style={bracecol,thin}]
  \node[sup,faded] (s2) at (5.0,0.3) {$s_2$};
  \node[sup,faded] (s1) at (1.9,1.4) {$s_1$};
  \node[sup,faded] (s3) at (8.1,1.4) {$s_3$};
  \node[int,half] (i1) at (3.1,2.4) {$i_1$};
  \node[int,faded] (i2) at (5.0,1.9) {$i_2$};
  \node[int,faded] (i3) at (6.8,2.4) {$i_3$};
  \node[int,half] (i4) at (3.5,3.9) {$i_4$};
  \node[int,faded] (i5) at (5.0,3.6) {$i_5$};
  \node[int,faded] (i6) at (6.5,3.9) {$i_6$};
  \node[man,faded] (m) at (5.0,5.3) {$m$};
  \draw[dim] (s1) -- (i1);
  \draw[dim] (s1) -- (i2);
  \draw[dim] (s2) -- (i1);
  \draw[dim] (s2) -- (i2);
  \draw[dim] (s2) -- (i3);
  \draw[dim] (s3) -- (i3);
  \draw[hot] (i1) -- (i4);
  \draw[dim] (i2) -- (i4);
  \draw[dim] (i2) -- (i5);
  \draw[dim] (i2) -- (i6);
  \draw[dim] (i3) -- (i5);
  \draw[dim] (i3) -- (i6);
  \draw[dim] (i4) -- (m);
  \draw[dim] (i5) -- (m);
  \draw[dim] (i6) -- (m);
  \node[anchor=west] at (8.5,1.4) {Supplier Site};
  \node[anchor=west] at (5.6,5.3) {Manufacturer};
  \draw[decorate,decoration={brace,amplitude=5pt}] (7.6,2.2) -- (7.6,4.1) node[midway,right=6pt] {Intermediate Sites};
  \node[anchor=east] (tmi) at (2.05,2.65) {TMI};
  \node[anchor=east] (wmi) at (2.05,3.15) {WMI};
  \node[anchor=east] (oti) at (2.05,3.65) {OTI};
  \draw[ann] (2.15,2.45) -- (2.15,3.85);
  \draw[lead] (2.33,3.15) -- (3.3,3.15);
  \node[anchor=east,align=right] at (-0.05,3.65) {$\text{RT}_{\mu} = \mathbb{E}(\text{OTI})$\\[2pt]$\text{RT}_{\beta} = \mathrm{CI}_{\beta}(\text{OTI})$};
  \draw[ann] (0.05,3.0) -- (0.05,4.3);
  \draw[lead] (0.23,3.65) -- (oti.west |- 0,3.65);
  \draw[ann] (2.25,4.55) -- (-3.0,4.55) node[midway,below=7pt,black] {$\text{DRT} = \textcolor{blue}{f_{\mathrm{ML}}}\bigl(\mathbf{x}(r,t)\bigr)$};
\end{tikzpicture}
\caption[Route-level indicators on the Supply Chain Graph]{Route-level indicators on the Supply Chain Graph. Building on the SCGM representation, the route-level indicators are computed along a single route between two vertices. The Traffic Meta Index (TMI), Weather Meta Index (WMI), and Overall Transit Index (OTI) are shown, together with the statistically derived Route Time (RT), and the Dynamic Route Time (DRT), obtained via a machine learning regression model on the feature vector \(\mathbf{x}(r,t)\).}
\label{fig:5}
\end{figure}

These indicators are local descriptors that characterize the transit conditions along a single route between two vertices, integrating real-time traffic and weather information with historical travel time data. They cannot capture the performance of a shipment across multiple consecutive routes, which requires the path-level LCDIs introduced in the next section. Figure~\ref{fig:5} illustrates the route-level LCDIs within the SCGM representation. The remainder of this section defines each route-level LCDI in detail, with a summary provided in Table~\ref{tab:C1}.

\paragraph{Overall Transit Index (OTI)}\label{overall-transit-index-oti}

The Overall Transit Index is derived from the statistical distribution of traversing times between two intermediate sites, each computed as the elapsed time between the first and last consecutive carrier-provided tracking events recorded at the different locations. This indicator is independent of any external estimation model, and it is consistent with the ORI, as it aggregates processing times across all carriers without distinction.

\paragraph{Traffic Meta-Index (TMI) and Weather Meta-Index (WMI)}\label{traffic-meta-index-tmi-and-weather-meta-index-wmi}

Traffic and weather conditions along a route are quantified through two indicators: the Traffic Meta Index (TMI) and the Weather Meta Index (WMI). They are defined as scalar functions that map a route and carrier-provided tracking time, mapping a normalized score in \(\lbrack 0,1\rbrack\), where the lower bound signifies minimum to mild impact, and the upper bound signifies maximum to severe impact. Both indicators are computed from aggregated data provided by external services, respectively traffic and weather sources, whose specific providers are detailed later in this article, alongside the other external data sources integrated into the LCDIs framework. The calculation basis of the TMI includes a heuristic classification of the travel mode (air, rail, road, or sea) from the geodesic distance and average speed of each shipment, by empirically derived distance-velocity thresholds. The TMI is then computed only for road transport, as the ratio between the estimated travel time under real traffic conditions and under ideal traffic conditions. The calculation basis of the WMI includes two variables from the external weather source, general weather conditions and temperature, each converted into a normalized score and combined by taking their maximum, so that the more severe of the two effects dominates. Since weather must be evaluated along the entire route rather than at a single point, the route is subdivided into evenly spaced points, and the WMI is computed at each point before retaining the maximum value observed, capturing the risk of a single severe weather event rather than averaging its effects.

\paragraph{Route Time (RT)}\label{route-time-rt}

The Route Time depends on the transit time distribution, as each route is assigned its OTI. Therefore, two forms exist for this indicator. The first one is formally the expected value of the distribution, that is
\begin{equation}
\text{RT}_{\mu}(r) = \mathbb{E}\left\lbrack \text{OTI}(r) \right\rbrack
\label{eq:10}\,.
\end{equation}
While the second one is the two-sided confidence interval
\begin{equation}
\text{RT}_{\beta}(r) = \text{CI}_{\beta}\left\lbrack \text{OTI}(r) \right\rbrack = \left\lbrack L_{\beta},U_{\beta} \right\rbrack
\label{eq:11}
\end{equation}
at an arbitrary confidence level \(\beta\) such that \(\mathbb{P}\left( L_{\beta} \leq X \leq U_{\beta} \right) \geq \beta\) where \(X \sim \text{OTI}(r)\).

\paragraph{Dynamic Route Time (DRT)}\label{dynamic-route-time-drt}

The Dynamic Route Time estimates the transit time of individual routes under real-time conditions, extending the RT indicator by integrating contextual information from the TMI and WMI. The relationship between route times and these contextual variables is complex and nonlinear: traffic and weather variability jointly affect travel time, but their individual contributions are uneven, and neither combines with the others in a simple additive way. Given this complexity, the calculation basis of this indicator is carried out as a supervised regression task using Machine Learning. Accordingly, this indicator takes two forms. The pointwise form is the output of a regression model, which estimates the actual travel time of a given route from a feature vector \(\mathbf{x}(r,t)\) defined in Table~\ref{tab:2}, comprising RT, TMI, WMI, and the route's geodesic distance \(d\), that is
\begin{equation}
\text{DRT}_{\mu}(r,t) = f_{\mathrm{ML}}\left( \mathbf{x}(r,t) \right)\ ,\ \ \left\{ \text{RT}_{\mu}(r)\ ,\ \text{TMI}(r,t)\ ,\ \text{WMI}(r,t)\ ,\ d(r) \right\} \in \mathbf{x}(r,t)
\label{eq:12}\,.
\end{equation}
While the second form is the two-sided uncertainty interval
\begin{equation}
\text{DRT}_{\beta}(r,t) = \left\lbrack l_{\beta}\ ,u_{\beta} \right\rbrack \cdot \text{DRT}_{\mu}(r,t)
\label{eq:13}
\end{equation}
with \(l_{\beta} = 1 - \beta\gamma\), and \(u_{\beta} = 1 + \beta\gamma\) where \(\beta\) is an arbitrary scaling factor, and \(\gamma\) is a relative margin derived from the MAPE calculated during the machine learning model training, used as a proxy for predictive uncertainty. This heuristic does not conform to the probabilistic framework of other indicators, however it presents a practical solution under limited data availability. This is because more rigorous approaches such as quantile regression~\citep{ref22_Koenker1978} or Bayesian methods~\citep{ref21_Neal1996} are data intensive, and are not suited for the available data for model evaluation, as illustrated in the Results section. This approach presents two limitations: it assumes prediction errors are uniform across routes, and it produces a symmetric interval around \(\text{DRT}_{\mu}(r,t)\), whereas actual errors may be skewed.

\subsubsection{Real-time Path-Level LCDIs}\label{real-time-path-level-lcdis}

\begin{figure}[htbp]
\centering
\begin{adjustbox}{max width=\linewidth}
\begin{tikzpicture}[x=1cm,y=-1cm,scale=0.95,
  >={Stealth[length=2.2mm]},
  sup/.style={circle,draw=black!70,fill=supplier,minimum size=7mm,inner sep=0pt},
  int/.style={circle,draw=black!70,fill=intermediate,minimum size=7mm,inner sep=0pt},
  man/.style={circle,draw=black!70,fill=manuf,minimum size=11mm,inner sep=0pt},
  faded/.style={draw opacity=0.2,fill opacity=0.25,text opacity=0.3},
  half/.style={draw opacity=0.6,fill opacity=0.7},
  dim/.style={->,black!22,thin},
  hot/.style={->,black,line width=1.1pt},
  ann/.style={decorate,decoration={brace,amplitude=5pt},bracecol,thin},
  lead/.style={bracecol,thin}]
  \node[sup,faded] (s2) at (5.0,0.3) {$s_2$};
  \node[sup,faded] (s1) at (1.9,1.4) {$s_1$};
  \node[sup,faded] (s3) at (8.1,1.4) {$s_3$};
  \node[int] (i1) at (3.1,2.4) {$i_1$};
  \node[int,faded] (i2) at (5.0,1.9) {$i_2$};
  \node[int,faded] (i3) at (6.8,2.4) {$i_3$};
  \node[int] (i4) at (3.5,3.9) {$i_4$};
  \node[int,faded] (i5) at (5.0,3.6) {$i_5$};
  \node[int,faded] (i6) at (6.5,3.9) {$i_6$};
  \node[man] (m) at (5.0,5.3) {$m$};
  \draw[dim] (s1) -- (i1);
  \draw[dim] (s1) -- (i2);
  \draw[dim] (s2) -- (i1);
  \draw[dim] (s2) -- (i2);
  \draw[dim] (s2) -- (i3);
  \draw[dim] (s3) -- (i3);
  \draw[hot] (i1) -- (i4);
  \draw[dim] (i2) -- (i4);
  \draw[dim] (i2) -- (i5);
  \draw[dim] (i2) -- (i6);
  \draw[dim] (i3) -- (i5);
  \draw[dim] (i3) -- (i6);
  \draw[hot] (i4) -- (m);
  \draw[dim] (i5) -- (m);
  \draw[dim] (i6) -- (m);
  \node[anchor=west] at (8.5,1.4) {Supplier Site};
  \node[anchor=west] at (5.6,5.3) {Manufacturer};
  \draw[decorate,decoration={brace,amplitude=5pt}] (7.6,2.2) -- (7.6,4.1) node[midway,right=6pt] {Intermediate Sites};
  \def\yc{4.0}\def\xa{0.54}\def\xb{2.5}
  \draw[ann] (\xb,\yc+2.0) -- (\xb,\yc-2.0);
  \draw[ann] (\xa,\yc-0.75) -- (\xa,\yc+0.75);
  \draw[lead] (\xa+0.18,\yc) -- (\xb-0.18,\yc);
  \node[cyan!70!blue,anchor=south] at ({(\xa+\xb)/2},\yc+0.03) {Path};
  \node[anchor=north] at ({(\xa+\xb)/2},\yc-0.03) {$\pi$};
  \node[anchor=east,align=right] at (\xa-0.08,\yc) {$\text{TT} = f(\text{ST})$\\[2pt]$\text{PT} = f(\text{VT},\,\text{RT},\,\text{DRT})$\\[2pt]$\text{TFST} = \alpha\cdot\text{TT} + (1-\alpha)\cdot\text{PT}$};
\end{tikzpicture}
\end{adjustbox}
\caption[Path-level indicators on the Supply Chain Graph]{Path-level indicators on the Supply Chain Graph. Building on the SCGM representation, the path-level indicators are computed along a sequence of routes from a supplier site to the manufacturer, exemplified for \(\pi = \left( i_{1},i_{4},m \right)\). The Transit Time (TT), derived from the Shipment Time (ST), and the Path Time (PT), derived from the Vertex Time (VT) and Dynamic Route Time (DRT), are shown, together with the Total Forecasted Shipment Time (TFST), obtained as their convex combination by the TFST weighting factor \(\alpha\) as hyperparameter.}
\label{fig:6}
\end{figure}
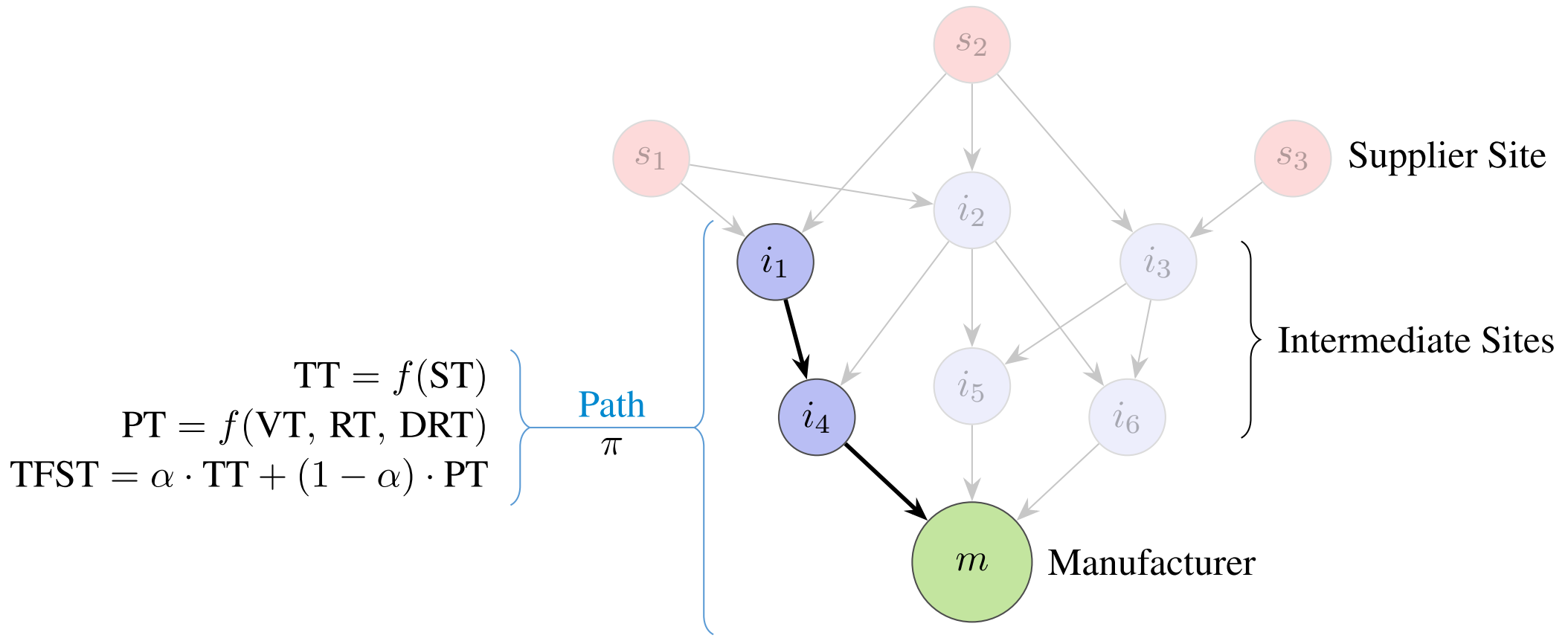

These indicators are global descriptors that aggregate route-level and vertex-level information along a path from a given vertex to the manufacturer, to estimate the remaining shipment duration. They cannot capture the dispatch stage of the order fulfilment process, which is instead addressed by the graph-level LCDI introduced in the next section. Figure~\ref{fig:6} illustrates the path-level LCDIs within the SCGM representation. The remainder of this section defines each path-level LCDI in detail, with a summary provided in Table~\ref{tab:C1}.

\paragraph{Transit Time (TT)}\label{transit-time-tt}

The Transit Time estimates the residual transport time of a shipment along a path from a supplier site with a specific carrier, based on historical data from the ST indicator. Therefore, two forms exist for this indicator. The first one is a scalar, evaluated as
\begin{equation}
\text{TT}_{\mu}(s,c,t) = \max\left( \text{ST}_{\mu}(s,c) - \left( t - t_{1} \right),\ 0 \right)
\label{eq:14}\,.
\end{equation}
While the second one is the two-sided confidence interval dependent on \(\text{ST}_{\beta}(s,c)\)
\begin{equation}
\text{TT}_{\beta}(s,c,t) = \max\left( \text{ST}_{\beta}(s,c) - \left( t - t_{1} \right),\ 0 \right)
\label{eq:15}
\end{equation}
at an arbitrary confidence level \(\beta\). Analogously to the \(\text{DRT}_{\beta}(r,t)\) this form does not conform to the probabilistic framework of other indicators, however it presents a practical solution to obtain an uncertainty bound, reflecting the potential variation in shipment duration based on past observations. Although this indicator is formally defined as a function of the elapsed shipment time \(\left( t - t_{1} \right)\), as it is recomputed at each carrier tracking update, this elapsed time can be interpreted as dependent of the shipment's current vertex \(v(t)\) along a path. Therefore, the TT is to be understood as being implicitly vertex and path-dependent.

\paragraph{Path Time (PT)}\label{path-time-pt}

The Path Time estimates the residual transport time of a shipment along a path from a supplier site with a specific carrier, based on real-time data from both the VT and RT indicators. Two forms exist for this indicator, the first one is a scalar, evaluated as followis. For any path \(\pi = \left( v_{0},v_{1},\ldots,v_{n} \right)\) and a route sequence \(\rho = \left( r_{0},r_{1},\ \ldots,r_{n - 1} \right)\), it is possible to define the residual sub-path as \(\pi^{\prime}\left( v_{i} \right) = \left( v_{i},v_{i + 1},\ldots,v_{n} \right)\) and residual sub-route as \(\rho^{\prime}\left( v_{i} \right) = \left( r_{i},\ r_{i + 1},\ldots,r_{n - 1} \right)\), with \(r_{i} = \left( v_{i},v_{i + 1} \right)\), then for a vertex \(v(t) \in \pi\) evaluated at a transit time \(t\), the indicator is
\begin{equation}
\text{PT}_{\mu}\left( s,c,v(t) \right) = \sum_{u \in \pi^{\prime}(v)}^{}{\text{VT}_{\mu}(u)} + \sum_{r \in \rho^{\prime}(v)}^{}{\text{RT}_{\mu}(r)}
\label{eq:16}\,.
\end{equation}
Also, the RT may be replaced with the DRT to incorporate data from traffic and weather conditions, yielding
\begin{equation}
\text{PT}_{\mu}\left( s,c,v(t) \right) = \sum_{u \in \pi^{\prime}(v)}^{}{\text{VT}_{\mu}(u)} + \sum_{r \in \rho^{\prime}(v)}^{}{\text{DRT}_{\mu}(r,t)}
\label{eq:17}\,.
\end{equation}
And the second one is the two-sided confidence interval, dependent on the interval-based versions at an arbitrary confidence level \(\beta\). That is,
\begin{equation}
\text{PT}_{\beta}\left( s,c,v(t) \right) = \sum_{u \in \pi^{\prime}(v)}^{}{\text{VT}_{\beta}(u)} + \sum_{r \in \rho^{\prime}(v)}^{}{\text{RT}_{\beta}(r)}
\label{eq:18}
\end{equation}
or, in the same fashion as the first form,
\begin{equation}
\text{PT}_{\beta}\left( s,c,v(t) \right) = \sum_{u \in \pi^{\prime}(v)}^{}{\text{VT}_{\beta}(u)} + \sum_{r \in \rho^{\prime}(v)}^{}{\text{DRT}_{\beta}(r,t)}
\label{eq:19}\,.
\end{equation}
Analogously to the \(\text{DRT}_{\beta}(r)\) this form does not conform to the probabilistic framework of other indicators, however it presents a practical solution to obtain an uncertainty bound, reflecting the potential variation in shipment duration based on past observations. The formal derivation of this indicator is detailed in the first author's MSc thesis~\citep{ref01_Volpato2025}.

\paragraph{Total Forecasted Shipment Time (TFST)}\label{total-forecasted-shipment-time-tfst}

The Total Forecasted Shipment Time balances information from the TT and PT indicators to provide an estimate of the residual transport time of a shipment along a path from a supplier site with a specific carrier. The TT relies on historical site-carrier data, and the PT incorporates real-time and path-dependent information, such as vertex residence times and route transit times. The calculation basis is a convex combination with a hyperparameter (weighing factor) \(\alpha\), and two forms exist for this indicator. The first one is a scalar, evaluated as:
\begin{equation}
\text{TFST}_{\mu}(s,c,t;\alpha) = \alpha \cdot \text{TT}_{\mu}(s,c,t) + (1 - \alpha) \cdot \text{PT}_{\mu}\left( s,c,v(t) \right)
\label{eq:20}\,.
\end{equation}
And the second one is the two-sided confidence interval, dependent on the interval-based versions at an arbitrary confidence level \(\beta\). That is,
\begin{equation}
\text{TFST}_{\beta}(s,c,t;\alpha) = \alpha \cdot \text{TT}_{\beta}(s,c,t) + (1 - \alpha) \cdot \text{PT}_{\beta}\left( s,c,v(t) \right)
\label{eq:21}\,.
\end{equation}
The weighting factor \(\alpha\) controls the relative contribution of each indicator, as higher values favour historical information, while lower values favour dynamic, path-specific information. The weighting factor is different for each site-carrier pair, and is initialized either as a constant equal to 0.5, or as a monotonically decreasing function of the fraction of elapsed shipment time and expected ST indicator, that is
\begin{equation}
\alpha(\tau) \in \lbrack 0,1\rbrack\ ,\ \ \tau\left( s,c,t_{1},t \right) = \frac{t - t_{1}}{\text{ST}_{\mu}(s,c)}\ ,\ \ \tau \geq 0
\label{eq:22}\,.
\end{equation}
Two versions of \(\alpha(\tau)\) were considered, a linear function \(\alpha(\tau) = \max(1 - \tau,0)\) and a polynomial function parametrized by \(p \in \lbrack 0,1\rbrack\), namely
\begin{equation}
\alpha(\tau;p) = \max\left\lbrack (1 - \tau)^{q(p)},0 \right\rbrack,\ \ q(p) = \frac{1 - p}{p}
\label{eq:23}\,.
\end{equation}
The linear function was chosen because it assigns dominant weight to the historical estimates in the low- \(\tau\) regime, near the start of the shipment, while shifting weight toward the real-time estimate as \(\tau\) increases toward the end of the shipment. For the parametrized function, it is possible to obtain an optimal value of \(p\) by minimizing the MSE between the historical values of remaining transit times \(\left( t_{n}(s,c) - t \right)\) with the \(\text{TFST}_{\mu}(s,c,t;\alpha)\), summing over each tracking time \(t\) and orders for a site-carrier pair. This results in a one-dimensional constrained optimization problem, efficiently solved by standard bounded scalar optimization methods.

\subsubsection{Real-time Graph-Level LCDIs}\label{real-time-graph-level-lcdis}

\begin{figure}[htbp]
\centering
\begin{adjustbox}{max width=\linewidth}
\begin{tikzpicture}[x=1cm,y=-1cm,scale=0.95,
  >={Stealth[length=2.2mm]},
  sup/.style={circle,draw=black!70,fill=supplier,minimum size=7mm,inner sep=0pt},
  int/.style={circle,draw=black!70,fill=intermediate,minimum size=7mm,inner sep=0pt},
  man/.style={circle,draw=black!70,fill=manuf,minimum size=11mm,inner sep=0pt},
  faded/.style={draw opacity=0.2,fill opacity=0.25,text opacity=0.3},
  half/.style={draw opacity=0.6,fill opacity=0.7},
  dim/.style={->,black!22,thin},
  hot/.style={->,black,line width=1.1pt},
  ann/.style={decorate,decoration={brace,amplitude=5pt},bracecol,thin},
  lead/.style={bracecol,thin}]
  \node[sup] (s2) at (5.0,0.3) {$s_2$};
  \node[sup,faded] (s1) at (1.9,1.4) {$s_1$};
  \node[sup,faded] (s3) at (8.1,1.4) {$s_3$};
  \node[int,faded] (i1) at (3.1,2.4) {$i_1$};
  \node[int] (i2) at (5.0,1.9) {$i_2$};
  \node[int,faded] (i3) at (6.8,2.4) {$i_3$};
  \node[int] (i4) at (3.5,3.9) {$i_4$};
  \node[int] (i5) at (5.0,3.6) {$i_5$};
  \node[int] (i6) at (6.5,3.9) {$i_6$};
  \node[man] (m) at (5.0,5.3) {$m$};
  \draw[dim] (s1) -- (i1);
  \draw[dim] (s1) -- (i2);
  \draw[dim] (s2) -- (i1);
  \draw[dim] (s2) -- (i2);
  \draw[dim] (s2) -- (i3);
  \draw[dim] (s3) -- (i3);
  \draw[dim] (i1) -- (i4);
  \draw[hot] (i2) -- (i4);
  \draw[hot] (i2) -- (i5);
  \draw[hot] (i2) -- (i6);
  \draw[dim] (i3) -- (i5);
  \draw[dim] (i3) -- (i6);
  \draw[hot] (i4) -- (m);
  \draw[hot] (i5) -- (m);
  \draw[hot] (i6) -- (m);
  \node[anchor=west] at (8.5,1.4) {Supplier Site};
  \node[anchor=west] at (5.6,5.3) {Manufacturer};
  \draw[decorate,decoration={brace,amplitude=5pt}] (7.6,2.2) -- (7.6,4.1) node[midway,right=6pt] {Intermediate Sites};
  \node[anchor=east] (edt) at (3.9,0.3) {EDT};
  \draw[lead] (edt.east) -- (s2.west);
  \def\yc{3.85}\def\xa{0.54}\def\xb{2.5}
  \draw[ann] (\xb,\yc+2.2) -- (\xb,\yc-2.2);
  \draw[ann] (\xa,\yc-0.75) -- (\xa,\yc+0.75);
  \draw[lead] (\xa+0.18,\yc) -- (\xb-0.18,\yc);
  \node[cyan!70!blue,anchor=south] at ({(\xa+\xb)/2},\yc+0.03) {Graph};
  \node[anchor=north] at ({(\xa+\xb)/2},\yc-0.03) {$\Pi(i_2)$};
  \node[anchor=east,align=right] at (\xa-0.08,\yc) {$\text{E-PT} = f(\text{PT})$\\[2pt]$\text{E-TFST} = \alpha\cdot\text{TT} + (1-\alpha)\cdot\text{E-PT}$\\[2pt]$\text{EODT} = \text{EDT} + (t-t_1) + \text{E-TFST}$};
\end{tikzpicture}
\end{adjustbox}
\caption[Graph-level indicators on the Supply Chain Graph]{Graph-level indicators on the Supply Chain Graph. Building on the SCGM representation, the graph-level indicator aggregates information across all possible paths from the current vertex to the manufacturer, exemplified here for the graph set \(\Pi(v)\) where \(v = i_{2}\). The Expected Path Time (E-PT) and Expected Total Forecasted Shipment Time (E-TFST) are shown, together with the Estimated Overall Delivery Time (EODT) as the final indicator.}
\label{fig:7}
\end{figure}
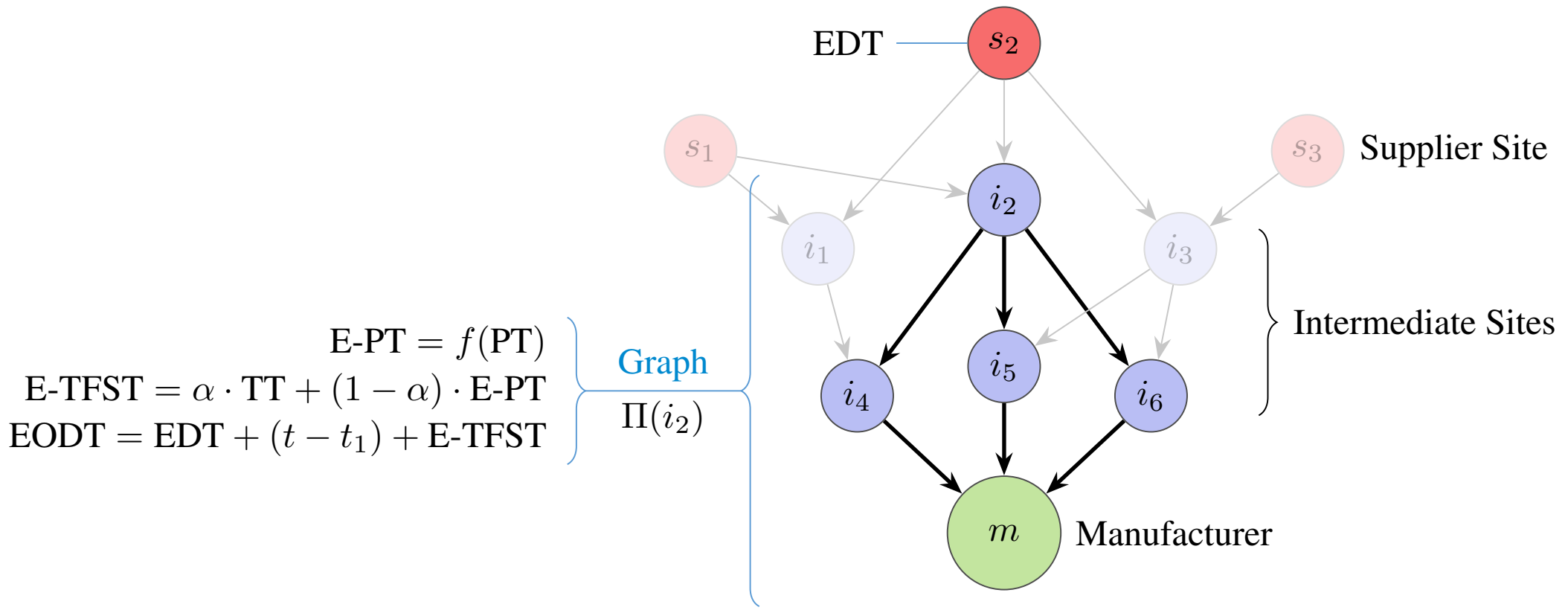

These indicators are semi-global descriptors that provide a comprehensive forecast of the total delivery time, combining the dispatch stage with the expected shipment duration across the set \(\Pi(v) = \left\{ \pi_{1},\pi_{2},\ \ldots,\pi_{\ell} \right\}\) of all possible paths from a vertex \(v\) to the manufacturer. They aggregate information from every lower level of the hierarchy into a single value compounding the probability estimates along each path \(\pi_{i} \in \Pi(v)\). Figure~\ref{fig:7} illustrates the graph-level LCDIs within the SCGM representation. The remainder of this section defines each path-level LCDI in detail, with a summary provided in Table~\ref{tab:C1}.

\paragraph{Expected Path Time (E-PT)}\label{expected-path-time-e-pt}

The Expected Path Time estimates the weighted probability expectation of residual transport times for a shipment potentially crossing all possible paths from the current vertex to the manufacturer, according to the transition matrix computed through the discrete-time Markov process formulation of the SCGM. A complete derivation is provided in Appendix~\ref{appendix-D}.

\paragraph{Expected Total Forecasted Shipment Time (E-TFST)}\label{expected-total-forecasted-shipment-time-e-tfst}

The Expected Total Forecasted Shipment Time balances information from the TT and E-PT indicators to provide an estimate of the weighted probability expectation of residual transport times for a shipment potentially crossing all possible paths from the current vertex to the manufacturer. The TT relies on historical site-carrier data, and the E-PT incorporates real-time and path-dependent information, such as vertex residence times and route transit times. The calculation basis is the same convex combination with a hyperparameter (weighing factor) \(\alpha \in \lbrack 0,1\rbrack\). A complete derivation is provided in Appendix~\ref{appendix-D}.

\paragraph{Estimated Overall Delivery Time (EODT)}\label{estimated-overall-delivery-time-eodt}

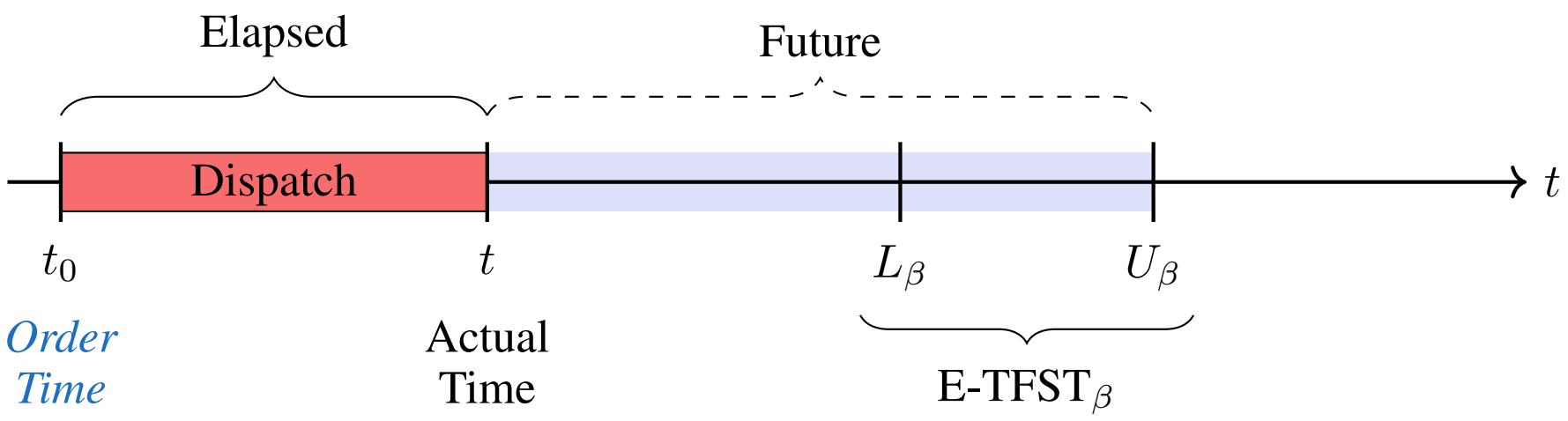
\begin{figure}[htbp]
\centering
\begin{tikzpicture}[x=1cm,y=1cm,scale=1]
  \fill[future] (3.2,-0.22) rectangle (8.2,0.22);
  \draw[->,thick] (-0.4,0) -- (11,0) node[right] {$t$};
  \fill[supplier] (0,-0.22) rectangle (3.2,0.22); \draw (0,-0.22) rectangle (3.2,0.22);
  \node at (1.6,-0.02) {Dispatch};
  \foreach \x in {0,3.2,6.3,8.2} \draw[thick] (\x,-0.3) -- (\x,0.3);
  \draw[decorate,decoration={brace,amplitude=8pt}] (0,0.5) -- (3.2,0.5) node[midway,above=9pt] {Elapsed};
  \draw[dashed,decorate,decoration={brace,amplitude=8pt}] (3.2,0.5) -- (8.2,0.5) node[midway,above=9pt] {Future};
  \node[below] at (0,-0.35) {$t_0$};
  \node[below] at (3.2,-0.35) {$t$};
  \node[below] at (6.3,-0.35) {$L_\beta$};
  \node[below] at (8.2,-0.35) {$U_\beta$};
  \node[labelblue,align=center,font=\itshape] at (0,-1.35) {Order\\Time};
  \node[align=center] at (3.2,-1.35) {Actual\\Time};
  \draw[decorate,decoration={brace,amplitude=6pt,mirror}] (6.0,-1.0) -- (8.5,-1.0) node[midway,below=8pt] {E-TFST$_\beta$};
\end{tikzpicture}
\caption[Time sequence for the interval-based EODT at the dispatching stage]{Time sequence for the interval-based EODT at the dispatching stage. The elapsed portion of the order fulfilment process spans only the dispatch time (EDT). Thus, the future portion of shipment time is estimated by the bounds of the interval-based E-TFST, with lower and upper bounds \(L_{\beta}\) and \(U_{\beta}\) denoting the \(\beta\)-level confidence intervals.}
\label{fig:8}
\end{figure}

\begin{figure}[htbp]
\centering
\begin{tikzpicture}[x=1cm,y=1cm,scale=1]
  \fill[future] (5.6,-0.22) rectangle (9.8,0.22);
  \draw[->,thick] (-0.4,0) -- (11,0) node[right] {$t$};
  \fill[supplier] (0,-0.22) rectangle (2.4,0.22); \draw (0,-0.22) rectangle (2.4,0.22);
  \fill[intermediate] (2.4,-0.22) rectangle (5.6,0.22); \draw (2.4,-0.22) rectangle (5.6,0.22);
  \node at (1.2,-0.02) {Dispatch}; \node at (4.0,-0.02) {Shipment};
  \foreach \x in {0,2.4,5.6,7.4,9.8} \draw[thick] (\x,-0.3) -- (\x,0.3);
  \draw[decorate,decoration={brace,amplitude=8pt}] (0,0.5) -- (5.6,0.5) node[midway,above=9pt] {Elapsed};
  \draw[dashed,decorate,decoration={brace,amplitude=8pt}] (5.6,0.5) -- (9.8,0.5) node[midway,above=9pt] {Future};
  \node[below] at (0,-0.35) {$t_0$};
  \node[below] at (2.4,-0.35) {$t_1$};
  \node[below] at (5.6,-0.35) {$t$};
  \node[below] at (7.4,-0.35) {$L_\beta$};
  \node[below] at (9.8,-0.35) {$U_\beta$};
  \node[labelblue,align=center,font=\itshape] at (0,-1.35) {Order\\Time};
  \node[labelblue,align=center,font=\itshape] at (2.4,-1.35) {Shipment\\Time};
  \node[align=center] at (5.6,-1.35) {Tracking\\Update};
  \draw[decorate,decoration={brace,amplitude=6pt,mirror}] (7.1,-1.0) -- (10.1,-1.0) node[midway,below=8pt] {E-TFST$_\beta$};
\end{tikzpicture}
\caption[Time sequence for the interval-based EODT at the shipping stage]{Time sequence for the interval-based EODT at the shipping stage. The elapsed portion of the order fulfilment process spans the dispatch time (EDT) and shipment time \(\left( t - t_{1} \right)\), forming the order time to the most recent tracking update \(\left( t - t_{0} \right)\). Thus, the future portion of shipment time is estimated by the bounds of the interval-based E-TFST, with lower and upper bounds \(L_{\beta}\) and \(U_{\beta}\) denoting the \(\beta\)-level confidence intervals.}
\label{fig:9}
\end{figure}
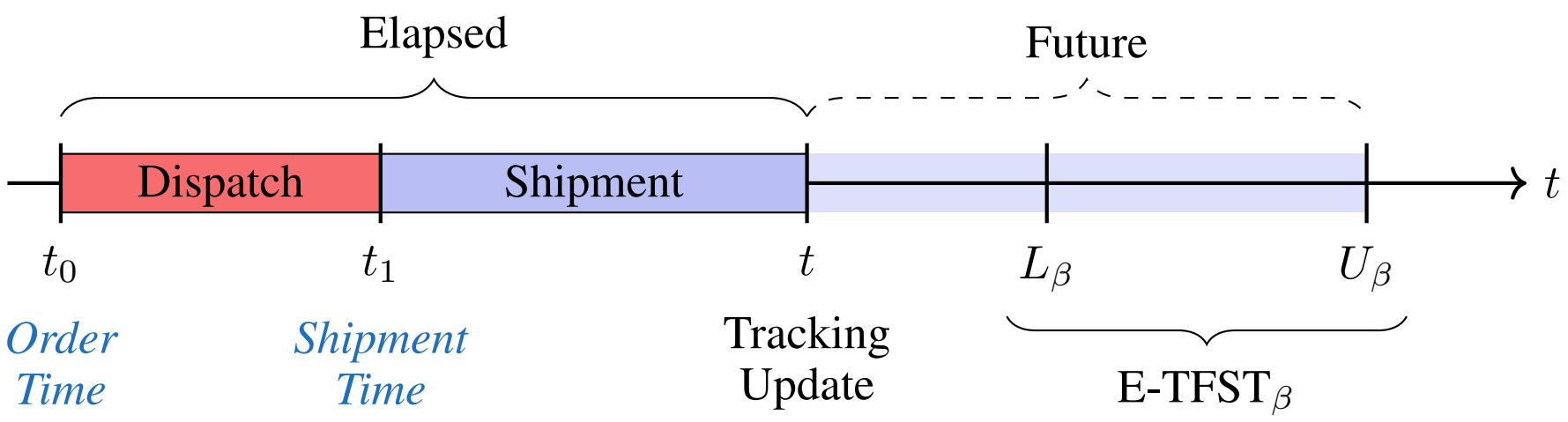

The Estimated Overall Delivery time provides a forecast of the total elapsed time of the logistics process, including dispatch and shipment stages, thus providing an estimate for the delivery time \(t_{n} = t_{0} + \text{EODT}_{\mu}\left( s,c,t_{0},t;\alpha \right)\) appearing in the time-sequence of the order delivery process in Figure~\ref{fig:3}. This indicator is exemplified on the time sequence of the order fulfilment process, as illustrated in Figure~\ref{fig:8} and Figure~\ref{fig:9}, for the dispatching and shipping stages respectively. They show how the interval-based version extends beyond the most recent time update to bound the remaining delivery time. The interval band corresponds to the shaded cyan band in the shipment estimate evolution plots presented in the EODT Estimation Results section.

The calculation basis consists of a simple summation of the EDT, which includes the distinction between dispatching and shipping stages, and the TFST. Two forms exist for this indicator, the first one is a scalar, evaluated as
\begin{equation}
\begin{split}
\text{EODT}_{\mu}\left( s,c,t_{0},t;\alpha \right) &= \text{EDT}\left( s,t_{0},t \right)\\
&\quad + \left\{\begin{aligned}
\left( t - t_{1} \right) + \text{E-TFST}_{\mu}(s,c,t;\alpha)\ ,\ \ t \geq t_{1}\ ,\ \  & \text{shipping stage} \\
\text{E-TFST}_{\mu}(s,c,t;\alpha)\ ,\ \ t < t_{1}\ ,\ \  & \text{dispatching stage}
\end{aligned}\right.
\end{split}
\label{eq:24}\,.
\end{equation}
While the second one is the two-sided confidence interval, dependent on \(\text{E-TFST}_{\beta}(s,c,t;\alpha)\) at an arbitrary confidence level \(\beta\), that is
\begin{equation}
\begin{split}
\text{EODT}_{\beta}\left( s,c,t_{0},t;\alpha \right) &= \text{EDT}\left( s,t_{0},t \right)\\
&\quad + \left\{\begin{aligned}
\left( t - t_{1} \right) + \text{E-TFST}_{\beta}(s,c,t;\alpha)\ ,\ \ t \geq t_{1}\ ,\ \  & \text{shipping stage} \\
\text{E-TFST}_{\beta}(s,c,t;\alpha)\ ,\ \ t < t_{1}\ ,\ \  & \text{dispatching stage}
\end{aligned}\right.
\end{split}
\label{eq:25}\,.
\end{equation}
\section{Use Case Implementation}\label{use-case-implementation}

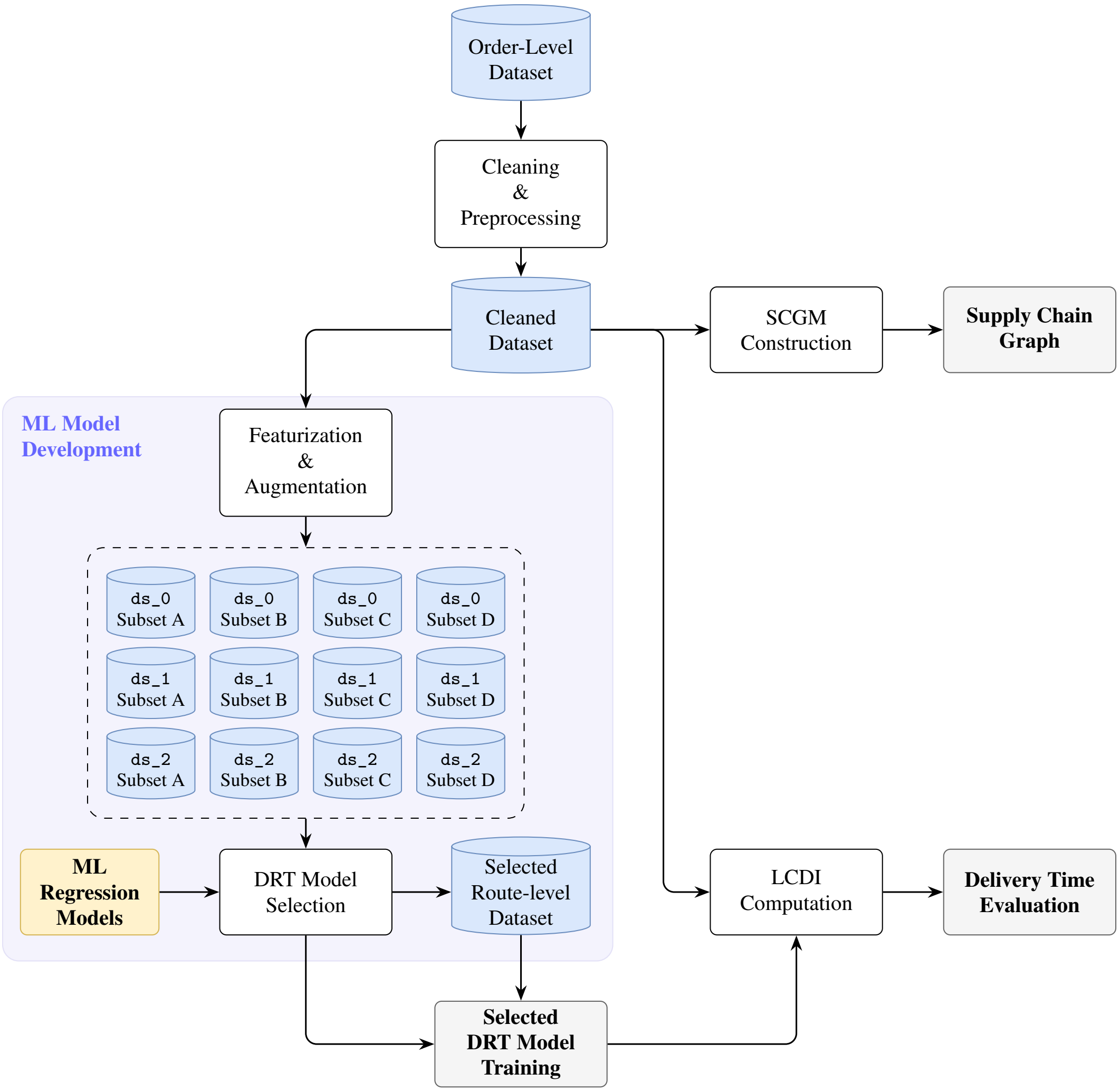
\begin{figure}[!ht]
\centering
\begin{adjustbox}{max width=\linewidth}
\begin{tikzpicture}[x=1cm,y=-1cm,font=\footnotesize,
  >={Stealth[length=2mm]},
  box/.style={draw=black,rounded corners=2.5pt,fill=white,align=center,inner sep=2pt,
              minimum width=2.25cm,minimum height=1.12cm},
  box3/.style={box,inner sep=4pt,minimum height=1.40cm},
  outbox/.style={box,draw=black!60,fill=black!4,font=\footnotesize\bfseries},
  ml/.style={box,draw=mlyellowdraw,fill=mlyellow,minimum width=1.81cm,font=\footnotesize\bfseries},
  db/.style={cylinder,shape border rotate=90,aspect=0.18,draw=dbdraw,fill=dbfill,align=center,
             inner sep=1pt,minimum width=1.81cm,minimum height=1.25cm},
  dbs/.style={db,minimum width=1.15cm,minimum height=0.93cm,font=\scriptsize,aspect=0.22,inner sep=2pt},
  flowarr/.style={->,black,line width=0.6pt,rounded corners=4pt}]
  \fill[mlpanel,rounded corners=6pt] (0.025,5.100) rectangle (8.000,12.475);
  \draw[mlpaneldraw,rounded corners=6pt] (0.025,5.100) rectangle (8.000,12.475);
  \node[anchor=north west,align=left,text=mllabel,font=\footnotesize\bfseries] at (0.150,5.225) {ML Model\\Development};
  \node[db] (order) at (6.800,0.688) {Order-Level\\Dataset};
  \node[box3] (clean) at (6.800,2.450) {Cleaning\\\&\\Preprocessing};
  \node[db] (cleaned) at (6.800,4.225) {Cleaned\\Dataset};
  \node[box] (scgm) at (10.400,4.225) {SCGM\\Construction};
  \node[outbox] (scg) at (13.450,4.225) {Supply Chain\\Graph};
  \node[box3] (feat) at (3.988,5.963) {Featurization\\\&\\Augmentation};
  \node[ml] (mlreg) at (1.163,11.575) {ML\\Regression\\Models};
  \node[box] (drtsel) at (3.988,11.575) {DRT Model\\Selection};
  \node[db] (selrl) at (6.800,11.575) {Selected\\Route-level\\Dataset};
  \node[box] (indic) at (10.400,11.575) {LCDI\\Computation};
  \node[outbox] (dte) at (13.450,11.575) {Delivery Time\\Evaluation};
  \node[outbox] (seldrt) at (6.800,13.562) {Selected\\DRT Model\\Training};
  \path let \p1=(feat.south), \p2=(drtsel.north) in coordinate (gc) at (3.988cm,{0.5*\y1+0.5*\y2-2.566pt});
  \node[dbs] (d00) at ($(gc)+(-2.025,-1.050)$) {\texttt{ds\_0}\\Subset A};
  \node[dbs] (d01) at ($(gc)+(-0.675,-1.050)$) {\texttt{ds\_0}\\Subset B};
  \node[dbs] (d02) at ($(gc)+(0.675,-1.050)$) {\texttt{ds\_0}\\Subset C};
  \node[dbs] (d03) at ($(gc)+(2.025,-1.050)$) {\texttt{ds\_0}\\Subset D};
  \node[dbs] (d10) at ($(gc)+(-2.025,0.000)$) {\texttt{ds\_1}\\Subset A};
  \node[dbs] (d11) at ($(gc)+(-0.675,0.000)$) {\texttt{ds\_1}\\Subset B};
  \node[dbs] (d12) at ($(gc)+(0.675,0.000)$) {\texttt{ds\_1}\\Subset C};
  \node[dbs] (d13) at ($(gc)+(2.025,0.000)$) {\texttt{ds\_1}\\Subset D};
  \node[dbs] (d20) at ($(gc)+(-2.025,1.050)$) {\texttt{ds\_2}\\Subset A};
  \node[dbs] (d21) at ($(gc)+(-0.675,1.050)$) {\texttt{ds\_2}\\Subset B};
  \node[dbs] (d22) at ($(gc)+(0.675,1.050)$) {\texttt{ds\_2}\\Subset C};
  \node[dbs] (d23) at ($(gc)+(2.025,1.050)$) {\texttt{ds\_2}\\Subset D};
  \node[draw,dashed,rounded corners=6pt,inner sep=7pt,fit=(d00)(d23)] (grp) {};
  \draw[flowarr] (order) -- (clean);
  \draw[flowarr] (clean) -- (cleaned);
  \draw[flowarr] (cleaned.east) -- (scgm.west);
  \draw[flowarr] (scgm) -- (scg);
  \draw[flowarr] (cleaned.west) -| (feat.north);
  \draw[flowarr] (cleaned.east) -- (8.662,4.225) |- (indic.west);
  \draw[flowarr] (feat.south) -- (grp.north);
  \draw[flowarr] (grp.south) -- (drtsel.north);
  \draw[flowarr] (mlreg) -- (drtsel);
  \draw[flowarr] (drtsel) -- (selrl);
  \draw[flowarr] (selrl.south) -- (seldrt.north);
  \draw[flowarr] (drtsel.south) |- (seldrt.west);
  \draw[flowarr] (seldrt.east) -| (indic.south);
  \draw[flowarr] (indic) -- (dte);
\end{tikzpicture}
\end{adjustbox}
\caption[SCGM-LCDI Framework Implementation Process]{SCGM-LCDI Framework Implementation Process. Starting from the order-level dataset, data cleaning and preprocessing yield the route-level dataset, which feeds two branches: SCGM construction, producing the Supply Chain Graph, and the featurization and augmentation phase, generating twelve dataset variants (three augmentation versions \texttt{ds\_0} to \texttt{ds\_2}, each with four feature subsets A to D). Candidate ML regression models are evaluated on these variants for the DRT indicator, resulting in the selection of a model trained on a specific dataset. The Supply Chain Graph, the selected dataset, and the DRT model are combined in the LCDI computation to produce the EODT estimates.}
\label{fig:10}
\end{figure}

The SCGM and LCDIs introduced in the previous sections provide the theoretical foundation for developing, validating, and assessing both the graph representation and the predictive components of the indicator framework in a systematic manner. Given the real-world constraints of supply chain datasets, such as imbalanced distribution of orders across supplier sites and carriers, incomplete tracking timestamps for a subset of shipment records, and heterogeneous formats and granularity of operational records across different data sources. The use case implementation shall follow a five-stage process, each isolating the contribution of a distinct component while remaining aligned with the goal of predicting delivery times under industrial conditions. In this context, a delay is understood in relation to the predicted delivery times: it occurs whenever the actual delivery time exceeds the expected value, or when the predicted delivery time itself grows unexpectedly large relative to historical norms, signalling that the shipment is at risk of running behind schedule. The implementation of the SCGM-LCDI framework is illustrated in Figure~\ref{fig:10}.

The implementation process as presented in the previous figure is structured as follows:

\begin{enumerate}
\item \textit{Order-level Dataset Preparation}. Dataset retrieval from the use case provider, used to construct the SCGM and to estimate the LCDIs. Data cleaning and preprocessing steps ensure that the raw order procurement records are cleaned, consistently formatted, and freed of missing or malformed entries that would bias the calculation of LCDIs, together with training and validation of machine learning models.
\item \textit{SCGM Construction}. The graph-based representation of suppliers, intermediate facilities, and the manufacturer is instantiated, together with the carrier-conditioned routing dynamics described in the Supply Chain Graph Model section.
\item \textit{Indicator Computation}. The historical and real-time indicators framework is applied to the graph to derive dispatch, shipment, and delivery estimates as described in the Logistic Chain Disruption Indicators section. Also, the LCDI framework is illustrated in the Modelling Approach section, so it is not discussed here.
\item \textit{DRT Model Development}. The machine learning model for the regression of the Dynamic Route Time (DRT) indicator is carried out, encompassing an augmentation of the route-level dataset, feature engineering, model selection, training, and validation. The detailed approach of this phase is reported here in summarized form, with full details available in the first author's MSc thesis~\citep{ref01_Volpato2025}.
\item \textit{Delivery Time Evaluation}. The complete framework is tested and validated under experimental settings on the use case, assessing the predictive performance of the resulting delivery time estimates. This phase is presented in the Results section, which reports overall performance metrics, a preliminary sensitivity analysis, and visualizations of the estimated time intervals for a representative sample of orders.
\end{enumerate}

As the core of this work, this staged process provides a transparent and reproducible pathway from raw operational data to interpretable prediction results, ensuring that each component, from the graph representation to the final delivery time estimates, can be independently inspected, validated, and, if needed, refined as more supply chain data becomes available.

\subsection{Order-level Dataset Preparation}\label{order-level-dataset-preparation}

The foundation of the proposed framework is a database provided by FAE Technology S.p.A., structured around order procurement records. Selected corresponding tables are provided in Appendix~\ref{appendix-B}. The list of database fields of interest such as tracking links and statuses provided in Table~\ref{tab:B1}. The most recent database dump in July 2025 contained 6551 order records, of which only 98 included valid tracking links from carriers. From these, 21 distinct supplier sites and 16 suppliers were identified, together with their associated orders. Table~\ref{tab:B2} reports, for each site, the associated supplier, geographic origin, country, and number of orders, and Table~\ref{tab:B3} reports the distribution of orders across carriers, confirming that a small set of major logistics providers, namely UPS, DHL, DHL Express, and FedEx, which accounted for the vast majority of shipments. Preliminary analysis shows that a single US-based site accounted for the majority of the orders, reflecting the pronounced imbalance discussed in the Logistic Chain Disruption Indicators section. This imbalance motivates the usage of a heuristic, rather than probabilistic, formulation for some of the interval-based route-level and path-level indicators. Featurising this foundational data consists of reconstructing sequences of shipping events from the tracking data provided by the various carriers, or extracting traffic and weather information. To this end, several external services were integrated, as summarized in Table~\ref{tab:1}.

\begin{table}[htbp]
\centering
\caption[External services integrated for dataset featurisation]{External services integrated for dataset featurisation. For each service, the type of information it provides, namely traffic, weather, or supplier holiday data, is reported alongside its source.}\label{tab:1}
\small
\begin{tabular}{@{}lll@{}}
\toprule
Name & Type & Source \\
\midrule
17Track & Shipping Events & \url{https://www.17track.net/en/api} \\
TomTom & Traffic & \url{https://docs.tomtom.com/traffic-api/api-explorer} \\
Visual Crossing & Weather & \url{https://www.visualcrossing.com/weather-api/} \\
Calendarific & Supplier Holidays & \url{https://calendarific.com/api-documentation} \\
\bottomrule
\end{tabular}
\end{table}

Data cleaning procedures for the order-level dataset addressed issues such as multiple tracking links per record and unreliable estimated delivery dates from the manufacturer's side. Comparison with carrier-provided delivery data revealed significant inaccuracies in these estimates, therefore this work relied exclusively on delivery dates provided by the carriers. For the route-level and vertex-level data used for computing the distributions for vertex-level residence time (ORI) and route-level transit time (OTI), data cleaning was limited to interquartile range (IQR) outlier removal, applied only when at least 10 observations were available. The detailed IQR procedure is illustrated in the first author's MSc thesis~\citep{ref01_Volpato2025}. Unlike dispatch times, which were adjusted for supplier holidays, carrier operations were assumed to be unaffected by national holidays, and no holiday-based filtering was applied for ORI and OTI computation. Consequently, the two distributions were summarized descriptive statistics, such as the sample mean, median, and quantiles, rather than being inferred. In addition, due to data scarcity, all ORI and OTI were aggregated across carriers. While reasonable, this assumption does not guarantee that different carriers use identical transportation methods or processing procedures at the same location; with more data, computing these metrics separately per carrier would be a natural refinement. Inference to the Gamma distribution was limited to a small subset of sites containing at minimum 30 orders, and was employed to evaluate historical LCDIs, namely the DT and ST. The fitting was performed via Maximum Likelihood Estimation and validated through Q-Q plots and a bootstrap Kolmogorov-Smirnov test, while the remaining, sparser samples were summarized using descriptive statistics, such as the sample mean, median, and quantiles. The detailed fitting procedure is illustrated Section 5.1 in the first author's MSc thesis~\citep{ref01_Volpato2025}.

\subsection{SCGM Construction}\label{scgm-construction}

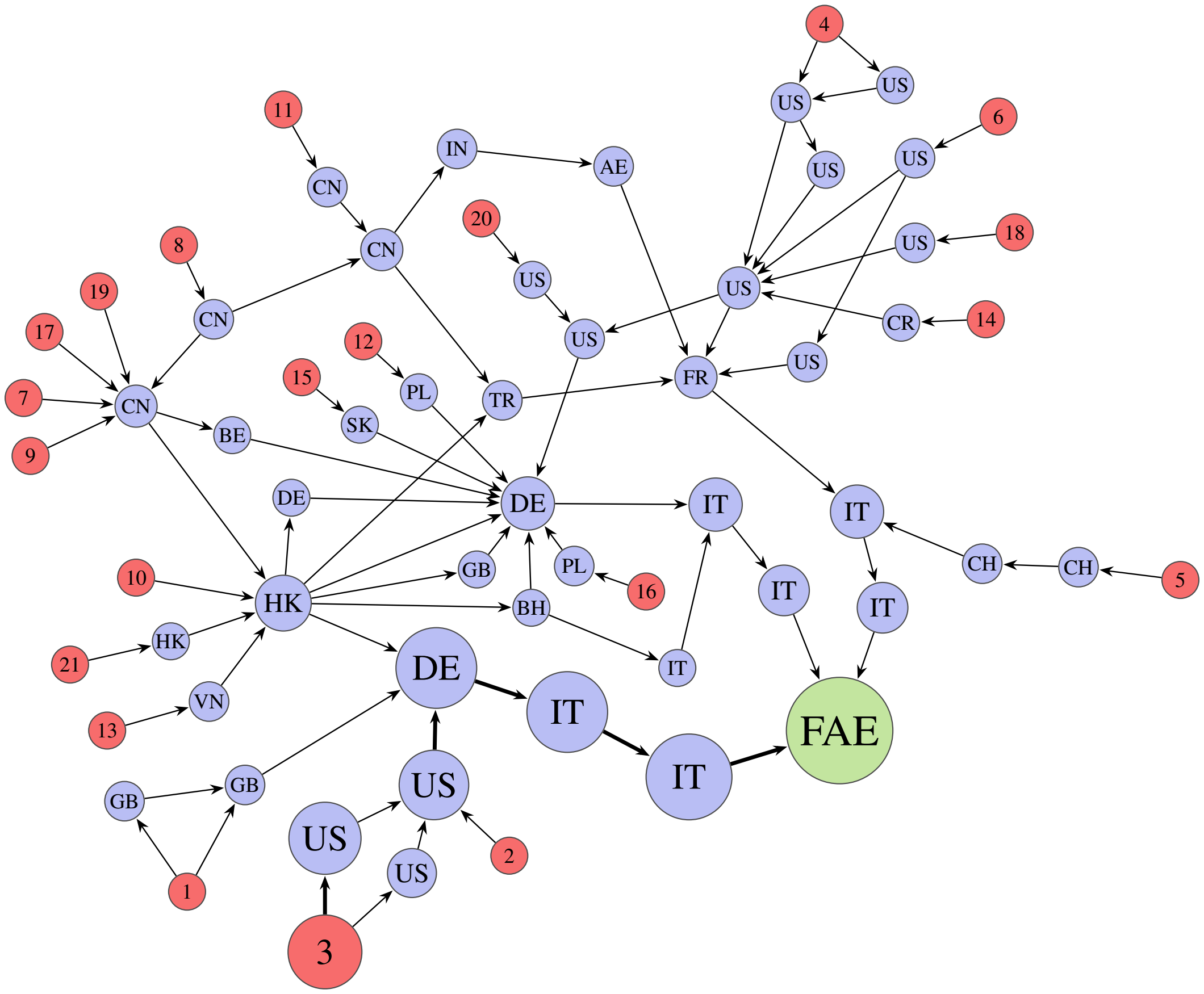
\begin{figure}[htbp]
\centering
\begin{tikzpicture}[x=1cm,y=-1cm,
  >={Stealth[length=1.6mm]},
  v/.style={circle,draw=black!70,inner sep=0pt,line width=0.4pt},
  sup/.style={v,fill=supplier}, int/.style={v,fill=intermediate}, man/.style={v,fill=manuf},
  arc/.style={->,black,line width=0.45pt,shorten >=0.3pt},
  flow/.style={->,black,line width=1.3pt,shorten >=0.3pt}]
  \node[sup,minimum size=0.43cm,font=\scriptsize] (s1) at (2.187,10.409) {1};
  \node[sup,minimum size=0.43cm,font=\scriptsize] (s2) at (5.940,9.990) {2};
  \node[sup,minimum size=0.86cm,font=\large] (s3) at (3.793,11.111) {3};
  \node[sup,minimum size=0.43cm,font=\scriptsize] (s4) at (9.612,0.297) {4};
  \node[sup,minimum size=0.43cm,font=\scriptsize] (s5) at (13.756,6.777) {5};
  \node[sup,minimum size=0.43cm,font=\scriptsize] (s6) at (11.637,1.377) {6};
  \node[sup,minimum size=0.43cm,font=\scriptsize] (s7) at (0.283,4.657) {7};
  \node[sup,minimum size=0.43cm,font=\scriptsize] (s8) at (2.092,2.876) {8};
  \node[sup,minimum size=0.43cm,font=\scriptsize] (s9) at (0.364,5.332) {9};
  \node[sup,minimum size=0.43cm,font=\scriptsize] (s10) at (1.593,6.750) {10};
  \node[sup,minimum size=0.43cm,font=\scriptsize] (s11) at (3.308,1.296) {11};
  \node[sup,minimum size=0.43cm,font=\scriptsize] (s12) at (4.239,3.996) {12};
  \node[sup,minimum size=0.43cm,font=\scriptsize] (s13) at (1.256,8.532) {13};
  \node[sup,minimum size=0.43cm,font=\scriptsize] (s14) at (11.489,3.740) {14};
  \node[sup,minimum size=0.43cm,font=\scriptsize] (s15) at (3.523,4.415) {15};
  \node[sup,minimum size=0.43cm,font=\scriptsize] (s16) at (7.533,6.912) {16};
  \node[sup,minimum size=0.43cm,font=\scriptsize] (s17) at (0.526,3.888) {17};
  \node[sup,minimum size=0.43cm,font=\scriptsize] (s18) at (11.826,2.727) {18};
  \node[sup,minimum size=0.43cm,font=\scriptsize] (s19) at (1.161,3.415) {19};
  \node[sup,minimum size=0.43cm,font=\scriptsize] (s20) at (5.616,2.565) {20};
  \node[sup,minimum size=0.43cm,font=\scriptsize] (s21) at (0.824,7.763) {21};
  \node[int,minimum size=0.43cm,font=\scriptsize] (usA) at (10.435,1.012) {US};
  \node[int,minimum size=0.46cm,font=\scriptsize] (usB) at (9.220,1.215) {US};
  \node[int,minimum size=0.43cm,font=\scriptsize] (usC) at (9.626,1.998) {US};
  \node[int,minimum size=0.46cm,font=\scriptsize] (usD) at (10.665,1.863) {US};
  \node[int,minimum size=0.46cm,font=\scriptsize] (usE) at (10.665,2.849) {US};
  \node[int,minimum size=0.49cm,font=\scriptsize] (usF) at (8.613,3.375) {US};
  \node[int,minimum size=0.43cm,font=\scriptsize] (cr) at (10.503,3.780) {CR};
  \node[int,minimum size=0.46cm,font=\scriptsize] (usG) at (9.409,4.239) {US};
  \node[int,minimum size=0.49cm,font=\scriptsize] (fr) at (8.114,4.415) {FR};
  \node[int,minimum size=0.46cm,font=\scriptsize] (usH) at (6.817,3.969) {US};
  \node[int,minimum size=0.43cm,font=\scriptsize] (usI) at (6.210,3.280) {US};
  \node[int,minimum size=0.46cm,font=\scriptsize] (ae) at (7.155,1.958) {AE};
  \node[int,minimum size=0.46cm,font=\scriptsize] (in) at (5.332,1.755) {IN};
  \node[int,minimum size=0.46cm,font=\scriptsize] (cnA) at (3.821,2.200) {CN};
  \node[int,minimum size=0.49cm,font=\scriptsize] (cnB) at (4.455,2.929) {CN};
  \node[int,minimum size=0.46cm,font=\scriptsize] (cnC) at (2.498,3.740) {CN};
  \node[int,minimum size=0.49cm,font=\scriptsize] (cnD) at (1.593,4.752) {CN};
  \node[int,minimum size=0.43cm,font=\scriptsize] (be) at (2.713,5.090) {BE};
  \node[int,minimum size=0.43cm,font=\scriptsize] (plA) at (4.887,4.590) {PL};
  \node[int,minimum size=0.43cm,font=\scriptsize] (sk) at (4.199,4.968) {SK};
  \node[int,minimum size=0.46cm,font=\scriptsize] (tr) at (5.859,4.684) {TR};
  \node[int,minimum size=0.62cm,font=\small] (deA) at (6.156,5.886) {DE};
  \node[int,minimum size=0.43cm,font=\scriptsize] (deB) at (3.402,5.819) {DE};
  \node[int,minimum size=0.62cm,font=\small] (itA) at (8.343,5.899) {IT};
  \node[int,minimum size=0.62cm,font=\small] (itB) at (9.990,5.981) {IT};
  \node[int,minimum size=0.46cm,font=\scriptsize] (chA) at (11.448,6.588) {CH};
  \node[int,minimum size=0.46cm,font=\scriptsize] (chB) at (12.569,6.628) {CH};
  \node[int,minimum size=0.59cm,font=\small] (itC) at (9.139,6.899) {IT};
  \node[int,minimum size=0.59cm,font=\small] (itD) at (10.287,7.101) {IT};
  \node[int,minimum size=0.43cm,font=\scriptsize] (gbA) at (5.562,6.655) {GB};
  \node[int,minimum size=0.46cm,font=\scriptsize] (plB) at (6.696,6.615) {PL};
  \node[int,minimum size=0.65cm,font=\small] (hkA) at (3.308,7.047) {HK};
  \node[int,minimum size=0.43cm,font=\scriptsize] (bh) at (6.197,7.101) {BH};
  \node[int,minimum size=0.43cm,font=\scriptsize] (hkB) at (1.998,7.492) {HK};
  \node[int,minimum size=0.43cm,font=\scriptsize] (itE) at (7.897,7.803) {IT};
  \node[int,minimum size=0.46cm,font=\scriptsize] (vn) at (2.443,8.194) {VN};
  \node[int,minimum size=0.94cm,font=\large] (deC) at (5.090,7.803) {DE};
  \node[int,minimum size=0.94cm,font=\large] (itF) at (6.615,8.316) {IT};
  \node[int,minimum size=1.00cm,font=\large] (itG) at (8.033,9.072) {IT};
  \node[int,minimum size=0.81cm,font=\large] (usJ) at (5.062,9.166) {US};
  \node[int,minimum size=0.84cm,font=\large] (usK) at (3.793,9.787) {US};
  \node[int,minimum size=0.57cm,font=\small] (usL) at (4.806,10.192) {US};
  \node[int,minimum size=0.46cm,font=\scriptsize] (gbB) at (2.862,9.166) {GB};
  \node[int,minimum size=0.46cm,font=\scriptsize] (gbC) at (1.458,9.355) {GB};
  \node[man,minimum size=1.24cm,font=\Large] (fae) at (9.787,8.532) {FAE};
  \draw[arc] (s4) -- (usB);
  \draw[arc] (s4) -- (usA);
  \draw[arc] (s6) -- (usD);
  \draw[arc] (s18) -- (usE);
  \draw[arc] (s14) -- (cr);
  \draw[arc] (s11) -- (cnA);
  \draw[arc] (s8) -- (cnC);
  \draw[arc] (s19) -- (cnD);
  \draw[arc] (s17) -- (cnD);
  \draw[arc] (s7) -- (cnD);
  \draw[arc] (s9) -- (cnD);
  \draw[arc] (s20) -- (usI);
  \draw[arc] (s12) -- (plA);
  \draw[arc] (s15) -- (sk);
  \draw[arc] (s16) -- (plB);
  \draw[arc] (s10) -- (hkA);
  \draw[arc] (s21) -- (hkB);
  \draw[arc] (s13) -- (vn);
  \draw[arc] (s1) -- (gbC);
  \draw[arc] (s1) -- (gbB);
  \draw[flow] (s3) -- (usK);
  \draw[arc] (s3) -- (usL);
  \draw[arc] (s2) -- (usJ);
  \draw[arc] (s5) -- (chB);
  \draw[arc] (usA) -- (usB);
  \draw[arc] (usB) -- (usC);
  \draw[arc] (usB) -- (usF);
  \draw[arc] (usC) -- (usF);
  \draw[arc] (usD) -- (usF);
  \draw[arc] (usD) -- (usG);
  \draw[arc] (usE) -- (usF);
  \draw[arc] (cr) -- (usF);
  \draw[arc] (usF) -- (fr);
  \draw[arc] (usF) -- (usH);
  \draw[arc] (usG) -- (fr);
  \draw[arc] (ae) -- (fr);
  \draw[arc] (in) -- (ae);
  \draw[arc] (cnB) -- (in);
  \draw[arc] (cnA) -- (cnB);
  \draw[arc] (cnB) -- (tr);
  \draw[arc] (cnC) -- (cnB);
  \draw[arc] (cnC) -- (cnD);
  \draw[arc] (usI) -- (usH);
  \draw[arc] (usH) -- (deA);
  \draw[arc] (tr) -- (fr);
  \draw[arc] (cnD) -- (be);
  \draw[arc] (be) -- (deA);
  \draw[arc] (cnD) -- (hkA);
  \draw[arc] (plA) -- (deA);
  \draw[arc] (sk) -- (deA);
  \draw[arc] (deB) -- (deA);
  \draw[arc] (hkA) -- (deB);
  \draw[arc] (hkA) -- (tr);
  \draw[arc] (hkA) -- (deA);
  \draw[arc] (hkA) -- (gbA);
  \draw[arc] (gbA) -- (deA);
  \draw[arc] (hkA) -- (bh);
  \draw[arc] (bh) -- (deA);
  \draw[arc] (bh) -- (itE);
  \draw[arc] (itE) -- (itA);
  \draw[arc] (deA) -- (itA);
  \draw[arc] (itA) -- (itC);
  \draw[arc] (fr) -- (itB);
  \draw[arc] (chA) -- (itB);
  \draw[arc] (chB) -- (chA);
  \draw[arc] (itB) -- (itD);
  \draw[arc] (plB) -- (deA);
  \draw[arc] (hkB) -- (hkA);
  \draw[arc] (vn) -- (hkA);
  \draw[arc] (hkA) -- (deC);
  \draw[arc] (gbB) -- (deC);
  \draw[arc] (gbC) -- (gbB);
  \draw[flow] (deC) -- (itF);
  \draw[flow] (itF) -- (itG);
  \draw[flow] (usJ) -- (deC);
  \draw[arc] (usK) -- (usJ);
  \draw[arc] (usL) -- (usJ);
  \draw[arc] (itC) -- (fae);
  \draw[arc] (itD) -- (fae);
  \draw[flow] (itG) -- (fae);
\end{tikzpicture}
\caption[Supply Chain Graph constructed from FAE Technology S.p.A. order-level dataset]{Supply Chain Graph constructed from FAE Technology S.p.A. order-level dataset. Following the same colour scheme as Figure~\ref{fig:2}. Vertex and edge size are proportional to the normalized flow associated with the corresponding node or edge. Country codes label intermediate vertices according to their geographic location. The graph comprises \(|S| = 21\) supplier sites and \(|I| = 44\) intermediate vertices, connected by \(\left| A_{SI} \right| = 24\) supplier-to-intermediate arcs, \(\left| A_{II} \right| = 55\) intermediate-to-intermediate arcs, and \(\left| A_{IM} \right| = 3\) intermediate-to-manufacturer arcs.}
\label{fig:11}
\end{figure}

The SCGM is constructed from the order-level dataset and the associated tracking information through an incremental procedure. For each delivered order, the corresponding supplier is matched to an existing vertex or, if none exists, a new supplier vertex is added to the graph. Similarly, for each consecutive pair of tracking events, an arc is created if not already present, with its carrier-specific flow initialized to zero; if the arc already exists, its flow is incremented by one unit. The transition probability matrix, representing the carrier-conditioned routing probabilities, is then updated to reflect the revised flow counts, as illustrated in the Supply Chain Graph Model section. This procedure incrementally maintains both the structural and probabilistic properties of the Supply Chain Graph as new shipment data is ingested. The resulting Supply Chain Graph, constructed from the dataset discussed in the Order-level Dataset Preparation section, is visualized in Figure~\ref{fig:11}.

\subsection{DRT Model Construction}\label{drt-model-construction}

As defined in the Dynamic Route Time (DRT) section, \(\text{DRT}_{\mu}(r,t) = f_{\mathrm{ML}}\left( \mathbf{x}(r,t) \right)\), where a supervised regression model (ML) is employed to predict the travel time of a route segment from a feature vector \(\mathbf{x}(r,t)\), with the route transit time as the model output. The following subsections describe, in order, the construction and augmentation of the route-level dataset, the feature engineering, and the selection, training and evaluation of the machine learning model.

\subsubsection{Route-level Dataset Construction and Augmentation}\label{route-level-dataset-construction-and-augmentation}

The route-level dataset is derived from the order-level dataset, whereas each sample corresponds to a single route segment traversed by a shipment, whose feature vector components is illustrated in Table~\ref{tab:2}.

\begin{table}[htbp]
\centering
\caption[Feature vector components for a single route-level dataset sample]{Feature vector components for a single route-level dataset sample. Each i-th dataset sample is described by a collection of features \(x_{i}(r,t)\) including the historical RT (\texttt{rt}), the current and averaged TMI and WMI (\texttt{tmi}, \texttt{avg\_tmi}, \texttt{wmi}, \texttt{avg\_wmi}), and the geodesic distance (\texttt{d}) between the route's origin and destination vertices.}\label{tab:2}
\small
\begin{tabular}{@{}lll@{}}
\toprule
Component & Unit & Description \\
\midrule
\texttt{rt} & {[}h{]} or {[}d{]} & Route Time indicator \(\text{RT}_{\mu}(r)\) \\
\texttt{tmi} & {[}0,1{]} & TMI computed for the route \\
\texttt{avg\_tmi} & {[}0,1{]} & Averaged TMI over observations on the route \\
\texttt{wmi} & {[}0,1{]} & WMI computed for the route \\
\texttt{avg\_wmi} & {[}0,1{]} & Averaged WMI over observations on the route \\
\texttt{d} & {[}km{]} & Geodesic distance between route vertices \\
\bottomrule
\end{tabular}
\end{table}

Starting from 483 samples, data cleaning removed 42, leaving with 441 samples for analysis. The dataset was split into training, validation and test sets with the ratio 70 : 10 : 20. The training set of 307 samples and test set of 89 samples was deemed limited for reliable statistical evaluation of model performance, thus a data augmentation strategy was employed by merging consecutive route segments belonging to the same path, producing samples that represent longer composite routes, but preserving temporal coherence. Traffic and weather features were aggregated proportionally, assuming local homogeneity over short spatial and temporal spans. Augmentation was applied twice to the training set. The first iteration consisted of merging pairs of consecutive routes, while the second iteration consisted of merging triples of consecutive routes. The resulting dataset versions are summarized in Table~\ref{tab:3}.

\begin{table}[htbp]
\centering
\caption[Data augmentation iterations for the route-level dataset]{Data augmentation iterations for the route-level dataset. Each version's size is obtained by adding its new samples to the size of the previous version. New samples are added exclusively to the training split, while the validation and test splits remain fixed across all versions, resulting in ratios 82 : 6 : 12 and 86 : 5 : 9 respectively; the small validation set splitting is acceptable because hyperparameters were not tuned.}\label{tab:3}
\small
\begin{tabular}{@{}lrrrrr@{}}
\toprule
Name & Iteration & Size & Training & Validation & Test \\
\midrule
\texttt{ds\_0} & 0 & 441 & 307 & 45 & 89 \\
\texttt{ds\_1} & 1 & 765 & 631 & 45 & 89 \\
\texttt{ds\_2} & 2 & 991 & 857 & 45 & 89 \\
\bottomrule
\end{tabular}
\end{table}

\subsubsection{Feature Engineering and Representation}\label{feature-engineering-and-representation}

A feature correlation analysis was conducted by computing Spearman rank correlations between the deviations of the features from their historical averages and the deviations of the actual travel time from the RT. This procedure is suitable for non-normally distributed data and monotonic relationships. Signed deviations showed a negligible correlation with travel time deviations, indicating that the direction of variation is not the main driver. Considering absolute deviations, traffic variability exhibited a moderate correlation with travel time deviations (0.58), where weather variability remained minimal (0.1). This suggested that the traffic fluctuations primarily influence predictability, while weather has a limited standalone effect. Weather features were however retained, as machine learning may leverage on them in conjunction with other predictors to improve accuracy. To systematically assess feature relevance, four dataset configurations are constructed, as shown in Table~\ref{tab:4}, by excluding geographical information and traffic and weather indicators, resulting in twelve distinct training sets when combined with the augmentation iterations.

\begin{table}[htbp]
\centering
\caption[Feature availability across the four subset configurations (A-D)]{Feature availability across the four subset configurations (A-D). Each subset includes ($\bullet$ mark) or excludes (empty) geographical information (\texttt{d}) and aggregated historical traffic and weather indicators (\texttt{avg\_tmi}, \texttt{avg\_wmi}), to test their individual contribution to prediction accuracy.}\label{tab:4}
\small
\begin{tabular}{@{}ccccccc@{}}
\toprule
Subset & \texttt{rt} & \texttt{tmi} & \texttt{avg\_tmi} & \texttt{wmi} & \texttt{avg\_wmi} & \texttt{d} \\
\midrule
A & $\bullet$ & $\bullet$ & $\bullet$ & $\bullet$ & $\bullet$ & $\bullet$ \\
B & $\bullet$ & $\bullet$ & & $\bullet$ & & $\bullet$ \\
C & $\bullet$ & $\bullet$ & $\bullet$ & $\bullet$ & $\bullet$ & \\
D & $\bullet$ & $\bullet$ & & $\bullet$ & & \\
\bottomrule
\end{tabular}
\end{table}

The subsets were designed to test feature importance hypotheses: whether historical averages provide additional predictive power beyond current conditions (Subset A vs B); whether contribution of geographical information impacts prediction accuracy (Subset A vs C). Subset D represents a minimal feature set to assess baseline predictive capability using current traffic and weather conditions and historical route times.

\subsubsection{Regression Models Selection}\label{regression-models-selection}

A Linear Regression~\citep{ref17_Seber2003} model served as a baseline, quantifying how much of the variance in RT value can be explained by a simple additive combination of the available features. This approach allows to assess the effectiveness of the supervised regression models for the DRT indicator, as any performance gain from more complex models over this baseline indicates the presence of genuine nonlinear structure in the data. Four regression models were evaluated, spanning a progression in complexity to isolate the contribution of nonlinear and interaction effects. To this end, ensemble tree-based methods were chosen because they are particularly well suited to tabular datasets employed in this work, in contrast to alternative supervised learning approaches. Unlike distance-based or kernel-based methods (e.g., k-nearest neighbours, support vector regression), tree-based ensembles make no assumptions about the feature space, are insensitive to feature scale, and natively handle mixed feature types, eliminating the need for extensive preprocessing or encoding. Random Forest~\citep{ref18_Breiman2001} was included as a bagging-based ensemble: it trains many decision trees independently, each on a random subset of the data and features, and averages their predictions to reduce variance and improve robustness to noise. This allows it to capture nonlinear relationships and interactions between features without extensive tuning, at the cost of treating each tree's contribution independently. Gradient boosting methods, such as XGBoost~\citep{ref19_Chen2016} and LightGBM~\citep{ref20_Ke2017} build trees sequentially rather than averaging independent trees, with each new tree trained specifically to correct the residual errors of the previous ones. They offer a sequential, error-correcting structure that is typically better suited than bagging to extracting weak or subtle signals, such as those from TMI and WMI. In fact, the predictive value of those indicators may only emerge through interaction with other features rather than through a strong standalone correlation.

\subsubsection{Regression Models Training}\label{regression-models-training}

To mitigate overfitting, and given the limited size of the dataset, default configurations for training and limited hyperparameters tuning were used. Each model was trained on all the dataset variants without hyperparameter fine-tuning, with optimization reserved for the best-performing configuration identified afterward. This process resulted in
\begin{equation*}
48\ \{\text{models}\} = 4\left\{ \begin{array}{c}
\text{regression} \\
\text{models}
\end{array} \right\} \times 4\left\{ \begin{array}{c}
\text{feature} \\
\text{sets}
\end{array} \right\} \times 3\left\{ \begin{array}{c}
\text{augmentation} \\
\text{iterations}
\end{array} \right\}\,.
\end{equation*}
For XGBoost and LightGBM, early stopping based on validation performance was applied to determine the optimal number of boosting iterations. The training procedure, including baseline hyperparameter configurations illustrated in Appendix~\ref{appendix-H}, is available at the first author's GitHub repository for the MSc thesis~\citep{ref25_Volpato2025b,ref24_Volpato2025a}. Each model was trained on all training set versions of the route-level dataset, while always validating and testing on the same splits, ensuring that improvements could be attributed to augmentation and feature engineering rather than changes in evaluation data.

\subsubsection{Regression Model Evaluation Metrics}\label{regression-model-evaluation-metrics}

Performance was evaluated using three metrics: mean absolute error (MAE), root mean squared error (RMSE) and mean absolute percentage error (MAPE). This combination of metrics captures both absolute deviations and proportional errors across routes of varying lengths. A preliminary grid search identified the most promising dataset-model combinations, where XGBoost and Random Forests emerged as the top performers, with XGBoost consistently outperforming alternatives across feature subsets. Results of the model performance evaluation are listed in Table~\ref{tab:E1} and Table~\ref{tab:E2} on {}Appendix~\ref{appendix-E}. Consequently, XGBoost was selected for further evaluation, consisting of training two instances the on subsets A, B, and C, as the best performing ones. The first instance by using standard hyperparameters, and the second instance by automatically optimizing hyperparameters via an iterative search on the validation set. The basis of evaluation was conducted using 10-fold cross-validation, ensuring that augmented samples were included only in training folds to preserve independence. Despite this, statistical confidence was limited, due to the size of the dataset, as reflected in the relatively wide 95\% confidence intervals computed using the Student's t-distribution. Results of the cross-validation are listed in Table~\ref{tab:E3}, where they revealed noticeably worse performance than the initial test-set results, accompanied by wide confidence intervals; moreover, hyperparameter optimization frequently degraded rather than improved model performance. Together, these observations indicate that the test and validation sets are too small to provide fully reliable performance estimates. Therefore, the final model was selected as a pragmatic choice: XGBoost trained on the full feature set (Subset A), without hyperparameter optimization. This choice favours practical robustness over performance: the full feature avoids architectural changes to the DRT model construction pipeline when additional data becomes available, and skipping hyperparameter tuning avoids overfitting to an unreliably small validation set.

\subsubsection{SCGM and LCDIs Hyperparameters}\label{scgm-and-lcdis-hyperparameters}

Configuration tables have been created to manage all hyperparameters for the SCGM and LCDIs, including the DRT model, and a representation of the database schema is available at the first author's GitHub repository for the MSc thesis~\citep{ref25_Volpato2025b}. Since the repository documents the complete system developed for the thesis, it also includes components and terminology not discussed in this paper; here, the authors report only the hyperparameter settings most relevant to this work's results:

\paragraph{SCGM Path Extraction} In a fully connected DAG with \(V\) vertices, the number of distinct paths between a source and a sink vertex can reach \(O\left( 2^{V - 2} \right)\) in the worst case. To keep computation tractable, only a limited number of the most probable paths is retained, rather than enumerating the entire path space. The corresponding path probabilities are rescaled to form a valid probability distribution over a restricted set, providing a practical heuristic to mitigate the combinatorial growth of path enumeration while preserving the most relevant routing behaviour.

\paragraph{LCDI Confidence Thresholds} Historical and real-time indicators can be evaluated with a form comprising two-sided confidence intervals \(\beta\), one for each indicator, which allows to adjust sensitivity and robustness of the estimations.

\paragraph{DRT Model Usage} Retrieving traffic and weather data incurs in computational and monetary costs, due to the number of requests required, as external services charge beyond certain usage thresholds. Thus, the DRT machine learning model is applied only for short-term estimations close to the shipment start time, also because forecasts become unreliable further into the future, and for the most probable extracted paths.

\paragraph{TFST Weighting Factor} Between the options of this functional (constant, linear, or polynomial), it is possible to optimize the polynomial function by an exponent parameter \(p \in \lbrack 0,1\rbrack\) by minimizing the error function MSE between the historical values of remaining transit times \(\left( t_{n}(s,c) - t \right)\) with the \(\text{TFST}_{\mu}(s,c,t;\alpha)\). Preliminary evaluation on a sample shipment showed that high variability in shipment times justifies assigning a lower weight to transit time, favouring predictive estimates. This approach can be extended to other site-carrier pairs as more data become available, allowing to reform the functional form of the weighting factor. Given the limited dataset, this optimization was for a single site-carrier pair (Site 3, Carrier UPS, with 53 orders), yielding an optimal value of \(p\) equal to 0.1809, leaving the initialized value of \(p\) equal to 0.5 for all other site-carrier pairs

\section{Results}\label{results}

The experimental evaluation is organized on two levels. The first evaluation concerns the DRT model, already introduced in the Regression Model section, where the machine learning model is first trained and assessed in isolation on the task of predicting the travel time of a single route segment, and compared against the historical baseline, as the pointwise RT which returns the historical mean route time. The second evaluation concerns the estimated overall delivery time across 7 model variants by means of 3 interval-based metrics.

\subsection{DRT Model Evaluation}\label{drt-model-evaluation}

This section reports the main findings from the evaluation described in the Regression Model section. The selected configuration for the DRT model is reported in Table~\ref{tab:5}.

\begin{table}[htbp]
\centering
\caption[Selected configuration for the DRT Model implementation]{Selected configuration for the DRT Model implementation. This table summarizes the chosen regression model, feature subset, augmentation level, and hyperparameter setting, as determined by the evaluation method.}\label{tab:5}
\small
\begin{tabular}{@{}ll@{}}
\toprule
Object & Description \\
\midrule
Model & XGBoost \\
Featurized Dataset & Full Feature Set, Subset A from Table~\ref{tab:4} \\
Data Augmentation & Single Iteration, \texttt{ds\_1} from Table~\ref{tab:3} \\
Hyperparameter Optimization & Default, from Table~\ref{tab:H2} \\
\bottomrule
\end{tabular}
\end{table}

The model evaluation produced a reduction in MAE from 1.60 to 1.12 days and nearly halved the MAPE, from 0.77\% to 0.44\% relative to the historical route time baseline, as detailed in Table~\ref{tab:E3}. The wide confidence intervals reported in the results reflect the limited size of the available dataset. Hyperparameter optimization was deliberately avoided, as it degraded performance on the small validation folds, which the authors infer it to be likely due to overfitting on limited data. In particular, these two values are also used as the relative margins \(\gamma\) of the interval-based DRT in the following subsection.

\subsection{Delivery Time Evaluation}\label{delivery-time-evaluation}

The evaluation follows a staged methodology. A time granularity for the calculation of LCDIs is first defined, after which a collection of seven model variants is explored. The first five variants are tested to evaluate the best-performing configuration of the TFST weighting factor, while the remaining two are tested to evaluate the best-performing Route Time implementation feeding into the EODT. Performance is assessed through three complementary metrics, sharpness, coverage, and interval score~\citep{ref23_Gneiting2007}, defined in {}Appendix~\ref{appendix-F}, computed on both the training and test sets to select the best configuration, with complete results listed in Appendix~\ref{appendix-G}. A preliminary sensitivity analysis is then performed on the best-performing models to evaluate their calibration, followed by a visual inspection of the evolution of the estimates for a sample of five orders; both analyses are restricted to the test set.

\subsubsection{Estimations Granularity}\label{estimations-granularity}

All indicators are estimated from time data at hourly resolution, but results are reported at two complementary granularities. At the hourly granularity, metrics are computed directly on the model's raw predictions, subsequently expressed in days; at the daily granularity, predictions are first discretized to integer business days, and the metrics are then evaluated on this scale. By construction, daily intervals are wider than hourly ones, trading sharpness for higher empirical coverage: the comparison between the two granularities therefore captures the trade-off between precision, favoured by the narrower but riskier hourly intervals, and robustness, favoured by the wider but better-calibrated daily intervals. The daily granularity also better matches the operational needs of the industrial use case, which favours stable business-day estimates over hour-level detail.

\subsubsection{TFST Model Variants}\label{tfst-model-variants}

A summary of the model variant names and their description is provided in Table~\ref{tab:6}, wherein various functional form of the TFST weighting factor are presented, together with the implementation of the DRT Model. The authors note that, due to the definition of the different weighting factor implementations, versions v0.0, v1, and v2 produce identical estimates for the dispatch stage: for these variants, the weighting factor is always evaluated to 1 at this stage, assigning full weight to TT since the shipment has not yet started. Consequently, in all the following analysis, the dispatch stage results will be grouped together.

\begin{table}[htbp]
\centering
\caption[Overview of TFST model variants]{Overview of TFST model variants. Variants v0.0-v0.2 employ fixed values of the weighting factor \(\mathbf{\alpha}\), variants v1.0-v2.0 employ linear and polynomial functional of \(\alpha\), and variants v3.0-v3.1 replace the baseline Route Time with the DRT Model, using the ML-derived best MAPE, and the baseline MAPE respectively.}\label{tab:6}
\small
\begin{tabular}{@{}lll@{}}
\toprule
Variant & TFST \(\mathbf{\alpha}\) Factor & Description \\
\midrule
v0.0 & 1.0 & TT-only contribution to TFST \\
v0.1 & 0.0 & PT-only contribution to TFST \\
v0.2 & 0.5 & Equal TT and PT contribution to TFST \\
v1.0 & linear & Linear functional between TT and PT contribution to TFST \\
v2.0 & polynomial & Polynomial functional between TT and PT contribution to TFST \\
v3.0 & polynomial & Similar to v2.0 with DRT contribution to TFST (\(\gamma\) = 0.44) \\
v3.1 & polynomial & Similar to v2.0 with DRT contribution to TFST (\(\gamma\) = 0.77) \\
\bottomrule
\end{tabular}
\end{table}

\subsubsection{TFST Weighting Factor Performance}\label{tfst-weighting-factor-performance}

Model variants from v0.0 to v2.0 were evaluated in order to analyse the different TFST weighting factor implementations across training and test sets, for both the dispatch and shipment stages in Table~\ref{tab:G1} and Table~\ref{tab:G2} respectively. Dispatch stage estimates are shown to be less precise than shipment-stage estimates, with coverage below the nominal confidence level at both granularities, indicating insufficient reliability. For the shipment stage, hourly estimates remain unreliable, whereas daily estimates achieve coverage close to the nominal 90\%, indicating acceptable calibration on both training and test data. Among the weighting-factor variants, TT-dominated models tend to produce wider intervals: v0.0 achieves the highest coverage across all splits and granularities, but at the cost of a too wide interval score. Conversely, v0.1 and v0.2 yield narrower intervals and lower interval scores, at the expense of reduced coverage, with v0.1 performing best on the test set and v2.0 performing best on the training set. Given the limited amount of available data, version v2.0 is selected for the remainder of the evaluation, as it achieves the best overall balance between coverage and interval width.

\subsubsection{TFST Model Evaluation}\label{tfst-model-evaluation}

Model variants from v2.0 to v3.1 were evaluated in order to analyse the DRT contributions to the TFST, and by extension, to the EODT implementations across training and test sets, as shown in Table~\ref{tab:4}. Two main observations emerge from this comparison. First, isolating the contribution of the DRT model itself requires comparing v2.0 and v3.1, which share the same relative margin (\(\gamma\) = 0.77). On the test set, v3.1 achieves narrower intervals at both granularities, at the cost of slightly reduced coverage. This confirms that the DRT estimate alone produces tighter predictions than the historical baseline. Second, when the ML-derived margin is also applied (v3.0, \(\gamma\) = 0.44), sharpness improves further and the interval score reaches its best value across all variants, despite a drop in coverage. Overall, the interval score consistently favours the ML-based approach, owing to its tighter intervals, although confidence intervals remain wide due to the limited sample size; additional DRT strategies should be explored as more data become available.

\begin{table}[htbp]
\centering
\caption[Performance comparison of model variants v2.0-v3.1]{Performance comparison of model variants v2.0-v3.1. Bold values indicate the variant with lower metric values obtained for the test set, evaluated at both granularities.}\label{tab:7}
\small
\begin{tabular}{@{}llccc@{}}
\toprule
\multirow{2}{*}{Set} & \multirow{2}{*}{Variant} & Sharpness & Coverage & Interval Score \\
& & Hourly / Daily & Hourly / Daily & Hourly / Daily \\
\midrule
\multirow{3}{*}{Train} & v2.0 & 1.03 / 1.18 & 0.52 / 0.84 & 1.45 / 1.63 \\
& v3.0 & 0.87 / 0.91 & 0.35 / 0.76 & 1.37 / 1.54 \\
& v3.1 & 0.99 / 1.11 & 0.50 / 0.82 & 1.41 / 1.59 \\
\midrule
\multirow{3}{*}{Test} & v2.0 & 1.30 / 1.55 & 0.63 / 0.86 & 1.99 / 2.20 \\
& \textbf{v3.0} & \textbf{1.11 / 1.25} & \textbf{0.44 / 0.81} & \textbf{1.92 / 2.11} \\
& v3.1 & 1.23 / 1.45 & 0.55 / 0.83 & 1.96 / 2.23 \\
\bottomrule
\end{tabular}
\end{table}

\subsubsection{Sensitivity Analysis}\label{sensitivity-analysis}

\begin{figure}[htbp]
\centering
\includegraphics[width=0.8\linewidth]{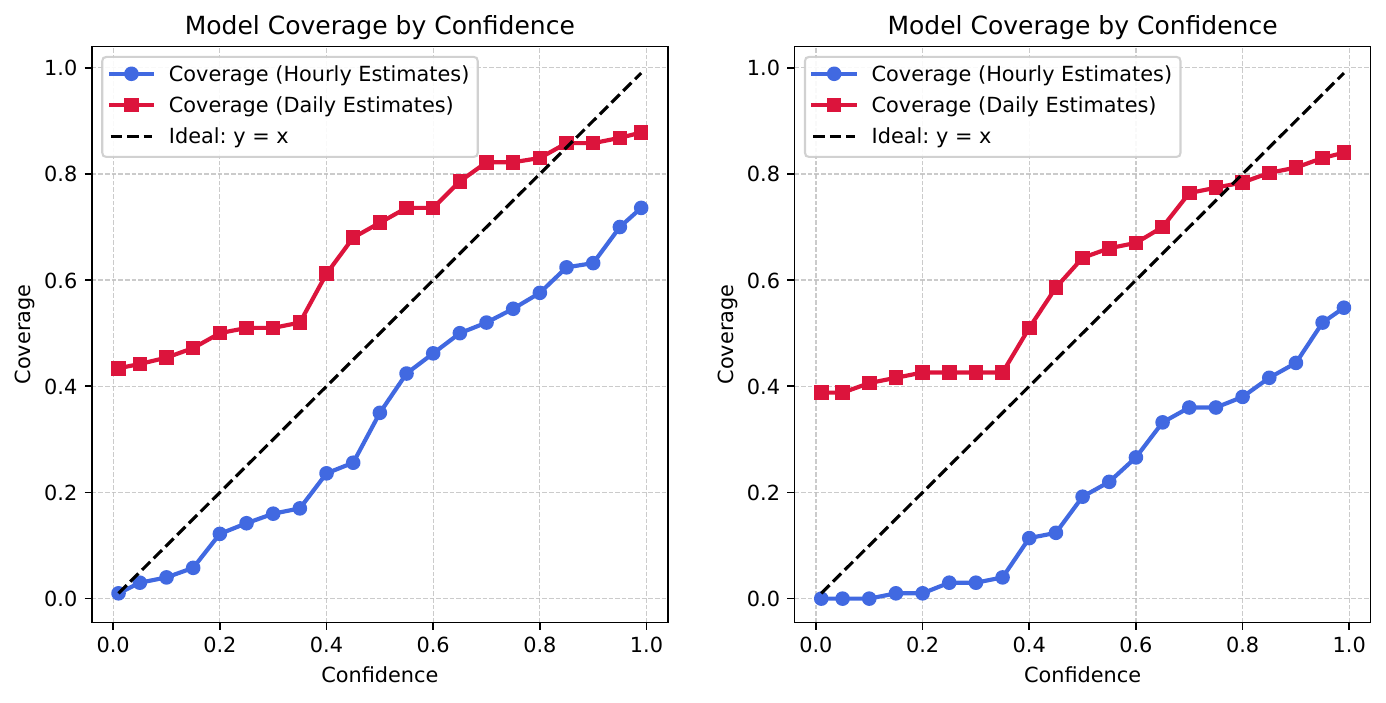}
\caption[Observed coverage against nominal confidence for variants v2.0 and v3.0]{Observed coverage against nominal confidence for variants v2.0 and v3.0. Both models have been evaluated on the test set at hourly and daily granularities. The dashed line indicates the ideal calibration (observed coverage equal to nominal confidence).}
\label{fig:12}
\end{figure}

A preliminary sensitivity analysis was performed on variants v2.0 and v3.0 to evaluate their calibration on the test set, whether the stated confidence level of each predicted interval matches its actual empirical reliability, a property distinct from sharpness and essential for the estimates to be trustworthy in operational decision-making. Figure~\ref{fig:12} reports the relationship between the nominal confidence level and the corresponding observed coverage, for both variants and at both hourly and daily estimation granularities.

Both variants exhibit similar behaviour, where hourly estimates fall well below the ideal trend in both cases, indicating unreliability, although the resulting curve remains roughly parallel to the ideal line. Daily estimates maintain a baseline coverage of approximately 0.4 at low nominal confidence levels, to then dropping below the ideal line beyond a nominal confidence of 0.8, indicating some under-reliability past this point. Overall, both variants have been shown to be unreliable at the hourly granularity, due to consistently low coverage, but provide meaningful insights at the daily level, achieving a reasonable balance between precision and calibration and remaining suitable for decision-making despite some residual under-reliability at higher confidence levels.

\subsubsection{EODT Estimation Results}\label{eodt-estimation-results}

\begin{figure}[htbp]
\centering
\includegraphics[width=0.8\linewidth]{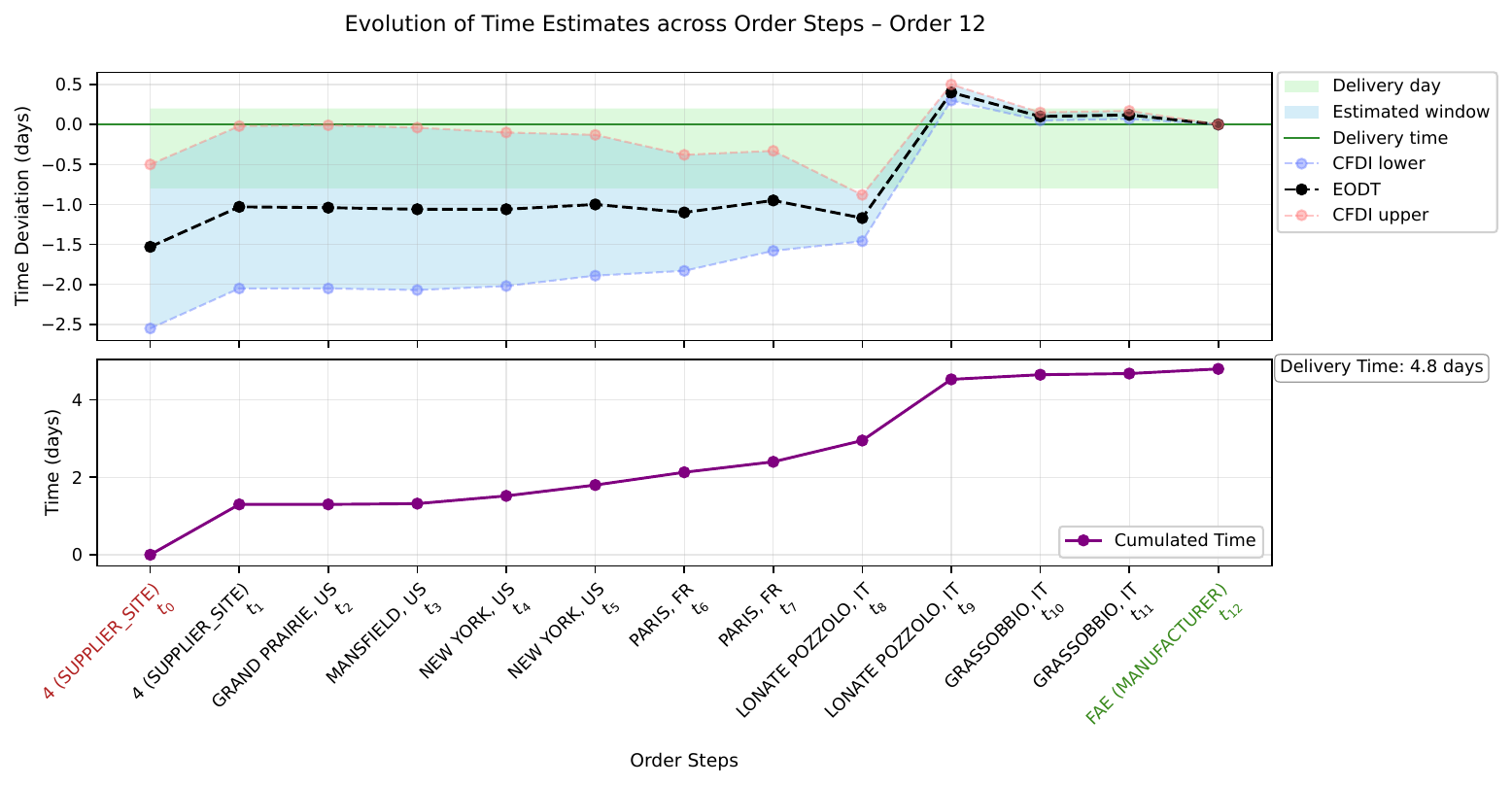}\\[4pt]
\includegraphics[width=0.8\linewidth]{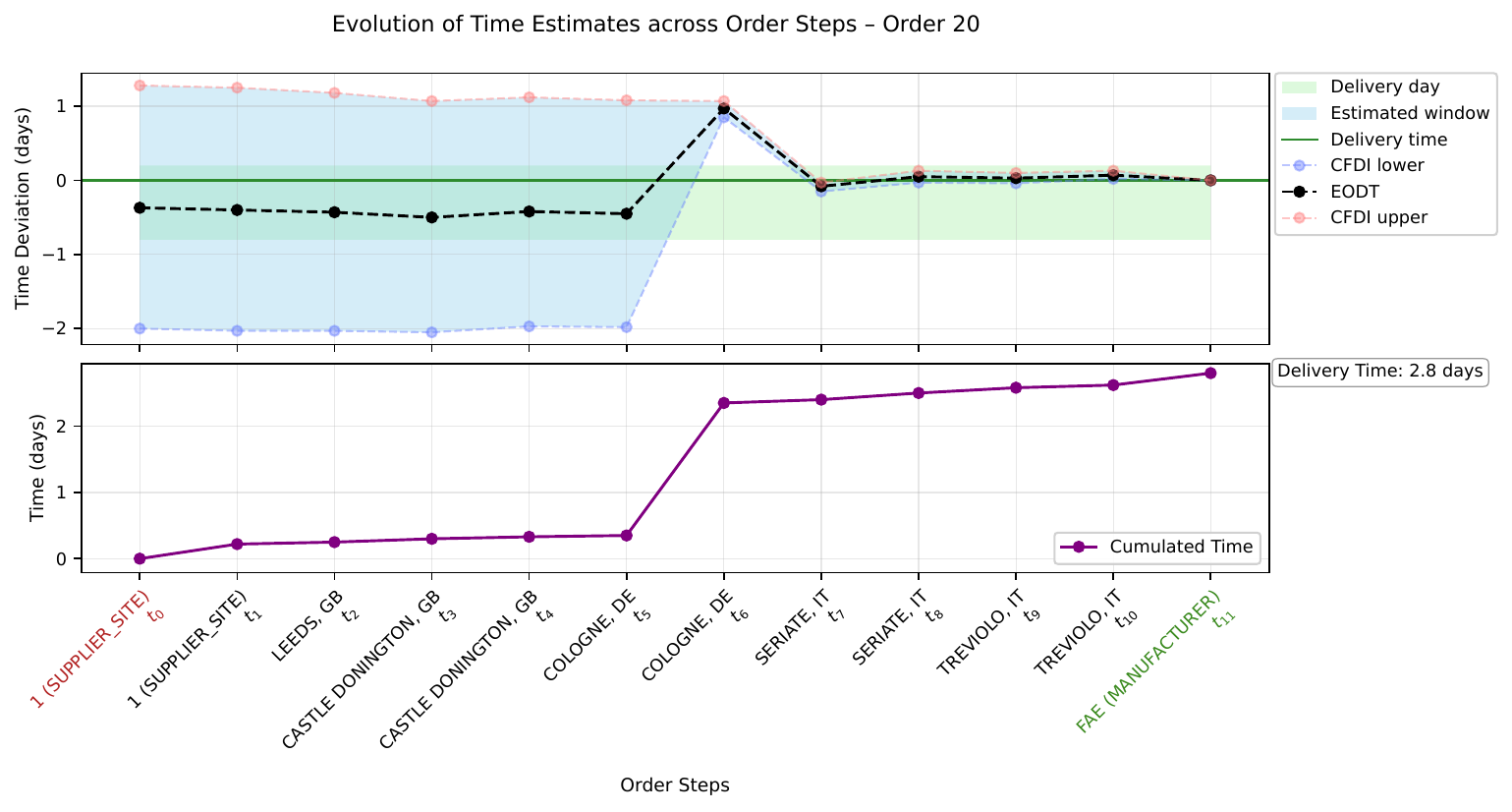}
\caption[Evolution of EODT estimates for supplier Site 1 (Order 20) and Site 4 (Order 12)]{Evolution of EODT estimates for supplier Site 1 (Order 20) and Site 4 (Order 12). Order 20's delivery time is initially underestimated, with the actual delivery falling outside the predicted interval, until a sharp correction occurs near the end of the shipment, after which the estimate becomes accurate and well-calibrated through delivery. Order 12's delivery time falls well within EODT bounds until a specific point, after which estimates are well-calibrated through delivery.}
\label{fig:13}
\end{figure}

\begin{figure}[htbp]
\centering
\includegraphics[width=0.8\linewidth]{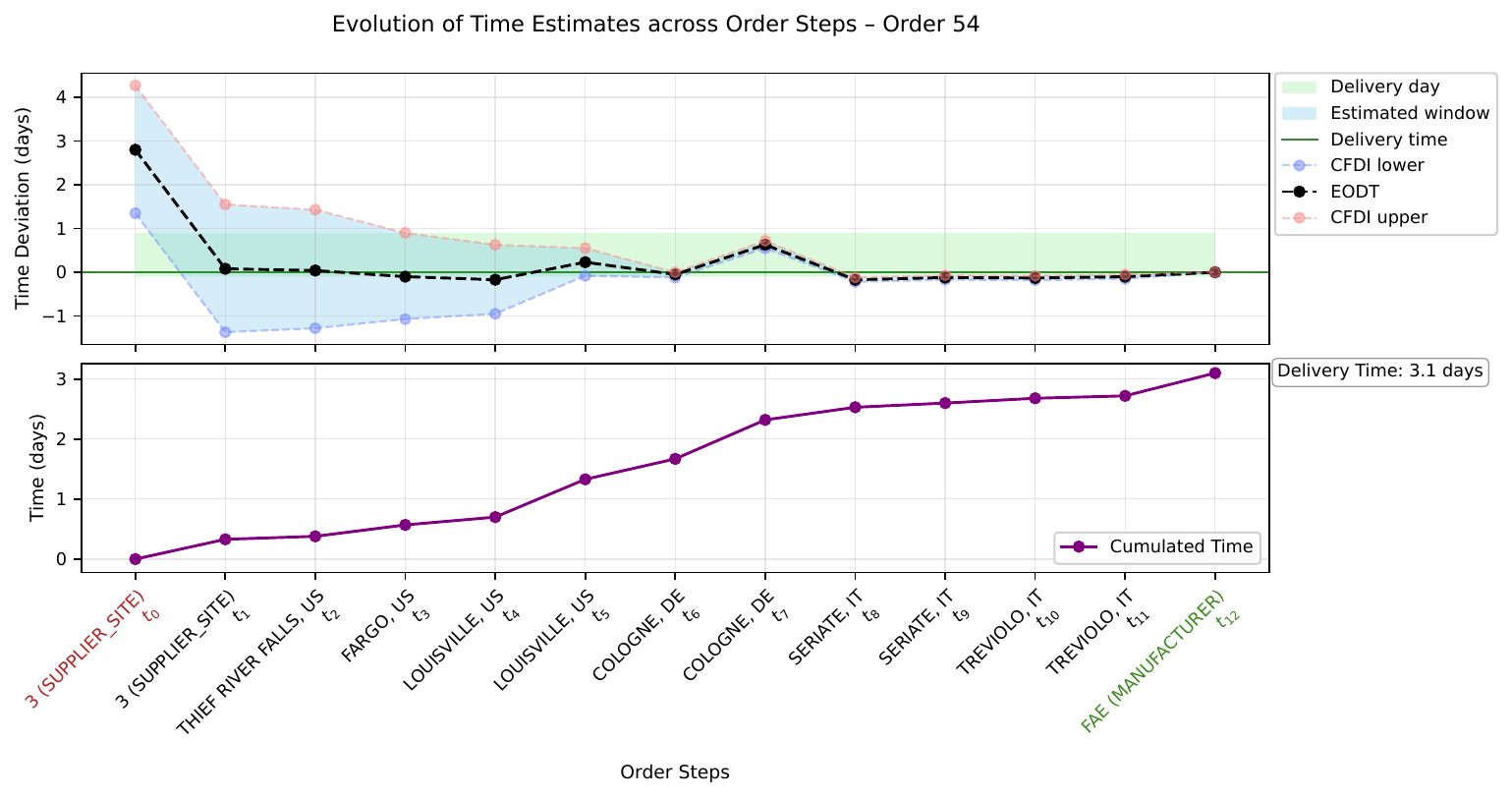}\\[4pt]
\includegraphics[width=0.8\linewidth]{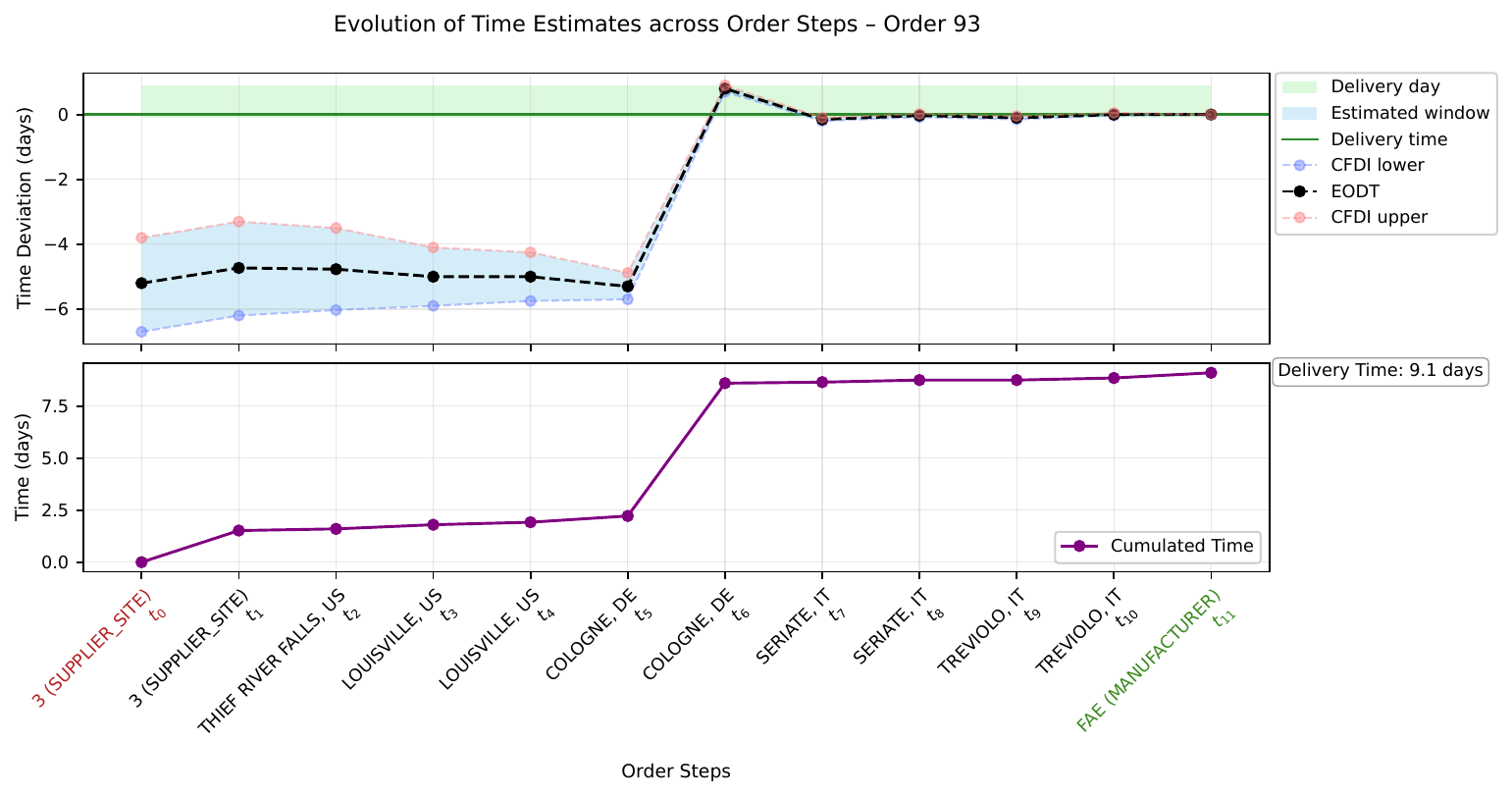}\\[4pt]
\includegraphics[width=0.8\linewidth]{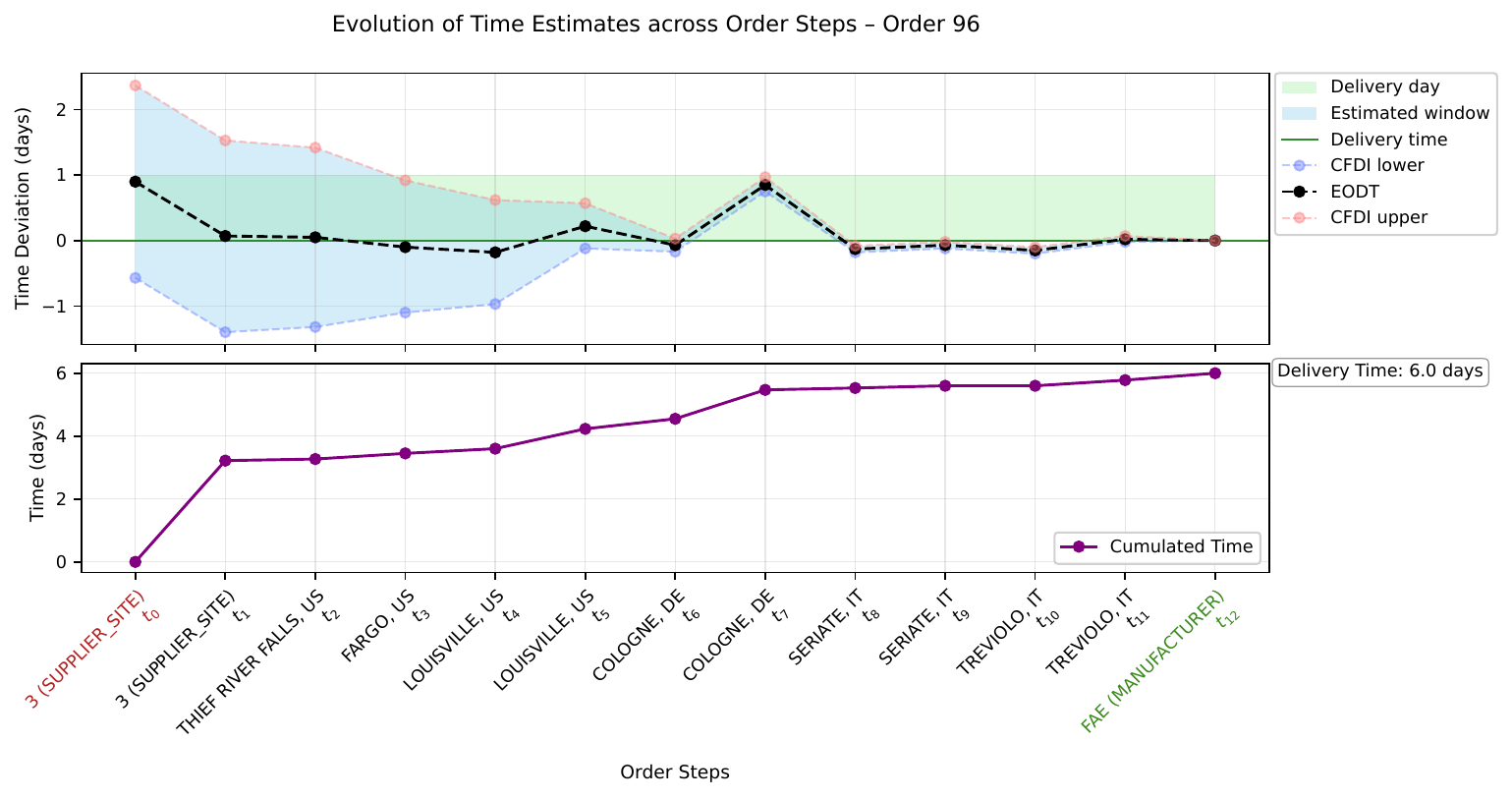}
\caption[Evolution of EODT estimates for supplier Site 3 (Orders 54, 93, and 96)]{Evolution of EODT estimates for supplier Site 3 (Orders 54, 93, and 96). Order 54 and Order 96 start noticeably overestimated, with the actual delivery time falling near or outside the EODT upper bound, before converging to accurate, well-calibrated estimates. Order 93 instead starts strongly underestimated, remaining outside the EODT interval until a sharp correction at the same point, after which it becomes accurate and well-calibrated through delivery.}
\label{fig:14}
\end{figure}

This subsection illustrates the evolution of the time estimates produced by the variant v3.0, which achieved the best interval score, for a small sample of five orders selected from the test set, ensuring that at least one order from each site is represented. Each figure is organized into two plots. The lower plot illustrates the actual shipment time progression, with locations shown on the x-axis and total elapsed time on the y-axis; the first data point corresponds to the order placement, representing the dispatch-stage estimate, while subsequent points correspond to the shipment steps registered by carrier tracking. The upper plot shows the deviation of the interval-based EODT from the actual delivery time, including the trends of the lower and upper bound estimates as described in Figure~\ref{fig:9}, labelled CFDI, a separate nomenclature adopted within the M4ESTRO project. {}Figure~\ref{fig:13} shows the shipment dynamics for two orders delivered from supplier sites 1 and 4, each associated with five orders in total, while Figure~\ref{fig:14} illustrates the evolution of three orders from site 3, which processed the highest number of orders overall. The related data regarding order-level information of supplier and site identifiers is reported in the corresponding Table~\ref{tab:B2}.

Overall, the plot highlights the effectiveness of the EODT estimates, with most orders closely tracking the actual shipment times. Three key observations emerge. First, regarding general accuracy, the model performs well across all visualized orders, with Order 93 being the only case showing significant deviations during the first half of the shipment; even in this case, however, the estimates quickly converge to the correct values. Second, regarding the convergence of uncertainty, a clear narrowing of the prediction interval occurs around the midpoint of the shipment in each plot, likely reflecting the reduced number and length of the remaining paths considered by the PT indicator as the shipment progresses. Third, the DRT model demonstrates the ability to adapt to real-time delay events, visible as steep segments in the underlying cumulated time plots.

\section{Discussion}\label{discussion}

The results from the use case implementation of the SCGM and LCDI framework demonstrate their effectiveness in predicting delivery times and, by extension, providing early-warning to potential delays along inbound logistics processes, addressing the real-time disruption detection requirement set out by the M4ESTRO project. The results support three main conclusions, discussed below: the validity of the LCDI framework as an implementation-independent formalism, the performance and trade-offs of the specific statistical and machine learning techniques adopted, and the practical relevance of the approach for industrial deployment. LCDIs were defined to be independent of their implementation: they specify the quantities of interest, such as transit, path, and forecasted shipment times, without prescribing the computational techniques used to estimate them. In this work, implementation choices were driven by data availability and project time constraints, resulting in a combination of statistical estimators and the XGBoost machine learning regression model for the DRT indicator. The power of the framework lies in its generality and potential to accommodate any advanced or alternative estimation techniques as additional data or computational resources become available. The SCGM effectively captures structural dependencies and delay-propagation dynamics along the order fulfilment process, and indicators derived jointly from graph properties and operational data, such as path structure, offer explanatory power that extends beyond traditional operational features. Even with limited data, the DRT model outperformed the historical route-time baseline, reducing the mean absolute error from 1.60 to 1.12 days and nearly halving the MAPE, particularly after the application of data augmentation. Comparable improvements were observed for the TFST and EODT estimates, as the interval score consistently favoured the machine learning-based route time model over the historical baseline, confirming that the integration of real-time traffic and weather data yields more informative predictions, even though the resulting gains are still modest given the limited amount of available data. A recurring trade-off emerged between estimation granularities: hourly estimates offer finer temporal resolution but exhibit poor calibration, with coverage consistently falling below the nominal confidence level, whereas daily estimates, though less precise, achieve better calibration and align with the operational needs of the industrial use case, which favours stable business-day estimates over hour-level detail. Performance also varied across operational stages: data scarcity continues to limit the reliability of dispatch-stage estimates, while shipment-stage estimates already provide actionable insights, benefiting from the progressive reduction of uncertainty as an order advances toward delivery. The SCGM and LCDI framework has been embedded in a module of the M4SCARA application prototype. This prototype persists raw measurements and indicator configurations in a relational data model~\citep{ref25_Volpato2025b} and exposes the indicators through a serverless architecture deployed on AWS Lambda (computing service provided by Amazon Web Services), with dedicated components for historical and real-time indicator computation, supply chain graph management and route time estimation~\citep{ref24_Volpato2025a}. This integration confirms that the indicators can be computed continuously as new tracking events are ingested, enabling the early disruption detection capability targeted by the M4ESTRO project. The main limitations of this work stem from the size and composition of the available data. Results have been derived from a single manufacturer, with a single test set, and an order-level dataset strongly concentrated on a single site-carrier pair, which limits the extent to which the TFST weighting factor and the DRT Model can be generalized to other routes. Limited statistical significance testing was performed on the differences between model variants, and the wide cross-validation confidence intervals reported for the DRT model in Table~\ref{tab:E3} reflects the same constraint. Consequently, the results presented in this work are internally consistent and validated on an industrial use case.

\section{Future Work}\label{future-work}

Future work should prioritize advanced uncertainty modelling techniques to overcome the limitations of the current DRT interval estimation, which relies on a symmetric margin derived from a global MAPE and therefore does not account for route-specific error behaviour that may generate asymmetries. Quantile regression~\citep{ref22_Koenker1978} could address this directly, by estimating conditional quantiles of travel time as a function of route-level features and producing intervals that adapt to each route's error distribution rather than assuming a fixed, symmetric bound around pointwise estimates. The approach introduced in this work has proven effective in closely related travel-time settings: quantile regression combined with deep architectures has been used to build probabilistic travel-time predictors that model error without assuming a fixed distributional form~\citep{ref26_Li2020}, and monotonic quantile regression frameworks have been applied to bus arrival time prediction to generate adaptive, calibration-aware prediction windows under varying horizons~\citep{ref27_Parslov2023}. Bayesian methods~\citep{ref28_Pomykacz2024} could offer a complementary benefit: by treating model parameters as distributions rather than point estimates, they naturally propagate uncertainty from limited or noisy route-specific data into a prediction that can be incrementally updated as new shipment observations become available, directly addressing the small-sample constraint that currently limits route-specific probabilistic modelling in the DRT. In a food-delivery setting facing similar data scarcity, Bayesian generalized linear models estimated via Hamiltonian Monte Carlo were shown to produce reliable delivery-time predictions without requiring the large training sets typically needed by neural approaches~\citep{ref28_Pomykacz2024}, while a comparative study on bus travel time found that, although Bayesian neural networks are more computationally demanding, they can outperform deep quantile regression in specific cases, suggesting that the two approaches are complementary rather than competing~\citep{ref27_Parslov2023}. More broadly, calibrated uncertainty-aware delay prediction has already demonstrated operational value in logistics, where reliable prediction intervals have been shown to help planners identify shipments at risk of significant delay~\citep{ref29_Faulkner2026}. Together, these approaches would move the DRT indicator closer to the probabilistic interval definition it was originally intended to satisfy, improving both the reliability and interpretability of the uncertainty bounds it produces. Beyond uncertainty modelling, once more data become available, the TFST optimization approach should be validated across multiple site-carrier pairs to assess the generalizability of the current results, possibly refining the choice of the family of functions used to model the weighting factor. The SCGM-LCDI framework can also be readily extended to multi-manufacturer environments, enabling data aggregation from multiple sources within a unified graph structure, a shift that would also help relieve the data scarcity currently limiting the adoption of more sophisticated uncertainty models. Finally, comprehensive validation across diverse manufacturing contexts, supply chain topologies, and disruption scenarios will be essential to further refine the proposed approach and strengthen the evidence base supporting its operational deployment.

\clearpage

\section*{Data and Software Availability}\label{data-and-software-availability}

The data underlying this study consist of commercially sensitive logistics records provided by FAE Technology S.p.A., the data owner, under a confidentiality agreement, and therefore cannot be made publicly available. Requests to access anonymized extracts may be addressed to the corresponding author and will be evaluated with the data owner on a case-by-case basis. The source code of the modelling framework, together with the first author's thesis~\citep{ref01_Volpato2025} and its related supplementary material are available at \url{https://doi.org/10.5281/zenodo.18615069}. License details are reported in the repository.

\section*{Competing Interests}\label{competing-interests}

No competing interests were disclosed.

\section*{Ethics and Consent Statement}\label{ethics-and-consent-statement}

Ethical approval and consent were not required for this study. The research involved evaluating private supply chain procurement data using non-invasive technologies and did not involve human participants, personal data collection, animal subjects, or any procedures requiring ethical review. All data gathering activities were conducted with the authorization of the relevant data owners (FAE Technology S.p.A., Italy).

\section*{Declaration of Generative AI and AI-Assisted Technologies}\label{declaration-of-generative-ai-and-ai-assisted-technologies-in-the-writing-process}

During the preparation of this manuscript, the authors used generative AI-based tools, specifically large language models such as Claude Sonnet 5 (Anthropic), solely to assist with language editing and consistency checks in limited sections. All substantive decisions regarding the design of the review, screening, data extraction, analysis and interpretation were made by the authors. The authors carefully reviewed and edited all AI-assisted text and take full responsibility for the integrity and accuracy of the manuscript's content.

\section*{Author Contributions}

Mattia: Conceptualization, Methodology, Formal Analysis, Software, Writing - Original Draft Preparation. Andrea: Conceptualization, Methodology, Writing - Original Draft Preparation, Writing - Review \& Editing. Bruno: Project Administration, Supervision, Writing - Review \& Editing.

\begin{ack}\label{acknowledgements}
This work was supported by the Horizon Europe Research and Innovation Programme under grant agreement No.~101138506 (M4ESTRO, Industrial Manufacturing As a sErvice STRategies and models for flexible, resilient, and reconfigurable value networks through Trusted and Transparent Operations). The authors gratefully acknowledge FAE Technology S.p.A. for their partnership in pilot validation.
\end{ack}

\clearpage

\bibliographystyle{plainnat}
\bibliography{bibliografia}

\begin{thebibliography}{43}
\providecommand{\natexlab}[1]{#1}
\providecommand{\url}[1]{\texttt{#1}}
\expandafter\ifx\csname urlstyle\endcsname\relax
  \providecommand{\doi}[1]{doi: #1}\else
  \providecommand{\doi}{doi: \begingroup \urlstyle{rm}\Url}\fi

\bibitem[Aljohani(2023)]{ref03_Aljohani2023}
Abeer Aljohani.
\newblock {Predictive Analytics and Machine Learning for Real-Time Supply Chain
  Risk Mitigation and Agility}.
\newblock \emph{Sustainability}, 15\penalty0 (20):\penalty0 15088, 2023.
\newblock \doi{10.3390/su152015088}.

\bibitem[Ayrapetov et~al.(2025)Ayrapetov, Larin, Kapski, Kapski, Khakimov, and
  Otakuziyev]{ref13_Ayrapetov2025}
Dmitriy Ayrapetov, O.~N. Larin, D.~V. Kapski, P.~D. Kapski, R.~M. Khakimov, and
  D.~I. Otakuziyev.
\newblock {Quantifying Supply Chain Resilience: A Markov Chain Approach}.
\newblock In \emph{ICTEA: International Conference on Thermal Engineering:
  Theory and Applications}, volume~1. Toronto Metropolitan University, 2025.
\newblock URL
  \url{https://journals.library.torontomu.ca/index.php/ictea/article/view/2523}.

\bibitem[Balster et~al.(2020)Balster, Hansen, Friedrich, and
  Ludwig]{ref41_Balster2020}
Andreas Balster, Ole Hansen, Hanno Friedrich, and Andr{\'e} Ludwig.
\newblock {An ETA Prediction Model for Intermodal Transport Networks Based on
  Machine Learning}.
\newblock \emph{Business \& Information Systems Engineering}, 62\penalty0
  (5):\penalty0 403--416, 2020.
\newblock \doi{10.1007/s12599-020-00653-0}.

\bibitem[Baryannis et~al.(2019)Baryannis, Validi, Dani, and
  Antoniou]{ref39_Baryannis2019}
George Baryannis, Sahar Validi, Samir Dani, and Grigoris Antoniou.
\newblock {Supply chain risk management and artificial intelligence: state of
  the art and future research directions}.
\newblock \emph{International Journal of Production Research}, 57\penalty0
  (7):\penalty0 2179--2202, 2019.
\newblock \doi{10.1080/00207543.2018.1530476}.

\bibitem[Bennett and Lemoine(2014)]{ref35_Bennett2014}
Nathan Bennett and G.~James Lemoine.
\newblock {What a difference a word makes: Understanding threats to performance
  in a VUCA world}.
\newblock \emph{Business Horizons}, 57\penalty0 (3):\penalty0 311--317, 2014.
\newblock \doi{10.1016/j.bushor.2014.01.001}.

\bibitem[Bodendorf et~al.(2023)Bodendorf, Sauter, and
  Franke]{ref06_Bodendorf2023}
Frank Bodendorf, Maximilian Sauter, and J{\"o}rg Franke.
\newblock {A mixed methods approach to analyze and predict supply disruptions
  by combining causal inference and deep learning}.
\newblock \emph{International Journal of Production Economics}, 256:\penalty0
  108708, 2023.
\newblock \doi{10.1016/j.ijpe.2022.108708}.

\bibitem[Borucka et~al.(2023)Borucka, Koz{\l}owski, Parczewski, Antosz, Gil,
  and Pieniak]{ref10_Borucka2023}
Anna Borucka, Edward Koz{\l}owski, Rafa{\l} Parczewski, Katarzyna Antosz,
  Leszek Gil, and Daniel Pieniak.
\newblock {Supply Sequence Modelling Using Hidden Markov Models}.
\newblock \emph{Applied Sciences}, 13\penalty0 (1):\penalty0 231, 2023.
\newblock \doi{10.3390/app13010231}.

\bibitem[Breiman(2001)]{ref18_Breiman2001}
Leo Breiman.
\newblock {Random Forests}.
\newblock \emph{Machine Learning}, 45\penalty0 (1):\penalty0 5--32, 2001. 
\newblock \doi{10.1023/A:1010933404324}.

\bibitem[Brintrup et~al.(2020)Brintrup, Pak, Ratiney, Pearce, Wichmann,
  Woodall, and McFarlane]{ref40_Brintrup2020}
Alexandra Brintrup, Johnson Pak, David Ratiney, Tim Pearce, Pascal Wichmann,
  Philip Woodall, and Duncan McFarlane.
\newblock {Supply chain data analytics for predicting supplier disruptions: a
  case study in complex asset manufacturing}.
\newblock \emph{International Journal of Production Research}, 58\penalty0
  (11):\penalty0 3330--3341, 2020.
\newblock \doi{10.1080/00207543.2019.1685705}.

\bibitem[Carbonneau et~al.(2008)Carbonneau, Laframboise, and
  Vahidov]{ref02_Carbonneau2008}
R{\'e}al Carbonneau, Kevin Laframboise, and Rustam Vahidov.
\newblock {Application of machine learning techniques for supply chain demand
  forecasting}.
\newblock \emph{European Journal of Operational Research}, 184\penalty0
  (3):\penalty0 1140--1154, 2008.
\newblock \doi{10.1016/j.ejor.2006.12.004}.

\bibitem[Chen and Guestrin(2016)]{ref19_Chen2016}
Tianqi Chen and Carlos Guestrin.
\newblock {XGBoost: A Scalable Tree Boosting System}.
\newblock In \emph{Proceedings of the 22nd ACM SIGKDD International Conference
  on Knowledge Discovery and Data Mining (KDD '16)}, pages 785--794, New York,
  NY, USA, 2016. ACM.
\newblock \doi{10.1145/2939672.2939785}.

\bibitem[Christopher and Peck(2004)]{ref36_Christopher2004}
Martin Christopher and Helen Peck.
\newblock {Building the Resilient Supply Chain}.
\newblock \emph{The International Journal of Logistics Management}, 15\penalty0
  (2):\penalty0 1--14, 2004.
\newblock \doi{10.1108/09574090410700275}.

\bibitem[Faulkner et~al.(2026)Faulkner, Zandehshahvar, Akhlaghi, Ouellet,
  Jordan, and Van~Hentenryck]{ref29_Faulkner2026}
Stefan Faulkner, Reza Zandehshahvar, Vahid~Eghbal Akhlaghi, Sebastien Ouellet,
  Carsten Jordan, and Pascal Van~Hentenryck.
\newblock {Uncertainty-Aware Delivery Delay Duration Prediction via Multi-Task
  Deep Learning}.
\newblock arXiv preprint arXiv:2602.20271, 2026.
\newblock URL \url{https://arxiv.org/abs/2602.20271}.

\bibitem[Ford and Fulkerson(1956)]{ref15_Ford1956}
L.~R. Ford and D.~R. Fulkerson.
\newblock {Maximal Flow Through a Network}.
\newblock \emph{Canadian Journal of Mathematics}, 8:\penalty0 399--404, 1956.
\newblock \doi{10.4153/CJM-1956-045-5}.

\bibitem[Ghadge et~al.(2020)Ghadge, Wurtmann, and Seuring]{ref33_Ghadge2020}
Abhijeet Ghadge, Hendrik Wurtmann, and Stefan Seuring.
\newblock {Managing climate change risks in global supply chains: a review and
  research agenda}.
\newblock \emph{International Journal of Production Research}, 58\penalty0
  (1):\penalty0 44--64, 2020.
\newblock \doi{10.1080/00207543.2019.1629670}.

\bibitem[Gneiting and Raftery(2007)]{ref23_Gneiting2007}
Tilmann Gneiting and Adrian~E. Raftery.
\newblock {Strictly Proper Scoring Rules, Prediction, and Estimation}.
\newblock \emph{Journal of the American Statistical Association}, 102\penalty0
  (477):\penalty0 359--378, 2007.
\newblock \doi{10.1198/016214506000001437}.

\bibitem[Hosseini et~al.(2019)Hosseini, Ivanov, and Dolgui]{ref38_Hosseini2019}
Seyedmohsen Hosseini, Dmitry Ivanov, and Alexandre Dolgui.
\newblock {Review of quantitative methods for supply chain resilience
  analysis}.
\newblock \emph{Transportation Research Part E: Logistics and Transportation
  Review}, 125:\penalty0 285--307, 2019.
\newblock \doi{10.1016/j.tre.2019.03.001}.

\bibitem[Ivanov(2020)]{ref31_Ivanov2020}
Dmitry Ivanov.
\newblock {Predicting the impacts of epidemic outbreaks on global supply
  chains: A simulation-based analysis on the coronavirus outbreak
  (COVID-19/SARS-CoV-2) case}.
\newblock \emph{Transportation Research Part E: Logistics and Transportation
  Review}, 136:\penalty0 101922, 2020.
\newblock \doi{10.1016/j.tre.2020.101922}.

\bibitem[Ivanov and Dolgui(2021)]{ref34_Ivanov2021}
Dmitry Ivanov and Alexandre Dolgui.
\newblock {OR-methods for coping with the ripple effect in supply chains during
  COVID-19 pandemic: Managerial insights and research implications}.
\newblock \emph{International Journal of Production Economics}, 232:\penalty0
  107921, 2021.
\newblock \doi{10.1016/j.ijpe.2020.107921}.

\bibitem[Ke et~al.(2017)Ke, Meng, Finley, Wang, Chen, Ma, Ye, and
  Liu]{ref20_Ke2017}
Guolin Ke, Qi~Meng, Thomas Finley, Taifeng Wang, Wei Chen, Weidong Ma, Qiwei
  Ye, and Tie-Yan Liu.
\newblock {LightGBM: A Highly Efficient Gradient Boosting Decision Tree}.
\newblock In \emph{Advances in Neural Information Processing Systems 30 (NIPS
  2017)}, pages 3146--3154. Curran Associates, Inc., 2017.
\newblock URL
  \url{https://proceedings.neurips.cc/paper/2017/hash/6449f44a102fde848669bdd9eb6b76fa-Abstract.html}.

\bibitem[Koenker and Bassett(1978)]{ref22_Koenker1978}
Roger Koenker and Gilbert Bassett.
\newblock {Regression Quantiles}.
\newblock \emph{Econometrica}, 46\penalty0 (1):\penalty0 33--50, 1978.
\newblock \doi{10.2307/1913643}.

\bibitem[Li et~al.(2020)Li, Ran, Zhu, and Du]{ref26_Li2020}
Linchao Li, Bin Ran, Jiasong Zhu, and Bowen Du.
\newblock {Coupled application of deep learning model and quantile regression
  for travel time and its interval estimation using data in different
  dimensions}.
\newblock \emph{Applied Soft Computing}, 93:\penalty0 106387, 2020.
\newblock \doi{10.1016/j.asoc.2020.106387}.

\bibitem[Liu et~al.(2022)Liu, Lin, Chu, Zheng, and Chu]{ref11_Liu2022}
Ming Liu, Tao Lin, Feng Chu, Feifeng Zheng, and Chengbin Chu.
\newblock {A general robust dynamic Bayesian network method for supply chain
  disruption risk estimation under ripple effect}.
\newblock \emph{IFAC-PapersOnLine}, 55\penalty0 (10):\penalty0 1453--1458,
  2022.
\newblock \doi{10.1016/j.ifacol.2022.09.595}.

\bibitem[Maggioni et~al.(2017)Maggioni, Potra, and
  Bertocchi]{ref09_Maggioni2017}
Francesca Maggioni, Florian~A. Potra, and Marida Bertocchi.
\newblock {A scenario-based framework for supply planning under uncertainty:
  stochastic programming versus robust optimization approaches}.
\newblock \emph{Computational Management Science}, 14\penalty0 (1):\penalty0
  5--44, 2017.
\newblock \doi{10.1007/s10287-016-0272-3}.
\newblock Preprint: arXiv:1611.06514, doi:10.48550/arXiv.1611.06514.

\bibitem[Neal(1996)]{ref21_Neal1996}
Radford~M. Neal.
\newblock \emph{{Bayesian Learning for Neural Networks}}, volume 118 of
  \emph{Lecture Notes in Statistics}.
\newblock Springer, New York, NY, 1996.
\newblock \doi{10.1007/978-1-4612-0745-0}.

\bibitem[Norris(1997)]{ref14_Norris1997}
J.~R. Norris.
\newblock \emph{{Markov Chains}}.
\newblock Cambridge Series in Statistical and Probabilistic Mathematics.
  Cambridge University Press, Cambridge, 1997.
\newblock \doi{10.1017/CBO9780511810633}.

\bibitem[Parslov et~al.(2023)Parslov, Petersen, and
  Rodrigues]{ref27_Parslov2023}
Anders Parslov, Niklas~Christoffer Petersen, and Filipe Rodrigues.
\newblock {Short-term bus travel time prediction for transfer synchronization
  with intelligent uncertainty handling}.
\newblock \emph{Expert Systems with Applications}, 232:\penalty0 120751, 2023.
\newblock \doi{10.1016/j.eswa.2023.120751}.

\bibitem[Pettit et~al.(2019)Pettit, Croxton, and Fiksel]{ref37_Pettit2019}
Timothy~J. Pettit, Keely~L. Croxton, and Joseph Fiksel.
\newblock {The Evolution of Resilience in Supply Chain Management: A
  Retrospective on Ensuring Supply Chain Resilience}.
\newblock \emph{Journal of Business Logistics}, 40\penalty0 (1):\penalty0
  56--65, 2019.
\newblock \doi{10.1111/jbl.12202}.

\bibitem[Pomykacz et~al.(2024)Pomykacz, Gibas, and
  Baranowski]{ref28_Pomykacz2024}
Jan Pomykacz, Justyna Gibas, and Jerzy Baranowski.
\newblock {Bayesian Modeling of Travel Times on the Example of Food Delivery:
  Part 2---Model Creation and Handling Uncertainty}.
\newblock \emph{Electronics}, 13\penalty0 (17):\penalty0 3418, 2024.
\newblock \doi{10.3390/electronics13173418}.

\bibitem[Putthakosa and Hirotani(2023)]{ref12_Putthakosa2023}
Suchera Putthakosa and Daisuke Hirotani.
\newblock {Developing a Two-Stage Supply Chain Model using a Discrete Time
  Markov Chain Model during Supply Chain Disruptions}.
\newblock \emph{Industrial Engineering \& Management Systems}, 22\penalty0
  (4):\penalty0 449--462, 2023.
\newblock \doi{10.7232/iems.2023.22.4.449}.

\bibitem[Roscoe et~al.(2022)Roscoe, Aktas, Petersen, Skipworth, Handfield, and
  Habib]{ref32_Roscoe2022}
Samuel Roscoe, Emel Aktas, Kenneth~J. Petersen, Heather~D. Skipworth, Robert~B.
  Handfield, and Farooq Habib.
\newblock {Redesigning global supply chains during compounding geopolitical
  disruptions: the role of supply chain logics}.
\newblock \emph{International Journal of Operations \& Production Management},
  42\penalty0 (9):\penalty0 1407--1434, 2022.
\newblock \doi{10.1108/IJOPM-12-2021-0777}.

\bibitem[Sallam et~al.(2023)Sallam, Mohamed, and Mohamed]{ref08_Sallam2023}
Karam Sallam, Mona Mohamed, and Ali~Wagdy Mohamed.
\newblock {Internet of Things (IoT) in Supply Chain Management: Challenges,
  Opportunities, and Best Practices}.
\newblock \emph{Sustainable Machine Intelligence Journal}, 2\penalty0
  (1):\penalty0 3, 2023.
\newblock \doi{10.61185/SMIJ.2023.22103}.

\bibitem[Seber and Lee(2003)]{ref17_Seber2003}
George A.~F. Seber and Alan~J. Lee.
\newblock \emph{{Linear Regression Analysis}}.
\newblock Wiley Series in Probability and Statistics. John Wiley \& Sons,
  Hoboken, NJ, 2 edition, 2003.
\newblock ISBN 9780471415404.
\newblock \doi{10.1002/9780471722199}.
\newblock 1st edition: G. A. F. Seber (sole author), Wiley, 1977, no DOI.

\bibitem[Seco and Vieira(2014)]{ref07_Seco2014}
Alexandra Seco and Carlos Vieira.
\newblock {A Multi-agent Supply Chain Simulation Analysis through a Statistical
  Mixed Model}.
\newblock \emph{Procedia Technology}, 16:\penalty0 163--171, 2014.
\newblock \doi{10.1016/j.protcy.2014.10.079}.

\bibitem[Servos et~al.(2020)Servos, Liu, Teucke, and Freitag]{ref42_Servos2020}
Nikolaos Servos, Xiaodi Liu, Michael Teucke, and Michael Freitag.
\newblock {Travel Time Prediction in a Multimodal Freight Transport Relation
  Using Machine Learning Algorithms}.
\newblock \emph{Logistics}, 4\penalty0 (1):\penalty0 1, 2020.
\newblock \doi{10.3390/logistics4010001}.

\bibitem[Stamelou et~al.(2025)Stamelou, Ayfantopoulou, and
  Mallidis]{ref43_Stamelou2025}
A.~Stamelou, G.~Ayfantopoulou, and I.~Mallidis.
\newblock {Critical Analysis of the Main Factors Causing Delays in Various
  Stages of a Supply Chain and the Relevant Machine Learning Models for Dynamic
  Lead Time Predictions}.
\newblock In C.~McNally, P.~Carroll, B.~Martinez-Pastor, B.~Ghosh,
  M.~Efthymiou, and N.~Valantasis-Kanellos, editors, \emph{Transport
  Transitions: Advancing Sustainable and Inclusive Mobility. TRAconference
  2024}, Lecture Notes in Mobility. Springer, Cham, 2025.
\newblock ISBN 9783031952845.
\newblock \doi{10.1007/978-3-031-95284-5_70}.

\bibitem[Tarjan(1972)]{ref16_Tarjan1972}
Robert~E. Tarjan.
\newblock {Depth-First Search and Linear Graph Algorithms}.
\newblock \emph{SIAM Journal on Computing}, 1\penalty0 (2):\penalty0 146--160,
  1972.
\newblock \doi{10.1137/0201010}.

\bibitem[Verschuur et~al.(2021)Verschuur, Koks, and Hall]{ref30_Verschuur2021}
Jasper Verschuur, Elco~E. Koks, and Jim~W. Hall.
\newblock {Observed impacts of the COVID-19 pandemic on global trade}.
\newblock \emph{Nature Human Behaviour}, 5\penalty0 (3):\penalty0 305--307,
  2021.
\newblock \doi{10.1038/s41562-021-01060-5}.

\bibitem[Volpato(2025{\natexlab{a}})]{ref01_Volpato2025}
M.~Volpato.
\newblock {Delay Prediction in Supply Chains: A Graph-Based Modeling Approach}.
\newblock Master's thesis, Universit{\`a} degli Studi di Milano-Bicocca, Milan,
  Italy, 2025{\natexlab{a}}.

\bibitem[Volpato(2025{\natexlab{b}})]{ref24_Volpato2025a}
M.~Volpato.
\newblock {Dynamic route time notebook}.
\newblock GitHub repository iFoxz17/sc-delay-prediction, file
  code/rt\_estimator.ipynb, 2025{\natexlab{b}}.
\newblock URL
  \url{https://github.com/iFoxz17/sc-delay-prediction/blob/main/code/rt_estimator.ipynb}.

\bibitem[Volpato(2025{\natexlab{c}})]{ref25_Volpato2025b}
M.~Volpato.
\newblock {Prototype data model}.
\newblock GitHub repository iFoxz17/sc-delay-prediction, file
  plots/data\_model.png, 2025{\natexlab{c}}.
\newblock URL
  \url{https://github.com/iFoxz17/sc-delay-prediction/blob/main/plots/data_model.png}.

\bibitem[Wasi et~al.(2024)Wasi, Islam, Akib, and Bappy]{ref04_Wasi2024}
Azmine~Toushik Wasi, MD~Shafikul Islam, Adipto~Raihan Akib, and
  Mahathir~Mohammad Bappy.
\newblock {Graph Neural Networks in Supply Chain Analytics and Optimization:
  Concepts, Perspectives, Dataset and Benchmarks}.
\newblock arXiv preprint arXiv:2411.08550, 2024.
\newblock URL \url{https://arxiv.org/abs/2411.08550}.

\bibitem[Zhou et~al.(2023)Zhou, Wang, Liu, Wu, Shen, and Leung]{ref05_Zhou2023}
Xin Zhou, Jinglong Wang, Yong Liu, Xingyu Wu, Zhiqi Shen, and Cyril Leung.
\newblock {Inductive Graph Transformer for Delivery Time Estimation}.
\newblock In \emph{Proceedings of the Sixteenth ACM International Conference on
  Web Search and Data Mining (WSDM '23)}, pages 679--687, New York, NY, USA,
  2023. ACM.
\newblock \doi{10.1145/3539597.3570409}.

\end{thebibliography}

\clearpage

\pdfbookmark[1]{Supplemental Information}{Supplemental Information}
\makeatletter
\renewcommand{\toclevel@section}{2}
\renewcommand{\toclevel@subsection}{3}
\renewcommand{\toclevel@subsubsection}{4}
\makeatother
\hypersetup{bookmarksnumbered=true}
\section*{Supplemental Information}
This section provides supporting material referenced throughout the main text. Appendix~\ref{appendix-A} reproduces the formal theoretical results underpinning the Supply Chain Graph model, including the proofs of the maximal path, flow conservation, and routing probability theorems. Appendix~\ref{appendix-B} summarizes the relevant data tables from the order-level dataset provided in the industrial use case. Appendix~\ref{appendix-C} summarizes the historical and real-time indicators defined in the Logistic Chain Disruption Indicators section. Appendix~\ref{appendix-D} contains the derivations of the E-PT and E-TFST, showing how the linearity of the expectation operator allows the contributions of Transit Time and Path Time to be aggregated over the distribution of possible paths. Appendix~\ref{appendix-E} reports the evaluation and cross-validation results supporting the selection of the regression model for the Dynamic Route Time indicator. Appendix~\ref{appendix-F} provides the definitions of the evaluation metrics used to assess predictive quality in the Results section. Appendix~\ref{appendix-G} reports the evaluation results for the TFST weighting factors, Appendix~\ref{appendix-H} lists the hyperparameter settings for all the models introduced in the Regression Models Training section, and Appendix~\ref{appendix-I} summarizes the notation used throughout the manuscript.

\clearpage

\appendix

\section{Formal Basis of the Supply Chain Graph Model}\label{appendix-A}
\setcounter{table}{0}\renewcommand{\thetable}{A\arabic{table}}\renewcommand{\theHtable}{A.\arabic{table}}
\setcounter{figure}{0}\renewcommand{\thefigure}{A\arabic{figure}}\renewcommand{\theHfigure}{A.\arabic{figure}}
\setcounter{equation}{0}\renewcommand{\theequation}{A.\arabic{equation}}\renewcommand{\theHequation}{A.\arabic{equation}}

This appendix reproduces the formal basis for the approach of the Supply Chain Graph Model (SCGM). Theorem A.1 justifies the procedure for path extraction, used to enumerate all shipment routes from a given vertex, Theorem A.2 establishes the flow conservation property underlying the graph's structural consistency, lastly Theorem A.3 ensures that the routing probabilities computed via the Markov chain model form a valid probability distribution.

\textbf{Theorem A.1 (Maximal Paths).} \textit{Let \(G = (V,A)\) be a Supply Chain Graph and \(v \in V\) be a vertex. A path \(\pi = (v_{0},v_{1},\ldots,v_{k})\) is maximal if it cannot be extended by a further vertex without forming a cycle or violating the edge set of the graph. This condition occurs if and only if \(v_{k} = m\), the manufacturer vertex.}

\emph{Proof.} (\(\Leftarrow\)) If \(v_{k} = m\), then since \(m\) is, by construction, a sink vertex with no outgoing arcs (\(d_{\mathrm{out}}(m) = 0\)), the path cannot be extended and is therefore maximal. (\(\Rightarrow\)) Let \(\pi\) be maximal and suppose, for contradiction, that \(v_{k} \neq m\). Since \(m\) is the unique sink of the graph, it holds \(d_{\mathrm{out}}(v_{k}) > 0\), so there exists at least one outgoing arc \(\left( v_{k},v_{k + 1} \right) \in A\). Because the graph is acyclic, \(v_{k + 1} \notin \{ v_{0},\ldots,v_{k}\}\), and \(\pi\) can be extended to \(\kappa = (v_{0},\ldots,v_{k},v_{k + 1})\), contradicting maximality. Hence it must follow that \(v_{k} = m\).\hfill $\blacksquare$

\textbf{Theorem A.2 (Flow Conservation).} \textit{Let $G = (V,A)$ be a Supply Chain Graph with $V = S \cup I \cup M$ and $A = A_{SI} \cup A_{II} \cup A_{IM}$ the set of arcs. Let $f : V^{2} \rightarrow \mathbb{N}$ be a flow function satisfying the flow balance constraint at each intermediate vertex. Then the total flow is conserved across the graph, that is, $f_{V}(S) = f_{V}(M)$.}

\emph{Proof.} The total outgoing flow from supplier sites is
\begin{equation}
f_{V}(S) = \sum_{(s,i) \in A_{SI}}^{}{f(s,i)}
\label{eq:A1}
\end{equation}
and the total inflow at the manufacturer vertex is
\begin{equation}
f_{V}(M) = \sum_{(i,m) \in A_{IM}}^{}{f(i,m)}
\label{eq:A2}
\end{equation}
since no arcs connect the supplier sites directly to the manufacturer. The flow balance constraint at each intermediate vertex \(i \in I\) ensures
\begin{equation}
\sum_{(v,i) \in A}^{}{f(v,i)} = \sum_{(i,v) \in A}^{}{f(i,v)}
\label{eq:A3}\,.
\end{equation}
Summing over all intermediate vertices, every arc in \(A_{II}\) appears exactly once on each side of the equality and cancels out, yielding
\begin{equation}
\sum_{(s,i) \in A_{SI}}^{}{f(s,i)} = \sum_{(i,m) \in A_{IM}}^{}{f(i,m)}
\label{eq:A4}
\end{equation}
that is, \(f_{V}(S) = f_{V}(M)\).\hfill$\blacksquare$

\textbf{Theorem A.3 (Path Probabilities).} \textit{Let $G = (V,A)$ be a Supply Chain Graph with $V = S \cup I \cup \{m\}$ and let $v \in V$ be a vertex. Let $\Pi(v) = \{\pi_{1},\pi_{2},\ldots,\pi_{\ell}\}$ denote the set of all maximal paths from $v$ to the manufacturer vertex $m$, with probability}
\begin{equation}
p_{i} = \mathbb{P}\left( \pi_{i} \,\middle|\, X_{0} = v \right) = \prod_{(u,w) \in \pi_{i}} \mathbb{P}\left( X = u \,\middle|\, X = w \right)\,.
\label{eq:A5}
\end{equation}
\textit{Then $p = \{p_{1},p_{2},\ldots,p_{\ell}\}$ is a valid probability distribution, following $p_{i} \in [0,1]$ and $p_{1} + p_{2} + \ldots + p_{\ell} = 1$.}

\textit{Proof.} The non-negativity of $p_{i}$ is immediate, since the transition probabilities $\mathbb{P}\left( X = u \,\middle|\, X = w \right) = f(w,u)/f_{V}(w)$ all lie in $[0,1]$. Normalization is proved by induction on the starting vertex. For the case $v = m$, the only maximal path is $\pi = (m)$, with $\mathbb{P}\left( \pi \,\middle|\, X_{0} = m \right) = P_{mm} = 1$. Inductive step: assume
\begin{equation}
\sum_{\pi \in \Pi(n)} \mathbb{P}\left( \pi \,\middle|\, X_{0} = n \right) = 1 \quad\text{for any neighbour}\quad n \in N(v) = \left\{ n \in V : (v,n) \in A \right\}.
\end{equation}
Every maximal path from $v$ decomposes uniquely as $\pi = (v,\kappa)$ with $\kappa \in \Pi(n)$ for some neighbour $n \in N(v)$, and by the Markov property $\mathbb{P}\left( \pi \,\middle|\, X_{0} = v \right) = P_{vn} \cdot \mathbb{P}\left( \kappa \,\middle|\, X_{0} = n \right)$. Therefore,
\begin{equation}
\sum_{i = 1}^{\ell}p_{i} = \sum_{\pi \in \Pi(v)}^{}{\mathbb{P}\left( \pi\  \middle| \ X_{0} = v \right)} = \sum_{n \in N(v)}^{}{\sum_{\kappa \in \Pi(n)}^{}{P_{vn}\mathbb{\cdot P}\left( \kappa\  \middle| \ X_{0} = n \right)}} = \sum_{n \in N(v)}^{}P_{vn} = 1
\label{eq:A6}
\end{equation}
because the transition probabilities from \(v\) form a distribution over its neighbours.\hfill$\blacksquare$

\clearpage
\section{Order-level Dataset Tables}\label{appendix-B}
\setcounter{table}{0}\renewcommand{\thetable}{B\arabic{table}}\renewcommand{\theHtable}{B.\arabic{table}}
\setcounter{figure}{0}\renewcommand{\thefigure}{B\arabic{figure}}\renewcommand{\theHfigure}{B.\arabic{figure}}
\setcounter{equation}{0}\renewcommand{\theequation}{B.\arabic{equation}}\renewcommand{\theHequation}{B.\arabic{equation}}

This appendix summarizes the relevant data tables from the order-level dataset provided by FAE Technology S.p.A.

\begin{table}[htbp]
\centering
\caption[Order-level dataset schema]{Order-level dataset schema. The schema reports for each order, the identifier, associated supplier site, order issuance and confirmed delivery timestamps, and carrier tracking link.}\label{tab:B1}
\small
\begin{tabular}{@{}lll@{}}
\toprule
Column & Type & Description \\
\midrule
\texttt{id} & int & Order identifier number \\
\texttt{distributorID} & int & Supplier identifier number \\
\texttt{createdAt} & datetime & Order issuance time \\
\texttt{confirmedDeliveryDate} & datetime & Actual delivery time \\
\texttt{trackingLink} & string & Shipment tracking URL from carrier \\
\bottomrule
\end{tabular}
\end{table}

\begin{table}[htbp]
\centering
\caption[Order procurement overview]{Order procurement overview. The data table reports for each supplier site ID, the associated supplier ID, geographic origin, country, and number of orders. Site ID 3 accounted for the majority of the orders, reflecting the data imbalance discussed in the Use Case Implementation section.}\label{tab:B2}
\small
\begin{tabular}{@{}rrllr@{}}
\toprule
Site ID & Supplier ID & Location of Origin & Country & Orders \\
\midrule
1 & 1 & Leeds, Leeds & GB & 5 \\
2 & 2 & Louisville, Kentucky & US & 3 \\
3 & 2 & Thief River Falls, Minnesota & US & 61 \\
4 & 3 & Grand Prairie, Texas & US & 5 \\
5 & 4 & Montagnola, Ticino & CH & 1 \\
6 & 4 & Peabody, Massachusetts & US & 3 \\
7 & 4 & Shenzhen, Guangdong & CN & 1 \\
8 & 5 & Dongguan, Guangdong & CN & 2 \\
9 & 5 & Shenzhen, Guangdong & CN & 1 \\
10 & 7 & Hong Kong, Hong Kong & HK & 1 \\
11 & 7 & Zhuhai, Guangdong & CN & 1 \\
12 & 8 & Warsaw, Mazovia & PL & 1 \\
13 & 9 & Ho Chi Minh City, Ho Chi Minh & VN & 1 \\
14 & 9 & San José, San José & CR & 1 \\
15 & 10 & Bratislava, Bratislava Region & SK & 1 \\
16 & 10 & Gdansk, Pomerania & PL & 3 \\
17 & 11 & Shenzhen, Guangdong & CN & 3 \\
18 & 12 & Boca Raton, Florida & US & 1 \\
19 & 13 & Shenzhen, Guangdong & CN & 1 \\
20 & 14 & Minneapolis, Minnesota & US & 1 \\
21 & 16 & Tsuen Wan, Tsuen Wan & HK & 1 \\
\bottomrule
\end{tabular}
\end{table}

\begin{table}[htbp]
\centering
\caption[Distribution of orders across carriers in the pilot dataset]{Distribution of orders across carriers in the pilot dataset. A set of logistics providers, namely UPS, FedEx, DHL, and DHL Express, accounted for the entirety of the observed shipments.}\label{tab:B3}
\small
\begin{tabular}{@{}lr@{}}
\toprule
Carrier & Orders \\
\midrule
DHL Express & 4 \\
DHL & 12 \\
FedEx & 13 \\
UPS & 69 \\
\bottomrule
\end{tabular}
\end{table}

\clearpage
\section{Summary of the Logistic Chain Disruption Indicators}\label{appendix-C}
\setcounter{table}{0}\renewcommand{\thetable}{C\arabic{table}}\renewcommand{\theHtable}{C.\arabic{table}}
\setcounter{figure}{0}\renewcommand{\thefigure}{C\arabic{figure}}\renewcommand{\theHfigure}{C.\arabic{figure}}
\setcounter{equation}{0}\renewcommand{\theequation}{C.\arabic{equation}}\renewcommand{\theHequation}{C.\arabic{equation}}

This appendix summarizes the historical indicators and real-time indicators as defined within the Logistic Chain Disruption Indicators section, for each one, the subscript \(\mu\) denotes the pointwise (scalar) version of an indicator and \(\beta\) denotes its interval-based version, building upon a confidence interval.

\begin{table}[htbp]
\centering
\caption[Summary of the LCDIs]{Summary of the LCDIs, grouped into historical indicators and real-time indicators at the vertex, route, path, and graph levels.}\label{tab:C1}
\small
\begin{tabular}{@{}>{\raggedright\arraybackslash}p{0.17\linewidth}>{\raggedright\arraybackslash}p{0.13\linewidth}>{\raggedright\arraybackslash}p{0.64\linewidth}@{}}
\toprule
Name & Type & Description \\
\midrule
\multicolumn{3}{@{}l}{\textbf{Historical LCDIs}} \\[2pt]
$\text{DT}_{\mu}$ & Scalar & Expected dispatch time, calculated per supplier site \\
$\text{DT}_{\beta}$ & Interval & Estimate of dispatch time, calculated per supplier site \\
$\text{ST}_{\mu}$ & Scalar & Expected shipment time, calculated per supplier site and carrier \\
$\text{ST}_{\beta}$ & Interval & Estimate of shipment time, calculated per supplier site and carrier \\
$\text{DLT}_{\mu}$ & Scalar & Expected total delivery time, calculated per supplier site and carrier \\
$\text{DLT}_{\beta}$ & Interval & Estimate of total delivery time, calculated per supplier site and carrier \\[4pt]
\multicolumn{3}{@{}l}{\textbf{Real-time vertex-level LCDIs}} \\[2pt]
EDT & Scalar & Estimated dispatch time, adjusted for non-operational days \\
ORI & Distribution & Residence time distribution, aggregating data across carriers \\
$\text{VT}_{\mu}$ & Scalar & Expected residence time \\
$\text{VT}_{\beta}$ & Interval & Estimate of the residence time \\[4pt]
\multicolumn{3}{@{}l}{\textbf{Real-time route-level LCDIs}} \\[2pt]
OTI & Distribution & Route time distribution, aggregating data across carriers \\
TMI & Scalar & Measure of traffic congestion \\
WMI & Scalar & Measure of adverse weather impact \\
$\text{RT}_{\mu}$ & Scalar & Expected route transit time \\
$\text{RT}_{\beta}$ & Interval & Estimate of the route transit time \\
$\text{DRT}_{\mu}$ & Scalar & Expected route transit time, with traffic and weather conditions \\
$\text{DRT}_{\beta}$ & Interval & Estimate of the route transit time, with traffic and weather conditions \\[4pt]
\multicolumn{3}{@{}l}{\textbf{Real-time path-level LCDIs}} \\[2pt]
$\text{TT}_{\mu}$ & Scalar & Expected remaining shipment time based on historical shipment data \\
$\text{TT}_{\beta}$ & Interval & Estimate of remaining shipment time based on historical shipment data \\
$\text{PT}_{\mu}$ & Scalar & Estimate of the remaining shipment time from the SCGM \\
$\text{PT}_{\beta}$ & Interval & Estimate of the remaining shipment time from the SCGM \\
$\text{TFST}_{\mu}$ & Scalar & Expected remaining shipment time combining $\text{TT}_{\mu}$ and $\text{PT}_{\mu}$ \\
$\text{TFST}_{\beta}$ & Interval & Estimate of the remaining shipment time combining $\text{TT}_{\beta}$ and $\text{PT}_{\beta}$ \\[4pt]
\multicolumn{3}{@{}l}{\textbf{Real-time graph-level LCDIs}} \\[2pt]
$\text{E-PT}_{\mu}$ & Scalar & Expected value of $\text{PT}_{\mu}$ over all the probable paths \\
$\text{E-PT}_{\beta}$ & Interval & Expected value of $\text{PT}_{\beta}$ over all the probable paths \\
$\text{E-TFST}_{\mu}$ & Scalar & Expected value of $\text{TFST}_{\mu}$ over all the probable paths \\
$\text{E-TFST}_{\beta}$ & Interval & Expected value of $\text{TFST}_{\beta}$ over all the probable paths \\
$\text{EODT}_{\mu}$ & Scalar & Expected total time from order placement to delivery \\
$\text{EODT}_{\beta}$ & Interval & Estimate of the total time from order placement to delivery \\
\bottomrule
\end{tabular}
\end{table}

\clearpage
\section{Derivation of the E-PT and E-TFST Indicators}\label{appendix-D}
\setcounter{table}{0}\renewcommand{\thetable}{D\arabic{table}}\renewcommand{\theHtable}{D.\arabic{table}}
\setcounter{figure}{0}\renewcommand{\thefigure}{D\arabic{figure}}\renewcommand{\theHfigure}{D.\arabic{figure}}
\setcounter{equation}{0}\renewcommand{\theequation}{D.\arabic{equation}}\renewcommand{\theHequation}{D.\arabic{equation}}

Let \(\Pi(v) = \left\{ \pi_{1},\pi_{2},\ldots,\pi_{\ell} \right\}\) be the set of maximal paths from the current vertex \(v\) to the manufacturer vertex \(m\), and let the set \(p^{c} = \left\{ p_{1}^{c},p_{2}^{c},\ \ldots,p_{\ell}^{c} \right\}\) be the corresponding carrier-conditioned path probabilities over \(c \in C\) as defined in the Carrier-conditioned Routing Model section which form a valid probability distribution, by Theorem A.3. Let \(\mathcal{P} \in \Pi(v)\) be a random path with associated probability \(p_{i}^{c}= \mathbb{P}(\mathcal{P} = \pi_{i}\  \land \ C = c)\). The Expected Path Time is defined as the expectation of the PT indicator over the path distribution,
\begin{equation}
\text{E-PT}\left( s,c,v \right) = \mathbb{E}_{\mathcal{P} \sim p^{c}}\left\lbrack \text{PT}(s,c,\mathcal{P}) \right\rbrack = \sum_{1\le i\le\ell} p_i^c \cdot \text{PT}\left( s,c,\pi_i \right)
\label{eq:D1}\,.
\end{equation}
As the sum is computed component-wise, two forms exist for this indicator. The first one is derived from the scalar form of the PT indicator, and the second one is derived from interval-based form of the PT indicator. The Expected Total Forecasted Shipment Time is defined as the expectation of the TFTS over the path distribution,
\begin{equation}
\begin{aligned}
\text{E-TFST}(s,c,t;\alpha) & = \mathbb{E}_{\mathcal{P} \sim p^{c}}\left\lbrack \text{TFST}(s,c,\mathcal{P}) \right\rbrack \\[6pt]
& = \mathbb{E}_{\mathcal{P} \sim p^{c}}\left\lbrack \alpha \cdot \text{TT}(s,c,t) + (1 - \alpha) \cdot \text{PT}(s,c,\mathcal{P}) \right\rbrack \\[6pt]
& = \alpha \cdot \mathbb{E}_{\mathcal{P} \sim p^{c}}\left\lbrack \text{TT}(s,c,t) \right\rbrack + (1 - \alpha) \cdot \mathbb{E}_{\mathcal{P} \sim p^{c}}\left\lbrack \text{PT}(s,c,\mathcal{P}) \right\rbrack \\[6pt]
& = \alpha \cdot \text{TT}(s,c,t) + (1 - \alpha) \cdot \text{E-PT}\left( s,c,v(t) \right)
\end{aligned}
\label{eq:D2}
\end{equation}
since the TT depends only on the supplier site, carrier, and elapsed shipment time (referred to as the current vertex \(v(t)\) on the logistics chain), and the weighting factor \(\alpha\) depends at most on the elapsed shipment time, neither quantity varies with the path distribution. This allows the linearity of the expectation operator to be exploited, and as this formulation aggregates contributions of PT and TT over the distribution of possible paths, producing either a scalar form or an interval-based form of the remaining shipment time.

\clearpage
\section{DRT Model Evaluation and Cross-validation Results}\label{appendix-E}
\setcounter{table}{0}\renewcommand{\thetable}{E\arabic{table}}\renewcommand{\theHtable}{E.\arabic{table}}
\setcounter{figure}{0}\renewcommand{\thefigure}{E\arabic{figure}}\renewcommand{\theHfigure}{E.\arabic{figure}}
\setcounter{equation}{0}\renewcommand{\theequation}{E.\arabic{equation}}\renewcommand{\theHequation}{E.\arabic{equation}}

This appendix reports the results mentioned in the Regression Model section, comprising the DRT model performance evaluation results split in Table \ref{tab:E1} and Table \ref{tab:E2}, for readability, together with summary statistics from 10-fold cross-validation of XGBoost with mean, standard deviation and 95\% confidence interval.

\begin{table}[htbp]
\centering
\caption[DRT Model performance evaluation for all datasets with featurized subsets A and B]{DRT Model performance evaluation for all datasets with featurized subsets A and B. The results include MAE (days), RMSE (days), and MAPE metrics, together with the historical baseline. Bold* values indicate the best performing combination in terms of each minimum metric across all performance tests, including featurized subsets C and D.}\label{tab:E1}
\small
\begin{tabular}{@{}ll*{6}{>{\centering\arraybackslash}p{\widthof{MAPE (\%)}}}@{}}
\toprule
 &  & \multicolumn{3}{c}{Subset A} & \multicolumn{3}{c}{Subset B} \\
\cmidrule(lr){3-5}\cmidrule(lr){6-8}
Model & Dataset & MAE & RMSE & MAPE (\%) & MAE & RMSE & MAPE (\%) \\
\midrule
Baseline & All & 1.1776 & 2.0302 & 0.9643 & 1.1776 & 2.0302 & 0.9643 \\
\midrule
LinReg & \multirow{4}{*}{\texttt{ds\_0}} & 1.4014 & 2.0558 & 1.3895 & 1.1884 & 1.9532 & 0.6812 \\
RndForest & & 1.0748 & 2.0248 & 0.7869 & 1.0837 & 2.1047 & 0.8400 \\
XGBoost & & 1.0451 & 2.1634 & 0.4940 & 1.1010 & 2.2753 & 0.5644 \\
LightGBM & & 1.3724 & 2.2733 & 1.4150 & 1.4249 & 2.3456 & 2.0049 \\
\midrule
LinReg & \multirow{4}{*}{\texttt{ds\_1}} & 1.3416 & 1.9569 & 1.4827 & 1.1700 & 1.8965 & 0.8295 \\
RndForest & & 1.0455 & 1.9238 & 0.8235 & 0.9846 & 1.7008 & 0.9950 \\
XGBoost & & 0.9230 & 1.7246 & 0.4128 & \textbf{0.7902\rlap{*}} & 1.6647 & 0.4601 \\
LightGBM & & 1.0420 & 1.8618 & 0.7314 & 1.0362 & 1.7081 & 1.1943 \\
\midrule
LinReg & \multirow{4}{*}{\texttt{ds\_2}} & 1.2794 & 1.9078 & 1.5687 & 1.1421 & 1.8697 & 1.0047 \\
RndForest & & 1.1027 & 2.1194 & 0.4577 & 0.9721 & 1.7480 & 0.7622 \\
XGBoost & & 0.9950 & 1.7657 & 0.6780 & 0.8464 & \textbf{1.5737\rlap{*}} & \textbf{0.3793\rlap{*}} \\
LightGBM & & 1.0447 & 1.9264 & 0.7371 & 1.0470 & 1.8543 & 1.0007 \\
\bottomrule
\end{tabular}
\end{table}

\begin{table}[htbp]
\centering
\caption[DRT Model performance evaluation for all datasets with featurized subsets C and D]{DRT Model performance evaluation for all datasets with featurized subsets C and D. This table is a continuation of results from Table~\ref{tab:E1} including MAE (days), RMSE (days), and MAPE metrics, together with the baseline.}\label{tab:E2}
\small
\begin{tabular}{@{}ll*{6}{>{\centering\arraybackslash}p{\widthof{MAPE (\%)}}}@{}}
\toprule
 &  & \multicolumn{3}{c}{Subset C} & \multicolumn{3}{c}{Subset D} \\
\cmidrule(lr){3-5}\cmidrule(lr){6-8}
Model & Dataset & MAE & RMSE & MAPE (\%) & MAE & RMSE & MAPE (\%) \\
\midrule
Baseline & All & 1.1776 & 2.0302 & 0.9643 & 1.1776 & 2.0302 & 0.9643 \\
\midrule
LinReg & \multirow{4}{*}{\texttt{ds\_0}} & 1.4665 & 2.0478 & 1.5430 & 1.2623 & 1.9361 & 0.8049 \\
RndForest & & 1.0838 & 2.1048 & 0.8400 & 1.0942 & 1.8193 & 1.5767 \\
XGBoost & & 1.1010 & 2.2753 & 0.5644 & 1.1608 & 2.3298 & 0.7349 \\
LightGBM & & 1.4249 & 2.3456 & 2.0049 & 1.2857 & 2.1548 & 1.5421 \\
\midrule
LinReg & \multirow{4}{*}{\texttt{ds\_1}} & 1.3720 & 1.9670 & 1.5537 & 1.2021 & 1.8980 & 0.9062 \\
RndForest & & 0.9846 & 1.7008 & 0.9950 & 1.0172 & 1.6274 & 1.5525 \\
XGBoost & & 0.7905 & 1.6449 & 0.4685 & 1.0959 & 2.0909 & 1.0819 \\
LightGBM & & 1.0343 & 1.7011 & 1.0832 & 1.0405 & 1.6091 & 0.8216 \\
\midrule
LinReg & \multirow{4}{*}{\texttt{ds\_2}} & 1.3009 & 1.9193 & 1.6189 & 1.1642 & 1.8735 & 1.0661 \\
RndForest & & 0.9721 & 1.7480 & 0.7622 & 1.0460 & 1.5908 & 1.6122 \\
XGBoost & & 0.8728 & 1.8532 & 0.5783 & 1.1352 & 2.1715 & 1.0516 \\
LightGBM & & 1.0470 & 1.8543 & 1.0007 & 1.1943 & 1.9036 & 1.0228 \\
\bottomrule
\end{tabular}
\end{table}

\begin{table}[htbp]
\centering
\caption[Cross-validation performance of XGBoost across the three best-performing feature subsets (A, B, C)]{Cross-validation performance of XGBoost across the three best-performing feature subsets (A, B, C). Each feature subset is paired with the dataset augmentation iteration yielding its best result in Table~\ref{tab:E1} and Table~\ref{tab:E2}: Subset A with \texttt{ds\_1} (minimum MAE), Subset B with \texttt{ds\_2} (minimum RMSE and MAPE), and Subset C with \texttt{ds\_1} (next-best minimum MAE). For each combination, results are reported both with and without hyperparameter optimization (HyOpt). The first row reports the historical baseline. Values are shown as mean $\pm$ standard deviation, with 95\% confidence intervals in brackets, computed from the Student's t-distribution over 10-fold cross-validation. Bold values indicate the configuration (Subset A, \texttt{ds\_1}, absent hyperparameter optimization) used for the use case implementation.}\label{tab:E3}
\small
\begin{tabular}{@{}llllll@{}}
\toprule
HyOpt & Subset & Dataset & MAE (days) & RMSE (days) & MAPE (\%) \\
\midrule
- & - & - & 1.60 $\pm$ 0.28 {[}1.40, 1.80{]} & 3.58 $\pm$ 1.50 {[}2.50, 4.65{]} & 0.77 $\pm$ 0.26 {[}0.58, 0.95{]} \\
\midrule
\multirow{3}{*}{No} & \textbf{A} & \texttt{ds\_1} & \textbf{1.12 $\pm$ 0.58 {[}0.70, 1.53{]}} & \textbf{3.01 $\pm$ 2.84 {[}0.98, 5.05{]}} & \textbf{0.44 $\pm$ 0.17 {[}0.32, 0.57{]}} \\
 & B & \texttt{ds\_2} & 1.14 $\pm$ 0.58 {[}0.72, 1.55{]} & 3.02 $\pm$ 2.63 {[}1.14, 4.89{]} & 0.54 $\pm$ 0.28 {[}0.34, 0.75{]} \\
 & C & \texttt{ds\_1} & 1.09 $\pm$ 0.57 {[}0.69, 1.50{]} & 2.84 $\pm$ 2.74 {[}0.88, 4.80{]} & 0.47 $\pm$ 0.22 {[}0.31, 0.63{]} \\
\midrule
\multirow{3}{*}{Yes} & A & \texttt{ds\_1} & 1.37 $\pm$ 0.57 {[}0.96, 1.78{]} & 3.15 $\pm$ 2.79 {[}1.16, 5.15{]} & 2.01 $\pm$ 0.97 {[}1.32, 2.70{]} \\
 & B & \texttt{ds\_2} & 1.18 $\pm$ 0.57 {[}0.77, 1.59{]} & 2.93 $\pm$ 2.31 {[}1.28, 4.58{]} & 0.58 $\pm$ 0.33 {[}0.35, 0.82{]} \\
 & C & \texttt{ds\_1} & 1.10 $\pm$ 0.56 {[}0.70, 1.50{]} & 2.79 $\pm$ 2.81 {[}0.78, 4.80{]} & 0.54 $\pm$ 0.21 {[}0.39, 0.69{]} \\
\bottomrule
\end{tabular}
\end{table}

\clearpage
\section{Evaluation Metrics}\label{appendix-F}
\setcounter{table}{0}\renewcommand{\thetable}{F\arabic{table}}\renewcommand{\theHtable}{F.\arabic{table}}
\setcounter{figure}{0}\renewcommand{\thefigure}{F\arabic{figure}}\renewcommand{\theHfigure}{F.\arabic{figure}}
\setcounter{equation}{0}\renewcommand{\theequation}{F.\arabic{equation}}\renewcommand{\theHequation}{F.\arabic{equation}}

This appendix lists the evaluation metrics used in the Results section. For a set of \(N\) predictions with both lower and upper bound intervals \(\left\lbrack l_{i},u_{i} \right\rbrack\) and observed values \(y_{i}\), it is possible to define the following metrics.

\textbf{Sharpness}: Measures the average width of the predicted intervals \(\left( u_{i} - l_{i} \right)\), independently of the observations of the model. Defined as:
\begin{equation}
\text{Sharpness} = \frac{1}{N}\sum_{i = 1}^{N}\left( u_{i} - l_{i} \right)
\label{eq:F1}
\end{equation}
where narrower intervals indicate more informative estimations.

\textbf{Coverage}: Quantifies the average observations falling within the predicted intervals. Defined as:
\begin{equation}
\text{Coverage} = \frac{1}{N}\sum_{i = 1}^{N}{\mathbf{1}\left\{ l_{i} \leq y_{i} \leq u_{i} \right\}}
\label{eq:F2}
\end{equation}
where a model with coverage close to the nominal confidence level is called well calibrated, else, if the coverage lies below or above the nominal levels, it indicates over-confidence or under-confidence in uncertainty estimates.

\textbf{Interval Score}: Combines sharpness and coverage in a single measure~\citep{ref23_Gneiting2007}. After a significance level \(\alpha\) (complement of the confidence level) is declared, the pointwise interval score \({IS}_{\alpha}\left( l_{i},u_{i};y_{i} \right)\) is defined for a single prediction as:
\begin{equation}
\text{IS}_{\alpha}\left( l_{i},u_{i};y_{i} \right) = \left( u_{i} - l_{i} \right) + \frac{2}{\alpha}\left( l_{i} - y_{i} \right)\,\mathbf{1}\left\{ y_{i} < l_{i} \right\} + \frac{2}{\alpha}\left( y_{i} - u_{i} \right)\,\mathbf{1}\left\{ y_{i} > u_{i} \right\}
\label{eq:F3}
\end{equation}
where \(\left( u_{i} - l_{i} \right)\) is the width of the predicted intervals, with two penalty factors scaled by \(2/\alpha\). It follows that narrow intervals containing the observation are rewarded, while other intervals are penalized proportionally to the distance between the bound and the observations. The interval Score is defined as:
\begin{equation}
\text{Interval Score}(\alpha) = \frac{1}{N}\sum_{i = 1}^{N}{\text{IS}_{\alpha}\left( l_{i},u_{i};y_{i} \right)}
\label{eq:F4}
\end{equation}
where lower integral score values indicate more informative and better calibrated prediction intervals.

\clearpage
\section{TFST Weighting Factor Evaluation Results}\label{appendix-G}
\setcounter{table}{0}\renewcommand{\thetable}{G\arabic{table}}\renewcommand{\theHtable}{G.\arabic{table}}
\setcounter{figure}{0}\renewcommand{\thefigure}{G\arabic{figure}}\renewcommand{\theHfigure}{G.\arabic{figure}}
\setcounter{equation}{0}\renewcommand{\theequation}{G.\arabic{equation}}\renewcommand{\theHequation}{G.\arabic{equation}}

This appendix reports the full evaluation results for the different TFST weighting factor implementations across training and test sets, for the dispatch and shipment stages in tables Table~\ref{tab:G1} and Table~\ref{tab:G2} respectively, complementing the summary discussion presented in the TFST Weighting Factor Performance section.

\begin{table}[htbp]
\centering
\caption[Performance comparison of model variants v0.0-v2.0 for the dispatching stage]{Performance comparison of model variants v0.0-v2.0 for the dispatch stage. Bold values indicate the variant with optimal metric values obtained for the test set, evaluated at both granularities.}\label{tab:G1}
\small
\begin{tabular}{@{}llccc@{}}
\toprule
\multirow{2}{*}{Set} & \multirow{2}{*}{Variant} & Sharpness & Coverage & Interval Score \\
& & Hourly / Daily & Hourly / Daily & Hourly / Daily \\
\midrule
\multirow{5}{*}{Train} & v0.0 & 2.15 / 2.20 & 0.28 / 0.56 & 4.22 / 4.26 \\
& v0.1 & 1.67 / 1.70 & 0.23 / 0.33 & 4.38 / 4.44 \\
& v0.2 & 1.91 / 2.36 & 0.31 / 0.55 & 4.24 / 4.43 \\
& v1.0 & 2.15 / 2.20 & 0.28 / 0.56 & 4.22 / 4.26 \\
& v2.0 & 2.15 / 2.20 & 0.28 / 0.56 & 4.22 / 4.26 \\
\midrule
\multirow{5}{*}{Test} & v0.0 & 2.87 / 3.00 & 0.30 / 0.50 & 7.81 / 7.44 \\
& v0.1 & 1.62 / 1.10 & 0.10 / 0.40 & 7.41 / 6.88 \\
& v0.2 & 2.25 / 2.80 & 0.30 / 0.50 & 7.59 / 7.24 \\
& v1.0 & 2.87 / 3.00 & 0.30 / 0.50 & 7.81 / 7.44 \\
& \textbf{v2.0} & \textbf{2.87 / 3.00} & \textbf{0.30 / 0.50} & \textbf{7.81 / 7.44} \\
\bottomrule
\end{tabular}
\end{table}

\begin{table}[htbp]
\centering
\caption[Performance comparison of model variants v0.0-v2.0 for the shipment stage]{Performance comparison of model variants v0.0-v2.0 for the shipment stage. Bold values indicate the variant with optimal metric values obtained for the test set, evaluated at both granularities.}\label{tab:G2}
\small
\begin{tabular}{@{}llccc@{}}
\toprule
\multirow{2}{*}{Set} & \multirow{2}{*}{Variant} & Sharpness & Coverage & Interval Score \\
& & Hourly / Daily & Hourly / Daily & Hourly / Daily \\
\midrule
\multirow{5}{*}{Train} & v0.0 & 1.84 / 1.87 & 0.66 / 0.96 & 1.97 / 2.03 \\
& v0.1 & 0.80 / 0.91 & 0.42 / 0.71 & 1.56 / 1.73 \\
& v0.2 & 1.32 / 1.38 & 0.55 / 0.84 & 1.65 / 1.82 \\
& v1.0 & 1.38 / 1.50 & 0.63 / 0.88 & 1.67 / 1.83 \\
& v2.0 & 1.03 / 1.18 & 0.52 / 0.84 & 1.45 / 1.63 \\
\midrule
\multirow{5}{*}{Test} & v0.0 & 2.33 / 2.41 & 0.80 / 0.91 & 2.72 / 2.93 \\
& v0.1 & 0.77 / 0.96 & 0.52 / 0.81 & 1.74 / 1.80 \\
& v0.2 & 1.55 / 1.57 & 0.68 / 0.88 & 2.05 / 2.17 \\
& v1.0 & 1.66 / 1.89 & 0.74 / 0.88 & 2.21 / 2.47 \\
& \textbf{v2.0} & \textbf{1.30 / 1.55} & \textbf{0.63 / 0.86} & \textbf{1.99 / 2.20} \\
\bottomrule
\end{tabular}
\end{table}

\clearpage
\section{Hyperparameter Settings of the Regression Models}\label{appendix-H}
\setcounter{table}{0}\renewcommand{\thetable}{H\arabic{table}}\renewcommand{\theHtable}{H.\arabic{table}}
\setcounter{figure}{0}\renewcommand{\thefigure}{H\arabic{figure}}\renewcommand{\theHfigure}{H.\arabic{figure}}
\setcounter{equation}{0}\renewcommand{\theequation}{H.\arabic{equation}}\renewcommand{\theHequation}{H.\arabic{equation}}

This appendix reports the hyperparameter settings for all the models introduced in the Regression Models Training section. The Python package \texttt{sklearn} is used for Linear Regression and Random Forest, and the dedicated \texttt{xgboost} and \texttt{lightgbm} packages for the remaining two, with the exact parameter values reported in Tables~\ref{tab:H1}--\ref{tab:H3}.

\textbf{Linear Regression:} The LinearRegression model, imported from \texttt{sklearn.linear\_model}, is initialized without setting any hyperparameters, relying on its default configuration.

\textbf{Random Forest:} The object \texttt{RandomForestRegressor} is imported from \texttt{sklearn.ensemble} and it is initialized with the splitting criterion \texttt{absolute\_error}, minimizing the MAE at each node, together with the following settings, reported in Table~\ref{tab:H1}.

\begin{table}[htbp]
\centering
\caption[Hyperparameter settings for the Random Forest model]{Hyperparameter settings for the Random Forest model. }\label{tab:H1}
\small
\begin{tabular}{@{}rll@{}}
\toprule
Setting & Value & Description \\
\midrule
\texttt{n\_estimators} & \texttt{100} & Number of trees in the forest \\
\texttt{max\_depth} & \texttt{10} & Maximum depth of each tree \\
\texttt{min\_samples\_split} & \texttt{2} & Minimum number of samples required to split an internal node \\
\texttt{min\_samples\_leaf} & \texttt{1} & Minimum number of samples required at a leaf node \\
\texttt{max\_leaf\_nodes} & \texttt{None} & No limit on the number of leaf nodes per tree \\
\texttt{bootstrap} & \texttt{True} & Trees are built on bootstrap samples of the training data \\
\texttt{max\_samples} & \texttt{0.8} & Fraction of samples drawn for each tree when bootstrapping \\
\texttt{oob\_score} & \texttt{True} & Out-of-bag samples are used to estimate model performance \\
\texttt{max\_features} & \texttt{sqrt} & Number of features considered at each split, set to square root of the total \\
\texttt{n\_jobs} & \texttt{-1} & All available processors used for parallel computation \\
\bottomrule
\end{tabular}
\end{table}

\textbf{XGBoost:} The Booster model is imported from the \texttt{xgboost} package, and is initialized with the learning objective set to \texttt{reg:absoluteerror}, minimizing the MAE, together with the following settings, reported in Table~\ref{tab:H2}.

\begin{table}[htbp]
\centering
\caption[Hyperparameter settings for the Random XGBoost model]{Hyperparameter settings for the Random XGBoost model. }\label{tab:H2}
\small
\begin{adjustbox}{max width=\linewidth}
\begin{tabular}{@{}rll@{}}
\toprule
Setting & Value & Description \\
\midrule
\texttt{booster} & \texttt{gbtree} & Type of booster used, tree-based gradient boosting \\
\texttt{max\_depth} & \texttt{8} & Maximum depth of each tree \\
\texttt{min\_child\_weight} & \texttt{3} & Minimum sum of instance weight (Hessian) required in a child node \\
\texttt{gamma} & \texttt{0.1} & Minimum loss reduction required to make a further split on a leaf node \\
\texttt{eta} & \texttt{0.1} & Learning rate, shrinking each tree's contribution \\
\texttt{subsample} & \texttt{0.8} & Fraction of training instances sampled for each tree \\
\texttt{colsample\_bytree} & \texttt{0.8} & Fraction of features sampled when constructing each tree \\
\texttt{lambda} & \texttt{0.1} & \(L^{2}\) regularization term on leaf weights \\
\texttt{alpha} & \texttt{0.0} & \(L^{1}\) regularization term on leaf weights \\
\texttt{num\_rounds} & \texttt{1000} & Maximum number of boosting rounds \\
\texttt{early\_stopping\_rounds} & \texttt{100} & Number of rounds without validation improvement before training stops \\
\bottomrule
\end{tabular}
\end{adjustbox}
\end{table}

\textbf{LightGBM:} The \texttt{lightgbm.train} function, imported from the \texttt{lightgbm} package, is initialized with the learning objective set to a standard regression task (\texttt{regression}), together with the following settings, reported in Table~\ref{tab:H3}.

\begin{table}[htbp]
\centering
\caption[Hyperparameter settings for the Random LightGBM model]{Hyperparameter settings for the Random LightGBM model. }\label{tab:H3}
\small
\begin{tabular}{@{}rll@{}}
\toprule
Setting & Value & Description \\
\midrule
\texttt{metric} & \texttt{mae} & Evaluation metric, mean absolute error \\
\texttt{boosting\_type} & \texttt{gbdt} & Boosting type, gradient boosting decision tree \\
\texttt{num\_leaves} & \texttt{31} & Maximum number of leaves per tree, controlling model complexity \\
\texttt{max\_depth} & \texttt{-1} & Maximum tree depth, unbounded \\
\texttt{lambda\_l2} & \texttt{1.0} & \(L^{2}\) regularization term on leaf weights \\
\texttt{lambda\_l1} & \texttt{0.1} & \(L^{1}\) regularization term on leaf weights \\
\texttt{feature\_fraction} & \texttt{0.8} & Fraction of features randomly sampled for each tree \\
\texttt{bagging\_fraction} & \texttt{0.8} & Fraction of data randomly sampled for each iteration \\
\texttt{bagging\_freq} & \texttt{5} & Frequency, in iterations, at which bagging is performed \\
\texttt{learning\_rate} & \texttt{0.05} & Shrinkage rate applied to each tree's contribution \\
\texttt{min\_data\_in\_leaf} & \texttt{20} & Minimum number of samples required in a leaf node \\
\texttt{min\_gain\_to\_split} & \texttt{0.01} & Minimum loss reduction required to perform a split \\
\texttt{stopping\_rounds} & \texttt{100} & Number of rounds without validation improvement before training stops \\
\bottomrule
\end{tabular}
\end{table}

\clearpage
\section{Summary of Notation}\label{appendix-I}
\setcounter{table}{0}\renewcommand{\thetable}{I\arabic{table}}\renewcommand{\theHtable}{I.\arabic{table}}
\setcounter{figure}{0}\renewcommand{\thefigure}{I\arabic{figure}}\renewcommand{\theHfigure}{I.\arabic{figure}}
\setcounter{equation}{0}\renewcommand{\theequation}{I.\arabic{equation}}\renewcommand{\theHequation}{I.\arabic{equation}}

This appendix summarizes the mathematical symbols and indices used throughout the manuscript, grouped by topic. The LCDIs and their scalar (\(\mu\)) and interval-based (\(\beta\)) versions are excluded, as they are listed separately in Appendix~\ref{appendix-C}. Where a symbol takes a different meaning in a specific context, the context is stated explicitly.

{\small
\begin{longtable}{@{}>{\raggedright\arraybackslash}p{0.27\linewidth}>{\raggedright\arraybackslash}p{0.69\linewidth}@{}}
\caption[Summary of notation]{Summary of notation.}\label{tab:I1}\\
\toprule
Symbol & Description \\
\midrule
\endfirsthead
\multicolumn{2}{@{}l}{\emph{Table~\thetable\ continued from previous page}}\\
\toprule
Symbol & Description \\
\midrule
\endhead
\midrule
\multicolumn{2}{r@{}}{\emph{Continued on next page}}\\
\endfoot
\bottomrule
\endlastfoot
\multicolumn{2}{@{}l}{\textbf{Graph structure}} \\[2pt]
\(G = (V,A,C)\) & DAG with vertex set \(V\), arc set \(A\) and carrier set \(C\) \\
\(V = S \cup I \cup \{m\}\) & Vertex set, partitioned into supplier, intermediates, and manufacturer \\
\(S\);\ \(s, s_{i} \in S\) & Set of supplier sites; a generic supplier site \\
\(I\);\ \(i, i_{i} \in I\) & Set of intermediate vertices; a generic intermediate vertex \\
\(m\) & Manufacturer vertex, the unique sink of the graph \\
\(u, v, w, n\) & Generic vertices of \(V\) \\
\(A = A_{SI} \cup A_{II} \cup A_{IM}\) & Arc set, partitioned into \(S\to I\), \(I\to I\) and \(I\to M\) arcs \\
\(C = \{c_{1},\ldots,c_{k}\}\);\ \(c, c_{i} \in C\) & Set of carriers; a generic carrier \\
\(\ell_{i}\) & Third-party logistics (3PL) provider (Figure~\ref{fig:1}) \\
\(N(v)\) & Set of out-neighbours of vertex \(v\), \(N(v) = \{n \in V : (v,n) \in A\}\) \\
\(d_{\mathrm{out}}(v)\) & Out-degree of vertex \(v\) \\
\(|\cdot|\) & Cardinality of a set \\[4pt]
\multicolumn{2}{@{}l}{\textbf{Routes and paths}} \\[2pt]
\(r, r_{i} = (v_{a},v_{b})\) & Route, a directed arc from vertex \(v_{a}\) to vertex \(v_{b}\) \\
\(\pi, \pi_{i} = (v_{0},\ldots,v_{n})\) & Path, an ordered sequence of connected vertices \\
\(\rho, \rho_{i} = (r_{0},\ldots,r_{n-1})\) & Route sequence equivalent to a path \\
\(\pi^{\prime}(v_{i}),\ \rho^{\prime}(v_{i})\) & Remaining path and route sequence from vertex \(v_{i}\) to the manufacturer \\
\(\Pi(v) = \{\pi_{1},\ldots,\pi_{\ell}\}\) & Set of maximal paths from vertex \(v\) to the manufacturer \\
\(\ell\) & Number of maximal paths in \(\Pi(v)\) \\
\(\kappa\) & Path obtained by extending or truncating another path (Appendix~\ref{appendix-A}) \\
\(v(t)\) & Current vertex of a shipment at time \(t\) \\[4pt]
\multicolumn{2}{@{}l}{\textbf{Flows and routing probabilities}} \\[2pt]
\(f(v_{a},v_{b})\) & Flow on arc \((v_{a},v_{b})\), i.e.\ the number of shipments that traversed it \\
\(f^{c}(v_{a},v_{b})\) & Carrier-specific flow, restricted to shipments transported by carrier \(c\) \\
\(f_{v}(v),\ f_{v}^{c}(v)\) & Total outgoing flow, and total outgoing carrier-specific flow, of vertex \(v\) \\
\(f_{V}(S),\ f_{V}(M)\) & Total flow leaving the supplier sites and total flow entering the manufacturer \\
\(\{X_{t}\}_{t \in \mathbb{N}}\) & Discrete-time Markov process describing the location of a shipment \\
\(P_{uv},\ P_{uv}^{c}\) & Transition probability \(u\to v\), and its carrier-conditioned version \\
\(p = \{p_{1},\ldots,p_{\ell}\}\) & Path probability distribution over \(\Pi(v)\)\\
\(p_{i}\) & Probability of path \(\pi_{i}\) \\
\(p^{c} = \{p_{1}^{c},\ldots,p_{\ell}^{c}\}\) & Carrier-conditioned path probability distribution\\
\(p_{i}^{c}\) & Probability of path \(\pi_{i}\) given carrier \(c\) \\
\(\mathcal{P}\) & Random path drawn from \(\Pi(v)\) \\[4pt]
\multicolumn{2}{@{}l}{\textbf{Time}} \\[2pt]
\(t_{0}(s)\) & Order time \\
\(t_{1}(s,c)\) & Shipment time, at which the goods are handed over to the carrier \\
\(t_{n}(s,c)\) & Delivery time \\
\(t\) & Current time, i.e.\ the time of the most recent tracking update \\
\(t - t_{0},\ t - t_{1}\) & Elapsed time since order placement and since shipment \\
\(\tau(s,c,t_{1},t)\) & Normalized elapsed shipment time, \((t - t_{1})/\text{ST}_{\mu}(s,c)\) \\[4pt]
\multicolumn{2}{@{}l}{\textbf{Operators, parameters and uncertainty}} \\[2pt]
\(\mathbb{P}(\cdot),\ \mathbb{E}[\cdot]\) & Probability and expectation\\
\(\mathbb{E}_{\mathcal{P} \sim p}\) & Expectation over the path distribution \(p\) \\
\(X \sim \mathcal{D}\) & Random variable \(X\) distributed according to \(\mathcal{D}\) \\
\(\text{CI}_{\beta}[\cdot]\) & Two-sided confidence interval at confidence level \(\beta\) \\
\(L_{\beta},\ U_{\beta}\) & Lower and upper bounds of a confidence interval at level \(\beta\) \\
\(\beta\) & Confidence level; in the \(\text{DRT}_{\beta}\) heuristic, a scaling factor \\
\(\alpha,\ \alpha(\tau),\ \alpha(\tau;p)\) & TFST weighting factor in \([0,1]\), fixed or a linear or polynomial function of \(\tau\) \\
\(p,\ q(p)\) & Parameter of the polynomial $\alpha$ function and exponent, \(q(p) = (1-p)/p\) \\
\(O(\cdot)\) & Asymptotic upper bound (big-O notation) \\[4pt]
\multicolumn{2}{@{}l}{\textbf{Machine learning}} \\[2pt]
\(f_{\text{ML}}\) & Regression model estimating the Dynamic Route Time \\
\(\mathbf{x}(r,t)\);\ \(x_{i}(r,t)\) & Feature vector of route \(r\) at time \(t\); features of the \(i\)-th dataset sample \\
\(d(r)\) & Geodesic distance between the origin and destination vertices of route \(r\) \\
\(\gamma\) & Relative margin derived from the MAPE of the regression model \\
\(l_{\beta},\ u_{\beta}\) & Lower and upper multiplicative factors of the \(\text{DRT}_{\beta}\) interval, \(1 \mp \beta\gamma\) \\
\(L^{1},\ L^{2}\) & Regularization terms based on the \(L^{1}\) and \(L^{2}\) norms (Appendix~\ref{appendix-H}) \\[4pt]
\multicolumn{2}{@{}l}{\textbf{Evaluation metrics}} \\[2pt]
\(N\) & Number of predictions \\
\([l_{i},u_{i}]; l_{i},u_{i}\) & Predicted interval of the \(i\)-th prediction; lower and upper bounds\\
\(y_{i}\) & Observed value of the \(i\)-th prediction \\
\(\mathbf{1}\{\cdot\}\) & Indicator function, equal to 1 if the condition holds and 0 otherwise \\
\(\alpha\) & In Appendix~\ref{appendix-F}, the significance level (complement of the confidence level) \\
\(\text{IS}_{\alpha}(l_{i},u_{i};y_{i})\) & Pointwise interval score of the \(i\)-th prediction at significance level \(\alpha\) \\
\end{longtable}
}

\end{document}